%% file: main.tex
\PassOptionsToPackage{dvipsnames}{xcolor}
\documentclass[10pt]{article} 
\usepackage[accepted]{TMLR/tmlr}

\input{TMLR/math_commands.tex}

\usepackage{hyperref}
\usepackage{url}
\usepackage{cleveref}
\usepackage{algorithm}   
\usepackage{algpseudocode}
\usepackage{graphicx}
\usepackage{amsmath,amsfonts,amssymb}
\usepackage{amsthm}
\usepackage{wrapfig}
\newtheorem{definition}{Definition}
\newtheorem{proposition}{Proposition}
\newtheorem{lemma}{Lemma}

\usepackage{makecell}

\input{TMLR/preamble}

\title{Learning Rate Transfer and Feature Learning Across Depth for Constrained Spectral Optimizers: Complete Scion}

\author{\name Eline Pot \email eline.pot@centralesupelec.fr \\
      \addr CVN CentraleSupélec, 91190 Gif-sur-Yvette, France \\
      Framatome, 92400 Courbevoie, France
      \AND
      \name Laurent Lefebvre \email laurent.lefebvre@framatome.com\\
      \addr Framatome, 92400 Courbevoie, France
      \AND
      \name  Mathieu Segond \email mathieu.segond@framatome.com\\
      \addr Framatome, 92400 Courbevoie, France
      \AND
      \name Antonio Silveti-Falls \email antonio.silveti-falls@centralesupelec.fr \\
      \addr CVN CentraleSupélec, 91190 Gif-sur-Yvette, France
      }

\def\month{10}
\def\year{2026}
\def\openreview{\url{https://openreview.net/forum?id=TGiJpGVPNA}} 

\begin{document}

\maketitle

\begin{abstract}
Hyperparameter transfer across model depth is instrumental in training deep models, as it enables learning rates calibrated on shallow networks to be effectively applied to deeper architectures. The recent CompleteP framework establishes a $1/L$ residual scaling rule that demonstrates this for Adam-family optimizers. We study whether similar transfer extends to constrained optimization methods, specifically Frank-Wolfe and projected non-Euclidean normalized steepest descent, where updates are determined by linear minimization oracles. We derive scaling rules for these methods which confirm that the $1/L$ residual scaling applies, although the arguments differ from the Adam-family optimizers. Our analysis supports depth-wise transfer and width-wise transfer of the learning rate for the Scion algorithm while ensuring feature learning and avoiding the kernel-regime. In experiments on NLP transformers (nanoGPT on FineWeb), vision transformers (DeiT on ImageNet), and vision MLPs (MLP-Mixer on CIFAR-10), we observe learning rate transfer across depth and feature learning for appropriately chosen residual scaling \footnote{The code is available at \url{https://github.com/elinepot/CompleteSCION}.}.
\end{abstract}

\input{sec-intro}
\input{sec-related}
\input{sec-analysis}
\input{sec-conclusion}

\subsubsection*{Acknowledgments}
This work uses computational resources from Framatome GPU clusters CRONOS and SELENA, two HPC systems used by Framatome and EDF Group for large-scale simulations and GPU-accelerated workloads. It additionally uses computational resources from the ``Mésocentre'' computing center of Université Paris-Saclay, CentraleSupélec and École Normale Supérieure Paris-Saclay supported by CNRS and Région Île-de-France (https://mesocentre.universite-paris-saclay.fr/). Some experiments were also performed using HPC resources from GENCI–IDRIS (Grant 2025-AD011016724).

\clearpage
\bibliography{bibliography.bib}
\bibliographystyle{TMLR/tmlr}

\clearpage
\appendix
\input{sec-appendix}

\end{document}

%% file: TMLR/math_commands.tex
\usepackage{amsmath,amsfonts,bm}

\def\eqref#1{equation~\ref{#1}}
\def\Eqref#1{Equation~(\ref{#1})}

\def\Twoalgref#1#2{Algorithms \ref{#1} and \ref{#2}}

\def\1{\bm{1}}

\DeclareMathAlphabet{\mathsfit}{\encodingdefault}{\sfdefault}{m}{sl}
\SetMathAlphabet{\mathsfit}{bold}{\encodingdefault}{\sfdefault}{bx}{n}

\newcommand{\R}{\mathbb{R}}

\DeclareMathOperator*{\argmin}{arg\,min}

%% file: TMLR/preamble.tex
\usepackage{dsfont}
\usepackage{nicefrac}

\usepackage{slantsc} 

\DeclareMathOperator{\clip}{clip}
\DeclareMathOperator{\spectralclip}{spectral\_clip}

\DeclareMathOperator{\orthog}{orthog}

\usepackage{braket}

\newcommand{\lmo}{\operatorname{lmo}}

\usepackage{xspace}

\newcommand{\Scion}{{\sc Scion}\xspace}
\newcommand{\uScion}{{\sc Unconstrained Scion}\xspace}
\newcommand{\pScion}{{\sc Projected Scion}\xspace}

\usepackage{relsize}
\usepackage{graphicx} 
\usepackage{hyperref}

\usepackage{booktabs}
\usepackage{tabularx}

\newcommand{\bScion}{%
  \texorpdfstring{\textbf{S\scalebox{0.8}{CION}}}{SCION}\xspace%
}
\newcommand{\ubScion}{%
  \texorpdfstring{\textbf{U\scalebox{0.8}{NCONSTRAINED} S\scalebox{0.8}{CION}}}{UNCONSTRAINED SCION}\xspace%
}
\newcommand{\pbScion}{%
  \texorpdfstring{\textbf{P\scalebox{0.8}{ROJECTED} S\scalebox{0.8}{CION}}}{PROJECTED SCION}\xspace%
}

\newcommand{\itScion}{%
  \texorpdfstring{\textit{S\scalebox{0.8}{CION}}}{SCION}\xspace%
  }

\newcommand{\muP}{\mu \text{P}}

\usepackage{thmtools, thm-restate}

\usepackage{cleveref}

\crefformat{equation}{Equation~(#2#1#3)}
\crefrangeformat{equation}{Equations~(#3#1#4) to (#5#2#6)}

\usepackage{subcaption}
\usepackage{makecell}

\usepackage[dvipsnames]{xcolor}

\usepackage{float}
\usepackage{tcolorbox}
\usepackage{ulem}

%% file: sec-intro.tex
\section{Introduction}

Since their successful use in ImageNet \citep{imagenet_NIPS2012_c399862d}, deep neural networks have been increasingly used in complex tasks such as image classification \citep{Simonyan2014VeryDC,he2016deep,SzegedyGoingDeeper}, speech recognition \citep{HintonDeepNeuralNetworksForAcousticModeling,Graves2013SpeechRW,Amodei2015DeepS2}, and natural language processing \citep{sutskever2014sequence,vaswani2017attention,devlin2019bert}, where they have led to significant performance gains across benchmarks and applications. They are deployed for both difficult and critical applications, including medicine \citep{esteva2017dermatologist,topol2019high}, aeronautics \citep{julian2019guaranteeing,sun2019review}, and nuclear engineering \citep{tang2022deep,hu2021data}. 
Performance on complex tasks has been observed to increase with model scale, motivating the use of larger and heavier neural architectures, involving a high number of parameters.

There are many ways to scale deep neural networks, with the most common being increasing its width or depth.
However, increased performance is not the only consequence of scale. Indeed, as the scale increases so does the cost of training and tuning the models, making the ability to reuse hyperparameters across scales a pressing practical concern.

\begin{figure}
    \centering
    \includegraphics[width=\linewidth]{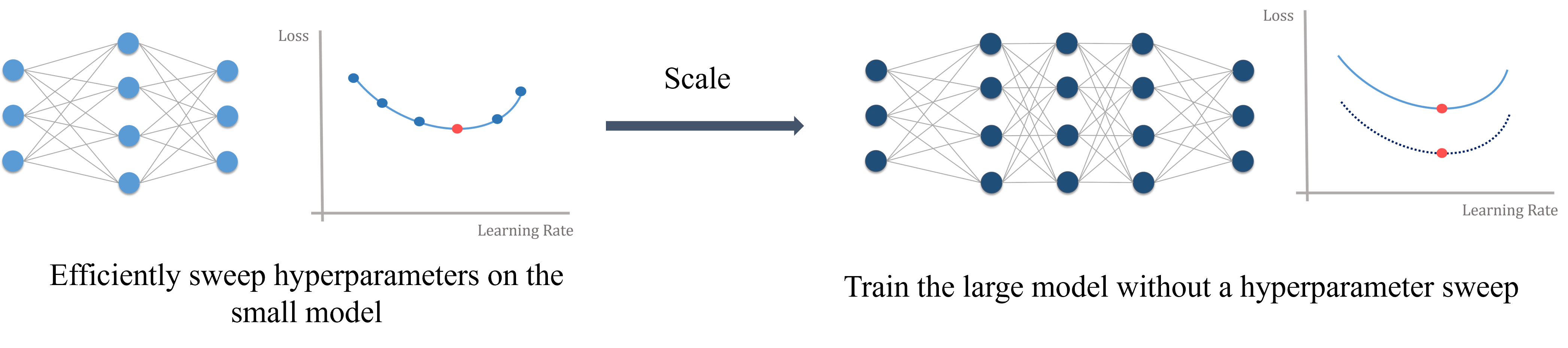}
    \caption{Practical learning rate transfer \citep{yang2021TPVZeroShot}. Define a small, computationally inexpensive version of the model (i.e. with small dimensions) and run a hyperparameter sweep to find the optimal learning rate $\gamma^*$.
    Then, train the large-scale target model directly using the optimal $\gamma^*$, bypassing expensive tuning at scale.}
    \label{protocol1}
\end{figure}

The Maximal Update Parametrization ($\mu$P) framework \citep{yang2021TPIVFeatureLeanringWidth,yang2023TPIVb} provides techniques to ensure that optimal hyperparameters remain invariant when the width of the model increases. 
This allows optimal hyperparameters found by tuning small scale models to be reliably transferred to larger ones.
Figure \ref{protocol1} provides practitioners a step-by-step framework for implementing and using the learning rate transfer. A well-designed parametrization guarantees that the optimal hyperparameter remains invariant across scales.
Additionally, $\mu$P prescribes parameter and learning‑rate scalings that preserve non‑degenerate \emph{feature learning} dynamics, in which the internal representations evolve to fit the data \citep{yang2021TPIVFeatureLeanringWidth}.
This avoids the \emph{lazy training} regime where the model dynamics behave like their linearization \citep{chizat2019lazy}.
However, the optimizer landscape has evolved significantly beyond the Stochastic Gradient Descent (SGD) and Adam \citep{kingma2014adam} updates originally considered in $\mu$P, leaving a gap.

Despite being introduced more than ten years ago \citep{carlson2015stochastic,carlson2015graphical,carlson2015preconditioned}, optimizers based on the spectral norm have recently reemerged as strong alternatives to standard coordinate-wise methods. In particular, \Scion \citep{pethick2025trainingdeeplearningmodels} (which also has an unconstrained variant) is a Conditional-Gradient-based algorithm \citep{frank_wolfe} that relies on a Linear Minimization Oracle (LMO) over norm-ball constraints and explores the choice of a norm across layers. Varying the norm at each layer induces layer-wise update magnitudes, which control the relative update scales between layers.
Updates based on the spectral norm have also attracted attention through their use in the Muon optimizer \citep{jordan2024muon}, which has gained popularity due to its performance in speed-running benchmarks \citep{modded_nanogpt_2024} and training industrial-scale large language models like the Kimi,
DeepSeek, and the Trinity series models \citep{team2025kimi,team2026kimi,deepseekai2026deepseekv4,singh2026arcee}.
Unlike SGD or Adam, these methods precondition the update to weight matrices in a layer-wise manner using matrix-level operations such as the matrix sign, rather than using coordinate-wise scaling. 
As a result, the update structure is fundamentally different, meaning that the standard $\mu$P parametrization and the Tensor Programs (TP) analytic framework \citep{yang2021TPIVFeatureLeanringWidth,yang2023TPIVFeatureLearningDepth} do not directly apply. 

While $\mu$P provides a principled framework for width-wise hyperparameter transfer, it does not directly address the effect of increasing depth. Several extensions have explored depth-wise hyperparameter transfer \citep{yang2023TPIVFeatureLearningDepth,bordelon2024depthwise,dey2026dontlazycompletepenables,qiu2025hyperparameter}. Among them, CompleteP \citep{dey2026dontlazycompletepenables} identifies three desiderata that enable joint width-and-depth-wise transfer: stable initialization, maximal feature updates, and a layer-wise non-lazy training regime. The lazy regime, related to the Neural Tangent Kernel regime, is one where features evolve like their linearization with respect to the parameters during training \citep{yang2021TPVZeroShot,dey2026dontlazycompletepenables}. 
However, existing instantiations of CompleteP focus on SGD-like optimizers and do not cover spectral-norm-based update rules.

In this work, we study the compatibility of the CompleteP parametrization with \Scion, referring to their combination as Complete Scion, and show that it satisfies all the desiderata outlined in CompleteP.
More precisely, we show that the learning rate tuned at small scale transfers to larger models without additional tuning, assuming a residual scaling of the form $L^{-\alpha}$ with $\alpha \in [\nicefrac{1}{2}, 1]$ and a learning rate scaled as $\gamma \propto L^{\alpha-1}$, for both \Scion and \uScion.

Our theoretical analysis establishes both width-wise and depth-wise hyperparameter transfer. 
For depth scaling, we confirm that the CompleteP desiderata (stable initialization and stable feature updates) hold for \Scion and \uScion,  
establishing depth-wise learning rate transfer. We show that stable feature updates are achieved in the $\alpha=1$ regime. 
We emphasize that, while our final optimization objectives align with the CompleteP framework, our derivations and proofs differ. 

Existing analyses in CompleteP focus on well-normalized coordinate-wise optimization methods such as Adam, whose updates are applied independently across parameters. In contrast, we focus on spectral optimizers like \Scion, which compute geometry-aware updates that are not regularized coordinate-wise, instead using linear minimization oracles over matrix-norm-ball constraints. As a consequence, the analytical techniques developed for coordinate-wise optimizers do not directly apply, and establishing depth-wise hyperparameter transfer requires a different derivation adapted to constrained and unconstrained spectral optimization.

For width scaling, we prove that the optimal learning rate is independent of hidden-layer width for \Scion (with weight decay), extending prior width-transfer results \citep{pethick2025trainingdeeplearningmodels,Yang2023ASpectralCondtion} to the constrained spectral setting. 
Finally, we consider a variant of \Scion which proposes to constrain the weights to be in a spectral ball or Stiefel manifold by using spectral clipping. We observe in our experiments that this projected version admits width-transfer.

We empirically validate these theoretical depth-wise and width-wise hyperparameter transfer results and the feature learning in multiple experiments across both vision and language tasks.
Learning rate transfer experiments on transformers (nanoGPT \citep{karpathy2023nanogpt} trained on FineWeb, and DeiT \citep{pmlr-v139-touvron21a} on ImageNet) and on MLPs (MLP-Mixer \citep{tolstikhin2021mlpmixerallmlparchitecturevision} trained on CIFAR‑10) show that hyperparameters tuned at small width and depth successfully transfer to larger models.

From a practical standpoint, by enabling principled width and depth-wise hyperparameter transfer, our results make these spectral-norm-based optimizers significantly cheaper to tune, in the same way that $\mu$P enabled practical scaling for Adam and SGD related optimization methods.

We summarize our contributions as follows:
\begin{itemize}
	\item We confirm that the CompleteP desiderata (stable initialization and stable feature updates) hold for \Scion and \uScion, under the $L^{-\alpha}$ residual scaling ($\alpha\in\{\nicefrac{1}{2},1\}$) with an appropriately scaled learning rate $\gamma \propto L^{\alpha-1}$, establishing depth-wise learning rate transfer. Our analysis is tailored to updates that take matrix structure into account, which are not addressed by the coordinate-wise analyses used in existing CompleteP formulations.
    \item We prove width-wise hyperparameter transfer for \Scion (with weight decay) and we empirically show it for Spectral Clipping, extending the width-transfer results of \cite{pethick2025trainingdeeplearningmodels,Yang2023ASpectralCondtion} to the constrained setting.
    \item We give practitioners a guide to tune layer-wise scale, and empirically validate the theoretical depth-wise and width-wise hyperparameter transfer results, as well as complete feature learning ($\alpha=1$ vs. $\alpha=\nicefrac{1}{2}$), on both transformers (modded nanoGPT on FineWeb, and DeiT on ImageNet) and MLPs (MLP-Mixer on CIFAR-10) across vision and language datasets.
\end{itemize}

%% file: sec-related.tex
\section{Related Work}

\paragraph{Width-wise hyperparameter transfer}
The Maximal Update Parametrization ($\muP$) \citep{yang2021TPIVFeatureLeanringWidth,yang2023TPIVb} initiated the Tensor Programs framework for hyperparameter transfer that determines how hyperparameters such as the learning rate or the initialization should vary with the model width going to infinity, ensuring that each layer exhibits stable learning and meaningful features (maximal feature learning). $\muP$ enables optimal hyperparameters tuned on a narrow model to transfer to wider models. This line of work established that width-wise hyperparameter transfer is well understood for coordinate-wise optimizers such as SGD and Adam, and inspired further analyses for width scaling \citep{yang2021TPVZeroShot,Yang2023ASpectralCondtion,bordelon2022self,chennuru2024feature}. 

\cite{kosson2025weight} study learning‑rate transfer for Adam under $\muP$, showing that the geometric alignment assumptions underlying $\muP$ hold only briefly at the start of training in practical large‑scale settings. They demonstrate that for Adam, it is instead independent weight decay that stabilizes representation updates across widths for most of training, reframing $\muP$’s effect as an implicit learning‑rate warmup.

Despite these refinements, works building on the $\mu$P framework focus on coordinate‑wise update structures and do not directly cover spectral‑norm‑based optimization methods.
More recent analyses have begun extending width-scaling to operator norm optimizers \citep{pethick2025training,xu2026widthscalingneuraloptimizers}, but do not address the depth-wise feature-learning framework studied in this paper.

\paragraph{Depth-wise hyperparameter transfer}
This idea has been extended, both in the Tensor Programs framework and beyond, to depth scaling for residual networks \citep{yang2023TPIVFeatureLearningDepth,bordelon2024depthwise,dey2026dontlazycompletepenables,qiu2025hyperparameter}. The insight behind these works is to scale the residual branch by a factor $L^{-\alpha}$ that depends on the depth $L$ of the network, with $\alpha \in [\nicefrac{1}{2}, 1]$, and to scale the learning rate accordingly.

More recently, CompleteP \citep{dey2026dontlazycompletepenables} formalized depth-wise hyperparameter transfer by introducing three desiderata: stable initialization, stable feature updates, and a non-lazy training regime. Within this framework, CompleteP shows that $\alpha=1$ ensures complete feature learning and improved performance compared to $\alpha=\nicefrac{1}{2}$ for Adam-based optimizers.\footnote{The authors of \cite{dey2026dontlazycompletepenables} argue that their results extend to any ``well-normed'' optimizer, and provide explicit derivations for SignSGD.} While CompleteP provides a general conceptual framework, its concrete instantiations and proofs focus on SGD-like update rules, leaving open the question of whether these desiderata can be satisfied by optimizers with fundamentally different update structures.

\cite{shigida2026learning} focus on the nGPT framework, restricting the model to a narrow class of architectures, and demonstrate learning rate transfer across both model dimension and token horizon. 
\cite{ren2026rethinking} constrain the model by design and implement a variant of $\mu$P using Frobenius norms for hypersphere parametrization. Utilizing the Muon optimizer under this Frobenius-sphere constraint, they achieve learning rate transfer across model width, depth, and training tokens by modifying the parametrization of the network itself.
In contrast, we operate in a more general setting, with constraints guaranteed through the optimization.

\cite{zheng2026spectral} study width-depth scaling in the $\mu$P setting from a spectral-condition perspective. In contrast, our work focuses on extending CompleteP-style depth-wise transfer to \Scion and \uScion, i.e. to a different optimizer class with geometry-aware spectral updates.

Our work builds on the CompleteP framework and instantiates it for a new family of optimizers. We adopt the same architectural setting and depth-dependent residual scaling, but study a different update rule based on \Scion. We show that, under appropriate scaling, both \Scion and \uScion satisfy the CompleteP desiderata, and we provide new proofs tailored to \Scion’s update dynamics. In this sense, our contribution extends CompleteP to a broader class of optimization methods.

\paragraph{Spectral-norm and geometry-based optimizers.}
A parallel line of work studies optimization methods that incorporate geometric or spectral information into their updates.
Shampoo \citep{gupta2018shampoo} introduces a tensor-wise preconditioning method, based on Kronecker-factored matrices. Shampoo uses these preconditioning matrices to whiten features along the left and right dimensions of the weight matrices. This significantly reduces the memory and computational cost compared to storing the full Hessian matrices at large scale.
SOAP \citep{vyas2025soap} extends Shampoo by combining tensor-structured preconditioning with Adam updates. It applies on-the-fly adaptive updates in the eigen-basis of the Kronecker preconditioner matrices \citep{pethick2025training}, improving the stability and reducing the computational cost of the training while preserving the scaling ability of Shampoo. In this work, we focus specifically on instant versions via LMO, noting that methods like \Scion operate precisely as instant Shampoo.

Spectral-norm-based optimizers constitute a distinct and increasingly popular class. 
Muon \citep{jordan2024muon} combines SGD with momentum and matrix normalization using Newton–Schulz iterations to produce orthogonal updates. The orthogonalization prevents updates from being dominated by a few directions, resulting in more balanced and robust training dynamics. Muon demonstrates rapid convergence and strong performances on large-scale deep learning tasks, including a speed record on CIFAR-10. 

Recent work \citep{filatov2026optimalscalingneedsoptimal} has also emphasized that the operator norm used for the last layer is fundamental to the scaling behavior in \Scion. \cite{filatov2026optimalscalingneedsoptimal} study scaling across both architecture and dataset with $\alpha=1$. Our work complements this perspective by analyzing the effect of $\alpha$ on both the scaling and feature learning aspects of \Scion. Similar to their work, we provide empirical results in Section~\ref{sec:radius_transfer} showing transfer of the layerwise learning rates as well.

\Scion \citep{pethick2025trainingdeeplearningmodels} is a Conditional Gradient algorithm that utilizes the LMO subject to norm-ball constraints, allowing for choice of layer-wise norms. 
From this perspective, \Scion plays a role for sign-based optimization \citep{Riedmiller1993ADA} analogous to that of Shampoo for Adam. Indeed, as Shampoo enriches Adam with structured matrix preconditioning, \Scion augments sign updates with explicit and more general geometric constraints.  
\Scion provides a unified framework on existing methods such as Muon \citep{jordan2024muon}, SignSGD \citep{bernstein2018signsgd} and NormalizedSGD \citep{hazan2015beyond}, that arise as special cases of unconstrained Stochastic Conditional Gradient under different norms (the spectral, the max-, and the Euclidean $L_2$ norms, respectively).
Thus, \Scion covers a broad family of geometric-aware methods, and has been implemented in the NVIDIA NeMo Emerging Optimizers library \citep{nemo_emerging_optimizers}.

Closest to this setting is the Modular Norm framework \citep{large2024scalable}, which establishes width‑wise and depth‑wise hyperparameter transfer using layer-wise normalized SGD. This direction was further developed and analyzed using linear minimization oracles in \cite{pethick2025trainingdeeplearningmodels}, which proves width‑wise transfer for \uScion and serves as the primary foundation for our analysis.
We highlight two important aspects that distinguish our work from \cite{large2024scalable}. First, we study the \Scion framework based on linear minimization oracles rather than layer-wise normalized SGD given in \cite{large2024scalable}; the update directions in \Scion are fundamentally different. Second, we study the effect of $\alpha$ beyond $\alpha=1$ on both hyperparameter transfer and feature learning.

Relatedly, \cite{qiu2025hyperparameter} study the width and depth scaling of the learning rate and the weight decay for matrix-preconditioned optimizers such as Shampoo, SOAP, and Muon+AdamW.
They do not consider $\alpha$ besides $\alpha=1$ and focus also on scaling the dataset size; in contrast, we are concerned with constrained optimizers, and the effect of $\alpha$ on their feature learning.

Additionally, \cite{islamov2026role} analyze hyperparameter scaling from a data-budget perspective, establishing principled rules for scaling the batch size, sequence length, and step size specifically for stochastic conditional gradient methods like \Scion.

%% file: sec-analysis.tex
\section{Complete \bScion}
\label{sec:complete_scion}

We first restate the desiderata required for effective feature learning and hyperparameter transfer.
We then show that these desiderata are satisfied by \Scion, both in its constrained and unconstrained forms. 

\subsection{Optimization problem}
\label{sec:optimization_pb}
    We consider training a neural network with $L$ hidden layers.
    We denote the full set of model parameters as $\mathbf{\Theta}=\{\theta^{(0)}, \theta^{(1)}, \dots, \theta^{(L+1)}\}$, where each $\theta^{(\ell)}, \ell \in \{0, \dots, L+1\}$ represents the parameters specific to layer $\ell$. 
    Each $\theta^{(\ell)}=(\theta^{(\ell, k)})_k$ collects all the parameters of the layer $\ell$.
    The training goal is to minimize the loss $\mathcal{L}(\mathbf{\Theta})$, formulated as
    \begin{equation}
        \min_{\mathbf{\Theta} \in \mathcal{X}} \mathcal{L}(\mathbf{\Theta})
    \end{equation}
    where $\mathcal{X}$ represents the parameter set, that is either unconstrained (full parameter space) or constrained to $\mathcal{X}=\mathcal{D}$. 
    We denote the partial gradient with respect to the $\ell$-th layer parameters as $\nabla_{\theta^{(\ell)}}\mathcal{L}(\mathbf{\Theta})$.
    The dependence on the full family of parameters $\mathbf{\Theta}$ will be implicit in the remainder of the paper, and we will simply write $\nabla_{\theta^{(\ell)}}\mathcal{L}$ and omit the dependence on $\mathbf{\Theta}$.
    
    In contrast to many prior works on hyperparameter transfer, we will not always assume $\mathcal{X}$ is the full parameter space, instead allowing for some constraint set $\mathcal{D}$ which is defined as the Cartesian product of layer-wise norm-balls with radii $\{\rho^{(0)}, \dots, \rho^{(L+1)}\}$, each constraining all parameter blocks of the corresponding layer.

    \begin{equation}
        \mathcal{D} = \prod_{\ell=0}^{L+1} \{ \theta^{(\ell)}: \lVert \theta^{(\ell, k)} \rVert \leq \rho^{(\ell, k)}, \forall k\}.
    \end{equation}

\begin{figure}[!ht]
        \centering
        \includegraphics[width=0.9\linewidth]{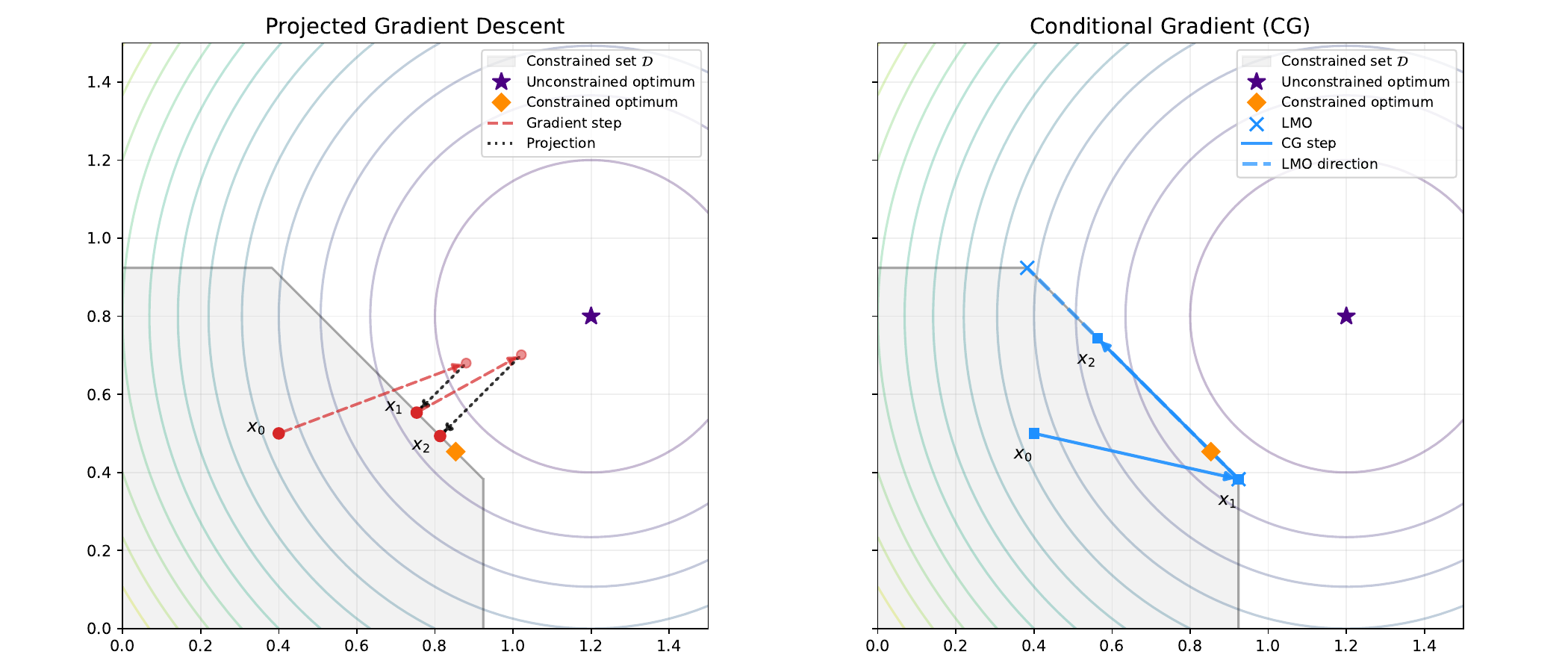}
        \caption{Projected gradient methods step outside the feasible set and then project back to the border of the constraint set, whereas Conditional Gradient methods preserve feasibility by updating through convex combinations with decreasing step sizes.}
        \label{fig:FW_vs_proj}
    \end{figure}

    There are two main ways to enforce these constraints: projection-based methods and Conditional-Gradient–type updates \citep{frank_wolfe}.
    In Conditional Gradient methods, each update is constructed as a convex combination of the current iterate and a point on the constraint set, ensuring that iterates remain within the feasible region. In contrast, projected gradient methods first take an unconstrained update step and then project the result back onto the constraint set at each iteration.
    An illustration of these two types of iterations can be found in Figure \ref{fig:FW_vs_proj}.
    In the next part, we detail the Stochastic version of Conditional Gradient, and the associated optimizer \Scion.

\subsection{\bScion and \ubScion}

    \Scion is a stochastic optimization framework based on the Stochastic Conditional Gradient (SCG), and its unconstrained version (uSCG), presented in \Twoalgref{alg:SCG}{alg:uSCG}, respectively.
    In uSCG, the parameters $\theta$ are optimized over an unconstrained parameter space denoted $\mathcal{X}$, as opposed to the constrained feasible set denoted $\mathcal{D}$ used in \Scion.
    Importantly, the use of weight decay can be interpreted within this framework as an instance of constrained version, \Scion, where the decay implicitly enforces norm‑ball constraints on the parameters, as discussed in \cite{pethick2025trainingdeeplearningmodels,pethick2025training}.

    \begin{figure}[ht!]
    \centering
        \begin{minipage}{0.48\linewidth}
        \begin{algorithm}[H]
        \caption{SCG}
        \label{alg:SCG}
        \textbf{Input:} Horizon $n$, initialization $\mathbf{\Theta}_1 \in \mathcal D$, $M_0 = 0$, 
        momentum $\beta_t\in [0,1)$, and stepsize $\gamma_t \in (0,1)$
        \begin{algorithmic}[1]
            \For{$t = 1, \dots, n$}
                \State Sample data $x \sim \mathcal P$
                \State $M_{t} \gets (1-\beta_{t}) \nabla_{\mathbf{\Theta}_t} \mathcal{L} + \beta_t M_{t-1}$
                \State $\mathbf{\Theta}_{t+1} \gets (1-\gamma_t) \mathbf{\Theta}_t + \gamma_t \lmo(M_t)$
            \EndFor
        \end{algorithmic}
        \end{algorithm}
    \end{minipage}
    \hfill
    \begin{minipage}{0.48\linewidth}
        \begin{algorithm}[H]
        \caption{Unconstrained SCG (uSCG)}
        \label{alg:uSCG}
        \textbf{Input:} Horizon $n$, initialization $\mathbf{\Theta}_1 \in \mathcal X$, $M_0 = 0$, 
        momentum $\beta_t \in [0,1)$, and stepsize $\gamma_t \in (0,1)$
        \begin{algorithmic}[1]
            \For{$t = 1, \dots, n$}
                \State Sample data $x\sim \mathcal P$
                \State $M_{t} \gets (1-\beta_{t}) \nabla_{\mathbf{\Theta}_t} \mathcal{L} + \beta_t M_{t-1}$
                \State $\mathbf{\Theta}_{t+1} \gets \mathbf{\Theta}_t + \gamma_t \lmo(M_t)$
            \EndFor
        \end{algorithmic}
        \end{algorithm}
    \end{minipage}
    \end{figure}
    
    SCG is classically used for constrained optimization problems as it ensures the iterates remain in a ball of constant norm. 
    These algorithms adapt the geometry to the problem by using norm-constrained Linear Minimization Oracles (LMOs). 
    The $\lmo$ is applied on the gradient, and its output minimizes a linear approximation to the function $\mathcal{L}$ we want to minimize under the norm-ball constraint set $\mathcal{D}$

     \begin{equation}
        \lmo(\nabla_{\theta^{(\ell)}} \mathcal{L}) \in \argmin_{\nu \in \mathcal{D}} \langle \nabla_{\theta^{(\ell)}}  \mathcal{L}, \nu \rangle.
    \end{equation}
    We note that the $\lmo$ provides only the \emph{direction} of the constrained steepest gradient descent.
    It does not give a step size or define the next iterate.
    Also, it is scale invariant: $\lmo(\mu  s)=\lmo(s)$ for any $\mu >0$.
        
    \Scion instantiates SCG and uSCG using layer-wise operator norms, resulting in the constrained variant (\Scion) and the unconstrained variant (\uScion),  respectively.
    By choosing the norm based on the architecture to be optimized, \Scion aligns the optimization direction with the model's structure, leading to faster and more efficient training, especially for large neural networks. 
    Update rules, associated LMOs, and per-layer learning rate scales are provided in Table \ref{tab:scion_rules} for typical configurations.
    Specific examples of the \uScion step and its layer-wise learning rate scales can be found in \Eqref{eq:exemple_uscion_lmos} and \Eqref{eq:exemple_updates}. 

    For instance, considering a one-step \uScion update with a $\texttt{Sign} \rightarrow \texttt{Spectral} \rightarrow \texttt{Sign}$ configuration for a 1-layer MLP with parameters $\{W^{(0)},W^{(1)},W^{(2)}\}$, the iterates at the iteration $t+1$ are given by:
    
    \begin{equation}
    \label{eq:exemple_uscion_lmos}
    \begin{aligned}
    \left[\begin{array}{l}W^{(0)}_{t+1} \\ W^{(1)}_{t+1} \\ W^{(2)}_{t+1}
    \end{array}\right]
    & = \left[\begin{array}{l}W^{(0)}_{t} \\ W^{(1)}_{t} \\ W^{(2)}_{t}\end{array}
    \right]
    +\gamma_t \lmo\left(\left[\begin{array}{c}M^{(0)}_t \\ M^{(1)}_t \\ M^{(2)}_t\end{array}\right]\right) 
    & \text { (by definition of \uScion) } \\ 
    & =\left[\begin{array}{l}W^{(0)}_{t} \\ W^{(1)}_{t} \\ W^{(2)}_{t}\end{array}\right]
    +\gamma_t\left[\begin{array}{l}\lmo_{\lVert \cdot \rVert_ {\text{Sign}}}\left(M^{(0)}_t\right) \\ \lmo_{\lVert \cdot \rVert_{\mathcal{S}_\infty}}\left(M^{(1)}_t\right) \\ \lmo_{\lVert \cdot \rVert_ {\text{Sign}}}\left(M^{(2)}_t\right)\end{array}\right]
    & \text { (the $\lmo$ is separable) } \\
    & =\left[\begin{array}{l}W^{(0)}_{t} \\ W^{(1)}_{t} \\ W^{(2)}_{t}\end{array}\right]-
    \gamma_t\left[\begin{array}{l}\rho_0\left(\frac{1}{d_{\text{in}}} \operatorname{sign}\left(M^{(0)}_t\right)\right) \\ \rho_1\left( \sqrt{\frac{d_1}{d_0}}\orthog\left(M^{(1)}_t\right)\right) \\ 
    \rho_2\left(\frac{1}{d_1} \operatorname{sign}\left(M^{(2)}_t\right)\right)\end{array}\right] 
    & \text { (compute the LMOs with the layerwise radii) }
    \end{aligned}
    \end{equation}
    where we used the separability of the $\lmo$ for the max norm $\lVert \mathbf{\Theta}\rVert=\max_{\ell}\{\lVert \theta^{(\ell)} \rVert_{\ell}\}$.

    Then, the difference between iteration $t+1$ and $t$ for each layer $\ell$, denoted $\Delta W^{(\ell)}$ can be expressed as:
    \begin{equation}
    \label{eq:exemple_updates}
        \begin{aligned}
            \Delta W^{(0)} &= \gamma_{t} \lmo_{\lVert \cdot \rVert_{\text{Sign}}}(M_t^{(0)}) =-\gamma_{t} \rho_{0} \left(\frac{1}{d_{\text{in}}} \operatorname{sign}\left(M^{(0)}_t\right)\right)\\
            \Delta W^{(1)} &= \gamma_{t} \lmo_{\lVert \cdot \rVert_{\mathcal{S}_\infty}}(M_t^{(1)}) =-\gamma_{t} \rho_{1} \left(\sqrt{\frac{d_1}{d_0}} \orthog \left( M^{(1)}_t \right)\right)\\
            \Delta W^{(2)} &= \gamma_{t} \lmo_{\lVert \cdot \rVert_{\text{Sign}}}(M_t^{(2)}) =-\gamma_{t} \rho_{2} \left(\frac{1}{d_{1}} \operatorname{sign}\left(M^{(2)}_t\right)\right).\\
        \end{aligned}
    \end{equation}
    
    \begin{table}[ht!]
        \centering
        \caption{Update rules and learning rate (LR) scaling for different architectural layers. This example utilizes the norm configuration $\texttt{Sign} \rightarrow \texttt{Spectral} \rightarrow \texttt{Sign}$ for the embedding, hidden, and un-embedding layers, respectively.
        We provide the specific operator norms and associated $\lmo$ applied to the average gradient $M^{(\ell)}$, whose reduced Singular Value Decomposition (SVD) is denoted $M^{(\ell)}=U^{(\ell)} \Sigma^{(\ell)} {V^{(\ell)}}^\top$ for all layers $\ell$.
        The parameters $\rho_e$, $\rho_h$, and $\rho_u$ designate the radii of the balls for the embedding, hidden, and un-embedding weights respectively.
        Following Recommendation 3.1 of \cite{pethick2025trainingdeeplearningmodels}, typical configurations include $\texttt{Spectral} \rightarrow \texttt{Spectral} \rightarrow \texttt{Sign}$ for image data, or $\texttt{ColNorm} \rightarrow \texttt{Spectral} \rightarrow \texttt{Sign}$ for $1$-hot inputs. For an exhaustive list of norms and associated LMOs, one can refer to \cite{pethick2025trainingdeeplearningmodels,pethick2025training}.}
        \scriptsize
        \begin{tabularx}{\textwidth}{l >{\centering\arraybackslash}X @{}c@{} >{\centering\arraybackslash}X @{}c@{} >{\centering\arraybackslash}X}
        \toprule
        & \textbf{Embedding} & $\rightarrow$ & \textbf{Hidden Layers} & $\rightarrow$ & \textbf{Un-embedding} \\
        & ($\ell=0$) &  & ($\ell \in [1, L]$) & & ($\ell=L+1$) \\
        \midrule
            \textbf{Norm} $\lVert \cdot \rVert$ & $\lVert \cdot \rVert_{\text{Sign}}$ &  & $\lVert \cdot \rVert_{\mathcal{S}_\infty}$ & & $\lVert \cdot \rVert_{\text{Sign}}$ \\
            \addlinespace
            \textbf{LMO} $\lmo (M^{(\ell)})$ & $-\rho_{e} \textcolor{Green}{\frac{1}{d_{\text{in}}}} \text{sign}(M^{(0)})$ & & $-\rho_{h} \textcolor{Green}{\sqrt{\frac{d_\ell}{d_{\ell -1}}}}U^{(\ell)}{V^{(\ell)}}^\top$ & & $-\rho_{u} \textcolor{Green}{\frac{1}{d_L}} \text{sign}(M^{(L+1)})$ \\
            \addlinespace
            \textbf{Scaled LR} $\gamma_\ell$ & $\gamma_{\text{base}}$ &  & $\textcolor{cyan}{L^{\alpha-1}}\gamma_{\text{base}}$ & & $\gamma_{\text{base}}$ \\
            \midrule
            \addlinespace
            \makecell[l]{\textbf{\ubScion Update} \\ $\Delta W^{(\ell)} = \gamma_\ell \lmo(M^{(\ell)})$} & $-\gamma_0 \rho_{e} \textcolor{Green}{\frac{1}{d_{\text{in}}}} \text{sign}(M^{(0)})$ && $-\gamma_\ell \rho_{h}\textcolor{Green}{\sqrt{\frac{d_\ell}{d_{\ell -1}}}}U^{(\ell)}{V^{(\ell)}}^\top$ && $-\gamma_{L+1} \rho_{u} \textcolor{Green}{\frac{1}{d_L}} \text{sign}(M^{(L+1)})$\\
            \addlinespace
            \makecell[l]{\textbf{\bScion  Update} \\ $\Delta W^{(\ell)} = \gamma_\ell ( \lmo(M^{(\ell)})-W^{(\ell)})$} 
            &  
            $-\gamma_0 (\rho_{e} \textcolor{Green}{\frac{1}{d_{\text{in}}}} \text{sign}(M^{(0)})+W^{(0)})$ 
            &&
            $-\gamma_\ell ( \rho_{h}\textcolor{Green}{\sqrt{\frac{d_\ell}{d_{\ell -1}}}}U^{(\ell)}{V^{(\ell)}}^\top + W^{(\ell)})$
             && 
             $-\gamma_{L+1} (\rho_{u} \textcolor{Green}{\frac{1}{d_L}} \text{sign}(M^{(L+1)}) + W^{(L+1)})$\\
            \bottomrule
        \end{tabularx}
        \label{tab:scion_rules}
    \end{table}
        
\subsection{Conditions for Depth-wise Transfer} 
\paragraph{Setup}

\begin{wrapfigure}[11]{r}{0.5\textwidth}
        \centering
        \includegraphics[width=0.5\textwidth]{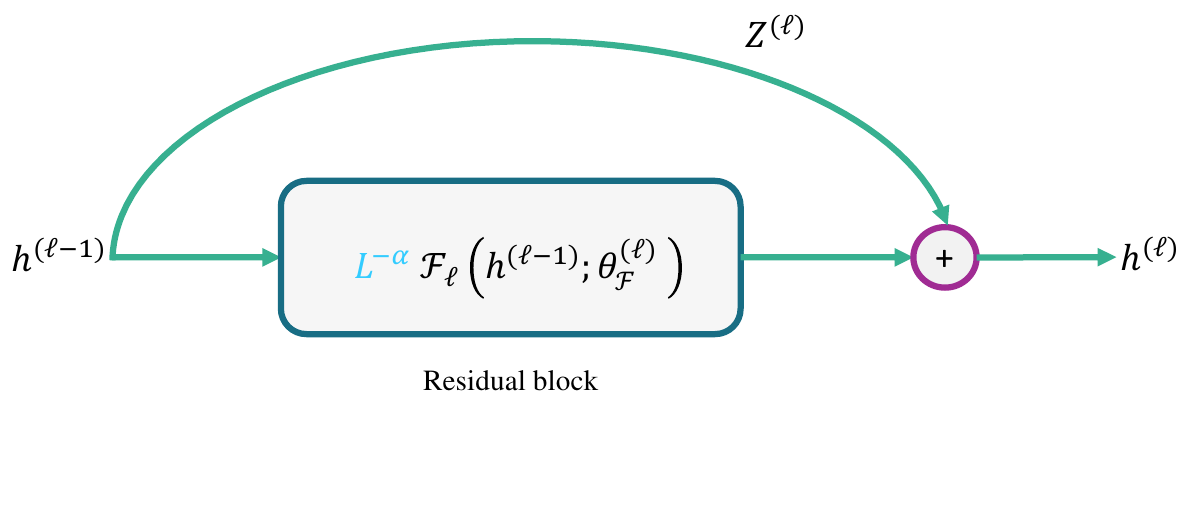}
        \vspace{-30pt}
        \caption{Residual block mechanism with depth-wise scaling \textcolor{cyan}{$L^{-\alpha}$}.}
        \label{fig:ResBlock}
        \vspace{50pt}
    \end{wrapfigure}

    In this section, we consider a residual architecture of depth $L$,  defined as
    \begin{equation}
    \label{eq:setup}
        \begin{aligned}
        h^{(0)} &= W^{(0)} x \\
        h^{(\ell)} &= Z^{(\ell)} h^{(\ell - 1)} + L^{-\alpha} \mathcal{F}_\ell (h^{(\ell - 1)} ; \theta_{\mathcal{F}}^{(\ell)}) 
            \quad (\forall \ell \in \{1, \dots, L\}) \\
        f &= W^{(L+1)} \phi(h^{(L)}),
        \end{aligned}
    \end{equation}   
    where $x \in \mathbb{R}^{d_{\text{in}}}$ are the input data, $h^{(\ell)} \in \mathbb{R}^{d_\ell}, \ell \in \{0, \dots, L\}$ are the hidden states, $Z^{(\ell)} \in \R^{d_\ell \times d_{\ell-1}}$, and $f \in \mathbb{R}^{d_{\text{out}}}$ is the output after nonlinear activation $\phi$. The embedding and un-embedding matrices are respectively $W^{(0)} \in \mathbb{R}^{d_0 \times d_{\text{in}}}$ and $W^{(L+1)} \in \mathbb{R}^{d_{\text{out}} \times d_L}$. The residual block $\mathcal{F}_\ell (h^{(\ell - 1)} ; \theta_{\mathcal{F}}^{(\ell)}), \ell\in \{1, \dots, L\}$ can be an attention block, or an MLP (cf. Figure \ref{fig:ResBlock}), and $\theta_{\mathcal{F}}^{(\ell)}$ is its trainable parameter.
    Following the notation introduced in Section \ref{sec:optimization_pb}, the parameters for the hidden layer $\ell$ are represented by the set $\theta^{(\ell)}=\{Z^{(\ell)}, \theta_{\mathcal{F}}^{(\ell)}\}$, and the parameters for the embedding and un-embedding layers are represented by $\theta^{(0)}=\{W^{(0)}\}$ and $\theta^{(L+1)}=\{W^{(L+1)}\}$ respectively.

    Essentially, the residual branch is scaled with a depth-dependent factor $L^{-\alpha}$, where the exponent $\alpha$ dictates how the contribution of each residual block scales as the total depth $L$ grows. This parametrization follows the framework of Tensor Programs and infinite-depth limits \citep{yang2023TPIVFeatureLearningDepth,bordelon2024depthwise}.

    The choice $\alpha=\nicefrac{1}{2}$ corresponds to the $\mu$P regime studied in Tensor Programs, where the cumulative contribution of the residual stream remains non-trivial in the infinite-depth limit and feature learning is preserved \citep{yang2023TPIVFeatureLearningDepth,yang2021TPVZeroShot,bordelon2024infinite}.
    
    Conversely, recent work suggests advantages for linear scaling. Indeed,    \cite{dey2026dontlazycompletepenables} argue that $\alpha=1$ is the only value that enables stable training, maximal updates and feature learning at infinite dimensions.

    CompleteP \citep{dey2026dontlazycompletepenables} defines three desiderata, that provide a principled framework for understanding why this parametrization works, and what criteria the parametrization should respect in order to observe hyper-parameter transfer both in width and depth, as well as non-lazy learning. 

    Following the literature \citep{dey2026dontlazycompletepenables,qiu2025hyperparameter,zheng2026spectral}, we enumerate the desiderata for depth-scaling below.

    \paragraph{Notation}
    In the analysis, we will use the following notation:
    \begin{equation*}
        \begin{aligned}
        f(n) = \Theta(g(n)) & \iff \big(\exists\,   (c, C) \in (0,+\infty)^2\big)  \quad c g(n) \leq f(n) \leq C g(n) \quad \text{as}\;n \rightarrow \infty, \\
        f(n) = \mathcal{O}(g(n)) & \iff \big(\exists\, C \in (0,+\infty)\big)\quad  f(n) \leq C g(n) \quad \text{as}\;n \rightarrow \infty, \\
        f(n) = \Omega(g(n)) & \iff \big(\exists\,c \in (0,+\infty)\big)\quad   cg(n) \leq f(n)\quad \text{as}\;n \rightarrow \infty.\\
        \end{aligned}
    \end{equation*}
    Note that the notation $\Theta()$ is different from the full parameter set mentioned in the previous subsections and written in bold as $\mathbf{\Theta}$.
    
    We will also use $\| \cdot \|_{2}$ and $\| \cdot \|_{\mathcal{S}_\infty}$ to denote the Euclidean norm and the spectral norm, respectively.
    
    \begin{restatable}[Stable Initialization]
    {des}{RestatableDesStableInit}
    \label{des:1}
        The hidden layers and the output remain stable (preventing signals from exploding or vanishing as they pass through many layers) at the initialization when $L$ and  $d_\ell$ become increasingly large:
        \begin{equation}
             (\forall \ell \in \{1,\ldots,L\})\;\;\frac{1}{d_\ell}\| h^{(\ell)}\|_{2}^2 \in \Theta (1)  \; \; \text{and} \;\; \lVert f\rVert_2^2 \in \mathcal{O}(1),
             \quad \text{as }\; d_\ell \rightarrow \infty, L \rightarrow \infty.
        \end{equation}
    \end{restatable}
        This property indicates how to adjust the variance of the initialization of the weights. In \cite{dey2026dontlazycompletepenables}, it is  highlighted that this imposes $\alpha \geq \frac{1}{2}$, since otherwise the residual stream variance diverges when the depth is scaled.

    \begin{restatable}[Stable Feature Update]
    {des}{RestatableDesStableUpdate}\footnote{The corresponding Desideratum in \cite{dey2026dontlazycompletepenables} has $\frac{1}{d_\ell} \lVert \Delta_{\theta^{(\ell)}} h^{(\ell)}\rVert_2^2 = \Theta(\frac{1}{L})$, which slightly differs from ours, as we discuss in Appendix \ref{appdx_sec:completeP_style_analysis}.}
    \label{des:2}
    The weights of each residual block should contribute with order $1/L$ to feature movements, and each non-residual block should contribute with constant order (i.e. each non-residual block has a meaningful impact on the residual stream):
        \begin{equation}
            \begin{aligned}
            & \frac{1}{\sqrt{d_\ell}}\lVert \Delta_{\theta_{\mathcal{F}}^{(\ell)}, Z^{(\ell)}} h^{(\ell)} \rVert_{2} \in \Theta \left(\frac{1}{L} \right) \quad (\forall \ell \in \{1,\ldots,L\}),\\
            & \frac{1}{\sqrt{d_0}}\lVert \Delta W^{(0)} x \rVert_{2}\in \Theta(1), \\
             & \frac{1}{\sqrt{d_{\mathrm{out}}}}\lVert \Delta W^{(L+1)} h^{(L)} \rVert_{2} \in \Theta(1), \\
            \end{aligned}
        \end{equation}
        where $\Delta$ designates the variation of the parameters between two successive iterations.
    \end{restatable}
         
        This criterion is the same as the maximal update condition when $L$ is a fixed constant, since we have  $\Theta(1/L)=\Theta(1)$ with respect to width-scaling \citep{yang2021TPIVFeatureLeanringWidth,Yang2023ASpectralCondtion}. Moreover, \cite{dey2026dontlazycompletepenables} show that this criterion imposes $\alpha \leq 1$, a result also proven in \cite{yang2023TPIVFeatureLearningDepth}.

        As we shall see in the next subsections, the first and second desiderata can be guaranteed to be satisfied by \Scion and \uScion for any $\alpha \in [\nicefrac{1}{2}, 1]$, under the condition that the learning rates, $\gamma$ and $\gamma_Z$, used to optimize the parameters $\{\theta_{\mathcal{F}}^{(\ell)}\}_{\ell \in \{1, \dots, L\}}$ and the operators $\{Z^{(\ell)}\}_{\ell \in \{1, \dots, L\}}$, respectively, are scaled as follows:
        \begin{equation}
            \gamma = \gamma_{\text{base}} L^{\alpha-1} \quad \text{ and } \quad  \gamma_Z = {\gamma_Z}_{\text{base}} L^{-1}.
        \end{equation}

        The learning rates relative to the embedding and un-embedding parameters do not need depth-wise scaling,
        \begin{equation}
        \begin{aligned}
            &\gamma_{W^{(0)}} =  {\gamma_{W^{(0)}}}_{\text{base}}  \quad \text{ and } \quad  \gamma_{W^{(L+1)}} =  {\gamma_{W^{(L+1)}}}_{\text{base}}.
        \end{aligned}
        \end{equation}

    We can interpret this in the unconstrained case as a \emph{scale-dependent} operator norm that encodes the scaling geometry automatically, so that the optimal learning rate does not change and these scalings are encoded solely in the norm choice.
    Specifically, the fixed magnitude of the $\lmo$ induces a guaranteed bound on the maximum norm of the parameters, ensuring stable scaling.

\subsection{\bScion and \ubScion satisfy Desiderata \ref{des:1} and \ref{des:2}}
\label{sec:derivations_depth_transfer_scion_uscion}

    \subsubsection{Fundamental setup} As a simplified model, consider the block $\mathcal{F}_\ell (h^{(\ell - 1)} ; \theta_{\mathcal{F}}^{(\ell)})$ given by a one-layer MLP, for $\ell \in \{1, \dots, L\}$, that is
    \begin{equation}
        h^{(\ell)} = Z^{(\ell)} h^{(\ell - 1)} + L^{-\alpha} W^{(\ell)} \phi(h^{(\ell - 1)}) 
    \end{equation}
    where $\phi$ is a Lipschitz-continuous activation function. In this case, the trainable parameter of the residual block $\theta_{\mathcal{F}}^{(\ell)}$ is the weight matrix $W^{(\ell)} \in \R^{d_\ell \times d_{\ell-1}}$ for $\ell \in \{1, \dots, L\}$. 
    In our analysis, we consider that the operator matrices applied to the residual stream $Z^{(\ell)} \in \R^{d_\ell \times d_{\ell-1}}$ for $\ell \in \{1, \dots, L\}$, are learned and updated during training.
    This generalizes the standard identity shortcut to a learnable linear transformation, aligning with recent architectural paradigms that introduce parameterized transformations into the residual path. 

    For instance, manifold-constrained hyper-connections (mHC) \citep{xie2025mhc}, which uses the $\lmo$ for the Birkhoff polytope, utilizes learnable mixing to mix features in the residual stream, and \cite{team2026attention} use an attention mechanism over the preceding layer output. In this case, following the notation defined in Section \ref{sec:optimization_pb}, the parameters for the hidden layer $\ell$ are simply the pair of matrices $\theta^{(\ell)}=\{Z^{(\ell)}, W^{(\ell)}\}$, and the parameters for the embedding and un-embedding layers are $\theta^{(0)}=\{W^{(0)}\}$ and $\theta^{(L+1)}=\{W^{(L+1)}\}$,  respectively.
    In our analysis, we omit biases, but, in practice, when included they are scaled in the same way as the weight matrices, consistent with \cite{dey2026dontlazycompletepenables}.
        
    \subsubsection{Modeling assumptions}
    \label{sec:assumptions_tools}

    We focus on the worst-case analysis for a batch size of $1$. 
    In practice and in our experiments, large batches are used and the results remain in accordance with our theoretical conclusions.
        
    \paragraph{Initialization}    
        As in \cite{pethick2025trainingdeeplearningmodels}, we suppose we initialize with either a semi-orthogonal or Gaussian matrix, and apply the $\lmo$ with the spectral norm for every layer $\ell$.
        We note that using a scaled sign operator norm on the embedding or un-embedding layers is also viable, consistent with the analysis in \cite{dey2026dontlazycompletepenables}.

        In particular, this leads (with high probability) to
        \begin{equation}
        \label{eq:hyp_init_and_update_weight}
             \lVert W^{(\ell)}\rVert_{\mathcal{S}_\infty} = \Theta \left(\sqrt{\frac{d_{\ell}}{d_{\ell-1}}} \right) 
             \quad \text{ and } \quad 
             \lVert \Delta W^{(\ell)}\rVert_{\mathcal{S}_\infty} = \Theta\left(\sqrt{\frac{d_{\ell}}{d_{\ell-1}}}\right).
        \end{equation}
        
        We suppose the same assumptions apply for the matrices $Z^{(\ell)}, \ell \in \{1, \dots,L\}$.

        For technical purposes, we will furthermore assume that the data x follows some distribution $\mathcal{P}$ such that $\lVert x \rVert_2=\Theta(\sqrt{d_{in}})$. This is often the case for images \citep{Yang2023ASpectralCondtion} for instance.

\paragraph{Scale of the activation function $\phi$}
        We assume that the hidden features $\phi(h^{(\ell)})$ have the same scale before and after the application of the activation function $\phi$ for all the layers $\ell$:
        \begin{equation}
        \label{eq:hyp_scale_activation_function}
            \lVert \phi(h^{(\ell)}) \rVert_2 = \Theta(\lVert h^{(\ell)} \rVert_2).
        \end{equation}
        It is the same assumption as in the Assumption 2 of \cite{Yang2023ASpectralCondtion}. This behavior is typical for most activation functions since they are designed to operate on inputs of order one and produce outputs that neither blow up nor collapse.

    \paragraph{Alignment of weight matrices}
        We assume that, for any step $t$, for all layers $\ell$, we have
        \begin{equation}
        \label{eq:hyp_alignment}
            \begin{aligned}
                \lVert W^{(\ell)} \phi(h^{(\ell-1)}) \rVert_2^2  & = \Theta(\lVert W^{(\ell)} \rVert_{\mathcal{S}_\infty}^2 \lVert \phi(h^{(\ell-1)})\rVert_2^2),\\
                 \lVert W^{(\ell)} h^{(\ell-1)} \rVert_2^2  & = \Theta(\lVert W^{(\ell)} \rVert_{\mathcal{S}_\infty}^2 \lVert h^{(\ell-1)}\rVert_2^2).\\
            \end{aligned}
        \end{equation}
        
        While this modeling assumption is not universally applicable, it has been empirically validated \citep{Yang2023ASpectralCondtion} and consistently yields predictive results.

        We suppose the same assumptions apply for the matrices $Z^{(\ell)}, \ell \in \{1, \dots,L\}$.  

        In Figure \ref{fig:weight_alignment_W_and_Z_with_phi_h}, we empirically verify that this is a reasonable assumption, showing that it holds for various depths and multiple steps.

\paragraph{Non-cancellation assumptions}
\label{par:non_cancellation_assumptions_depth_transfer}
We make the following assumptions on the alignment between updates and the current parameters. For any layer $\ell$, and training step $t$:
\begin{enumerate}
    \item The addition of the residual block with the previous state does not perfectly cancel.
    \begin{equation}
    \label{eq:alignment_sum_res_block}
    \lVert h^{(\ell)} \rVert_2^2 = \Theta \left( \lVert Z^{(\ell)} h^{(\ell-1)} \lVert_2^2+ \frac{1}{L^{2\alpha}} \lVert W^{(\ell)} \phi(h^{(\ell -1)}) \rVert_2^2 \right).
    \end{equation}
    \item The updates of $Z^{(\ell)}$ and $W^{(\ell)}$ do not perfectly cancel.
    \begin{equation}
    \label{eq:alignment_sum_update_res_block}
    \lVert \Delta_{W^{(\ell)}, Z^{(\ell)}}h^{(\ell)} \rVert_2^2 = \Theta \left( \lVert \Delta Z^{(\ell)} h^{(\ell-1)} \lVert_2^2+ \frac{1}{L^{2\alpha}} \lVert \Delta W^{(\ell)} \phi(h^{(\ell -1)}) \rVert_2^2 \right).
    \end{equation}
    \item The updates and the current quantities do not perfectly cancel.
    \begin{equation}
    \label{eq:alignment_init_and_update}
        \begin{aligned}
            \lVert W_{t+1}^{(\ell)} \rVert_{\mathcal{S}_\infty}^2
            & = \Theta \left( 
            \lVert W_{t}^{(\ell)} \rVert_{\mathcal{S}_\infty}^2
            + \lVert \Delta W^{(\ell)} \rVert_{\mathcal{S}_\infty}^2
            \right) \\
            \text{ and } \quad
             \lVert Z_{t+1}^{(\ell)} \rVert_{\mathcal{S}_\infty}^2
            & = \Theta \left( 
            \lVert Z_{t}^{(\ell)} \rVert_{\mathcal{S}_\infty}^2
            + \lVert \Delta Z^{(\ell)} \rVert_{\mathcal{S}_\infty}^2
            \right) \\
        \end{aligned}
    \end{equation}
    \item The current quantities and the $\lmo$s do not perfectly cancel.
    \begin{equation}
    \label{eq:alignment_sum_update_res_block_with_scion}
    \begin{aligned}
         &\lVert \Delta W^{(\ell)} \phi(h^{(\ell-1)})\rVert_2^2
         = \Theta \left( \lVert \gamma  W^{(\ell)} \phi(h^{(\ell-1)})\rVert_2^2 + \lVert \gamma  \lmo({M_t^{W^{(\ell)}}})
         \phi(h^{(\ell-1)})\rVert_2^2
         \right) \\
        \text{ and } \quad
        &\lVert\Delta Z^{(\ell)} h^{(\ell-1)}\lVert_2^2
        = \Theta \left(
        \lVert \gamma_Z  Z^{(\ell)} h^{(\ell-1)}\lVert_2^2 + \lVert \gamma_Z  \lmo({M_t^{Z^{(\ell)}}}) h^{(\ell-1)}\lVert_2^2
        \right).
    \end{aligned}
    \end{equation}
\end{enumerate}
These assumptions are empirically verified across depths and optimization steps, in Figures
\ref{fig:alignment_sum_res_block},
\ref{fig:alignment_sum_update_res_block}, and
\ref{fig:alignment_sum_update_res_block_with_scion} (the second row of Figure \ref{fig:alignment_sum_update_with_scion_width} also gives an analogous result to \Eqref{eq:alignment_init_and_update}). 

\paragraph{Update expression of $W^{(\ell)}$}
Since we consider a residual network, we need to compute, for every layer $\ell$
\begin{equation}
\label{eq:delta_W_uscion_scion}
    \Delta W^{(\ell)} = 
    \left \{
    \begin{array}{ll}
        \gamma \lmo(\nabla_{W^{(\ell)}}\mathcal{L})  & \text{ with \uScion} \\
         \gamma \lmo(\nabla_{W^{(\ell)}}\mathcal{L}) -\gamma W^{(\ell)} & \text{ with \Scion}.
    \end{array}
    \right.
\end{equation}

Further details about the updates and the LMOs expressions can be found in Table \ref{tab:scion_rules}, in \Eqref{eq:exemple_uscion_lmos} and in \Eqref{eq:exemple_updates}.

The same update expressions apply for the matrices $Z^{(\ell)}, \ell \in \{1, \dots,L\}$.  

To simplify the analysis, we consider the momentum $\beta^{(\ell)}=0$ for each layer $\ell$. As established in our analysis, the desiderata can be ensured by imposing suitable conditions on the spectral norm of the updates. Since the LMO has a fixed spectral norm regardless of whether momentum is used, we expect its behavior to remain unchanged. This expectation is further supported by our experiments, in which we considered nonzero momentum values.

\subsubsection{Desideratum \ref{des:1} is satisfied for \bScion and \ubScion}
\label{sec:condition1_for_scion}

    Our next result considers semi-orthogonally initialized matrices with $\lVert W^{(\ell)} \rVert_{\mathcal{S}_\infty} = \Theta(\sqrt{\frac{d_{\ell}}{d_{\ell-1}}})$ and $\lVert Z^{(\ell)} \rVert_{\mathcal{S}_\infty} = \Theta(\sqrt{\frac{d_{\ell}}{d_{\ell-1}}}), \forall \ell \in \{1, \dots, L\}$, and shows that Desideratum \ref{des:1} holds, extending the result in \cite{dey2026dontlazycompletepenables} for Gaussian variables initialization.

    \begin{restatable}[\Scion and \uScion verify Desideratum \ref{des:1}]{proposition}{ScionFirstDes}
    \label{prop:des1}
        Under the assumptions defined in the subsection \ref{sec:assumptions_tools}, the hidden layers and the output remain stable at the initialization when $L$ and $d_\ell$ become increasingly large.
    \end{restatable}

    We provide a sketch of the demonstration for \Scion and \uScion, where the same arguments hold. Complete derivations are available in Appendix \ref{appdx-proof:cond1}.
    
    \begin{proof}[Proof sketch of Proposition \ref{prop:des1}]
    \label{proof_sketch_Cond1}
         Let $\ell \in \{1, \dots, L\}$.
         By using the alignment of weight matrices in \Eqref{eq:hyp_alignment}, the scale of the activation function $\phi$ in \Eqref{eq:hyp_scale_activation_function}, and the weight initialization in \Eqref{eq:hyp_init_and_update_weight}, we have
             \begin{equation*}
                \begin{aligned}
                 \lVert h^{(\ell)} \rVert_2^2 &= \Theta( \lVert Z^{(\ell)} h^{(\ell - 1)} \rVert_2^2 + \frac{1}{L^{2\alpha}} \lVert W^{(\ell)} \phi (h^{(\ell - 1)}) \rVert_2^2) \\
                 & = \Theta \left(\lVert Z^{(\ell)} \rVert_{\mathcal{S}_\infty}^2 \lVert h^{(\ell - 1)} \rVert_2^2 + \frac{1}{L^{2\alpha}} \lVert W^{(\ell)}   \rVert_{\mathcal{S}_\infty}^2 \lVert h^{(\ell - 1)} \rVert_2^2 \right)
                 \\
                  & = \Theta \left(\prod_{k=1}^{\ell} (\frac{d_k}{d_{k-1}} \times(1 + \frac{1}{L^{2\alpha}})) d_0 \right)\\
                   & = \Theta \left( d_\ell \prod_{k=1}^{\ell} (1 + \frac{1}{L^{2\alpha}}) \right).
                 \end{aligned}
            \end{equation*}
            
            Then, $\frac{1}{d_L} \lVert h^{(L)} \rVert_2^2 = \Theta  \left( (1 + \frac{1}{L^{2\alpha}} )^L  \right)$ converges when $L$ goes to infinity if and only if $\alpha \geq \frac{1}{2}$.
                       
             Moreover,
             \begin{equation}   
                \begin{aligned}
                \frac{1}{d_\ell} \lVert h^{(\ell)} \rVert_2^2 & = \Theta (1) \quad \forall \ell \in \{1, \dots, L\} \text{ for } \alpha \geq \frac{1}{2},\\
                \end{aligned}
            \end{equation}
             
             and, for $f = W^{(L+1)} \phi(h^{(L)})$, we have
             \begin{equation}
                 \lVert f \rVert_2^2 =\Theta( \lVert W^{(L+1)} \rVert_{\mathcal{S}_\infty}^2  \lVert h^{(L)}\rVert_2^2 )
                 = \Theta(\frac{d_{\text{out}}}{d_L} \times d_L) = \mathcal{O}(1).
            \end{equation}
             
    \end{proof}

\subsubsection{Desideratum \ref{des:2} is satisfied for \bScion and \ubScion}
\label{sec:condition2_for_scion}

    Following the formulation of Desideratum \ref{des:2}, we study the change in $h^{(\ell)}$ when we update $\theta^{(\ell)}$ and $Z^{(\ell)}$ only. In our setting, we analyze the change in $h^{(\ell)}$ induced by the updates of $W^{(\ell)}$ and $Z^{(\ell)}$. 

    \begin{restatable}[\Scion and \uScion verify Desideratum \ref{des:2}]{proposition}{ScionSecondDes}
    \label{prop:des2}
        Under the assumptions defined in subsection \ref{sec:assumptions_tools}, for \Scion or \uScion, for any training step $t$, the weights of each block contribute with order $1/L$ to feature movements, and each non-residual block contributes with constant order when $L$ becomes increasingly large.
    \end{restatable}

    We provide a sketch of the demonstrations for \Scion. Complete derivations for \Scion and \uScion are available in Appendix \ref{appdx-proof:cond2_Scion} and \ref{appdx-proof:cond2_uScion} for the first training step.
    In Appendix \ref{appdx-proof:cond1_2_induction}, we provide a proof that Propositions \ref{prop:des1} and \ref{prop:des2} hold for any training step $t$.

    \begin{proof}[Proof sketch of Proposition \ref{prop:des2} for \itScion]
    \label{proof_sketch_Cond1_Scion}
            Let $\ell \in \{1, \dots, L\}$.
            We have
            \begin{equation}
             \begin{aligned}
            \lVert \Delta_{W^{(\ell)}, Z^{(\ell)}}h^{(\ell)} \rVert_2^2 
            &= \lVert \Delta Z^{(\ell)}h_t^{(\ell - 1)} + L^{-\alpha} \Delta W^{(\ell)} \phi(h^{(\ell-1)}_{t}) \rVert_2^2 \\
            &= \Theta \left(
            \lVert \gamma_Z Z^{(\ell)}  h^{(\ell-1)}_{t} \rVert_2^2 
            +
            \lVert \gamma_Z \lmo(\nabla_{Z^{(\ell)}} \mathcal{L}) h^{(\ell-1)}_{t}
            \rVert_2^2 \right.\\
            & \quad \left. +
            \lVert L^{-\alpha} \gamma W^{(\ell)}  \phi(h^{(\ell-1)}_{t}) \rVert_2^2 
            +
            \lVert L^{-\alpha} \gamma \lmo(\nabla_{W^{(\ell)}} \mathcal{L}) \phi(h^{(\ell-1)}_{t}) \rVert_2^2 
            \right).
            \\
            \end{aligned}
            \end{equation}
            
            For the first term, using the alignment of the weights in \Eqref{eq:hyp_alignment} and the weights initialization in \Eqref{eq:hyp_init_and_update_weight}, we have
            \begin{equation}
                \begin{aligned}
                \lVert \gamma_Z Z^{(\ell)}  h^{(\ell-1)}_{t} \rVert_2^2 
                &= \Theta \left( \gamma_Z^2 \lVert Z^{(\ell)} \rVert_{\mathcal{S}_\infty}^2 \lVert h^{(\ell-1)}_{t} \rVert_2^2 \right)
                \\
                &=  \Theta \left( \gamma_Z^2 d_{\ell} \right) 
                \quad \text{from Desideratum \ref{des:1}.}
                \\
                \end{aligned}
            \end{equation}

            For the third term, using the alignment of the weights in \Eqref{eq:hyp_alignment} and the scale of the activation function $\phi$ in \Eqref{eq:hyp_scale_activation_function} as well as the weights initialization in \Eqref{eq:hyp_init_and_update_weight}, we have
            \begin{equation}
                \begin{aligned}
                \lVert L^{-\alpha} \gamma W^{(\ell)}  \phi(h^{(\ell-1)}_{t}) \rVert_2^2 
                &=  \Theta \left( L^{-2\alpha} \gamma^2 \lVert W^{(\ell)} \rVert_{\mathcal{S}_\infty}^2 \lVert h^{(\ell-1)}_{t} \rVert_2^2 \right)
                \\
                &=  \Theta \left( L^{-2\alpha} \gamma^2 d_{\ell} \right)
                \quad \text{from Desideratum \ref{des:1}.}
                \end{aligned}
            \end{equation}

            For the second and fourth terms, plugging spectral $\lmo$ expressions leads to
            \begin{equation*}
                \begin{aligned}
                &\lVert \gamma_Z \lmo(\nabla_{Z^{(\ell)}} \mathcal{L}) h^{(\ell-1)}_{t} \rVert_2^2 +  \lVert L^{-\alpha} \gamma \lmo(\nabla_{W^{(\ell)}} \mathcal{L}) \phi(h^{(\ell-1)}_{t}) \rVert_2^2 \\
                = \quad &  \Theta \left( \frac{d_{\ell}}{d_{\ell-1}} 
                \gamma_Z^2 \lVert h^{(\ell-1)}_{t} \rVert^2_2
                +
                L^{-2\alpha} 
                \frac{d_{\ell}}{d_{\ell-1}} 
                \gamma^2 \lVert \phi(h^{(\ell-1)}_{t}) \rVert^2_2 \right).
            \end{aligned}
            \end{equation*}

            Using the scale of the activation function assumption in \Eqref{eq:hyp_scale_activation_function}, and Desideratum \ref{des:1}, we have
            \begin{equation*}
                \begin{aligned}
                 & \lVert \gamma_Z \lmo(\nabla_{Z^{(\ell)}} \mathcal{L}) h^{(\ell-1)}_{t} \rVert_2^2 +  \lVert L^{-\alpha} \gamma \lmo(\nabla_{W^{(\ell)}} \mathcal{L}) \phi(h^{(\ell-1)}_{t}) \rVert_2^2 \\
                = \quad &  \Theta \left( 
                d_{\ell}( \gamma_Z^2 + L^{-2\alpha} \gamma^2 )
                \right).\\
                \end{aligned}
            \end{equation*}

            Finally, we have
             \begin{equation}
                \begin{aligned}
                 \lVert \Delta_{W^{(\ell)}, Z^{(\ell)}}h^{(\ell)} \rVert_2^2 = \Theta \left(d_{\ell}( \gamma_Z^2 + L^{-2\alpha} \gamma^2 ) \right).
                 \end{aligned}
            \end{equation}
            
            Therefore, to ensure that the entries of $\Delta h^{(\ell)}$ are $\Theta(1/L)$, we must set
            \begin{equation}
                \gamma = \gamma_{\text{base}} L^{\alpha-1}, \text{ and } \gamma_Z = {\gamma_Z}_{\text{base}}L^{-1}.
            \end{equation} 
    \end{proof}

    \textbf{Remark}: If we do not learn the $Z^{(\ell)}$, then we have that $Z^{(\ell)}_{t+1} = Z^{(\ell)}_t$, and $\Delta Z^{(\ell)}=0$, 
    and we conclude by only requiring $\gamma = \gamma_{\text{base}} L^{\alpha-1}$.

\begin{figure}[H]
        \minipage{0.5\textwidth}
            \includegraphics[width=\linewidth]{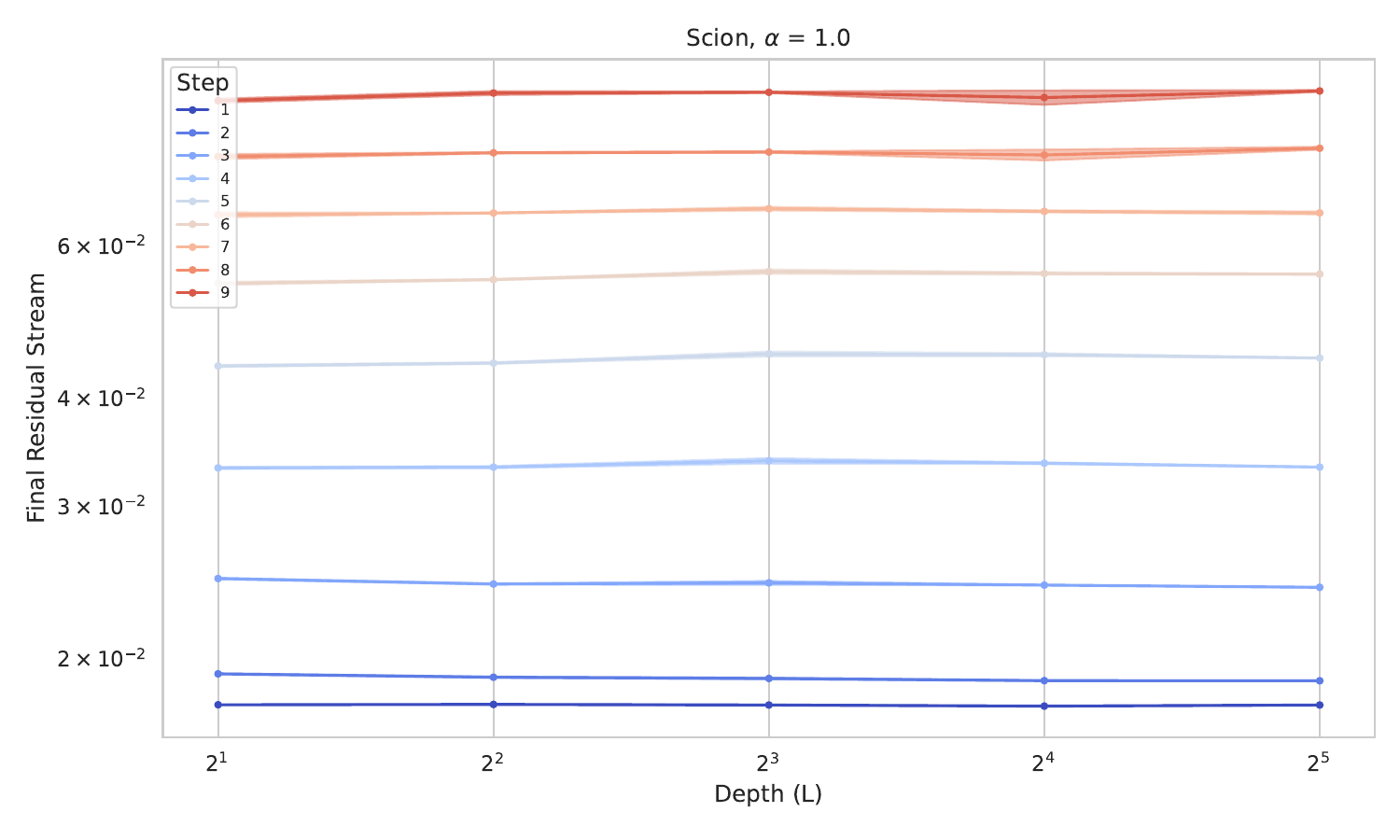}
        \endminipage\hfill
        \minipage{0.5\textwidth}
            \includegraphics[width=\linewidth]{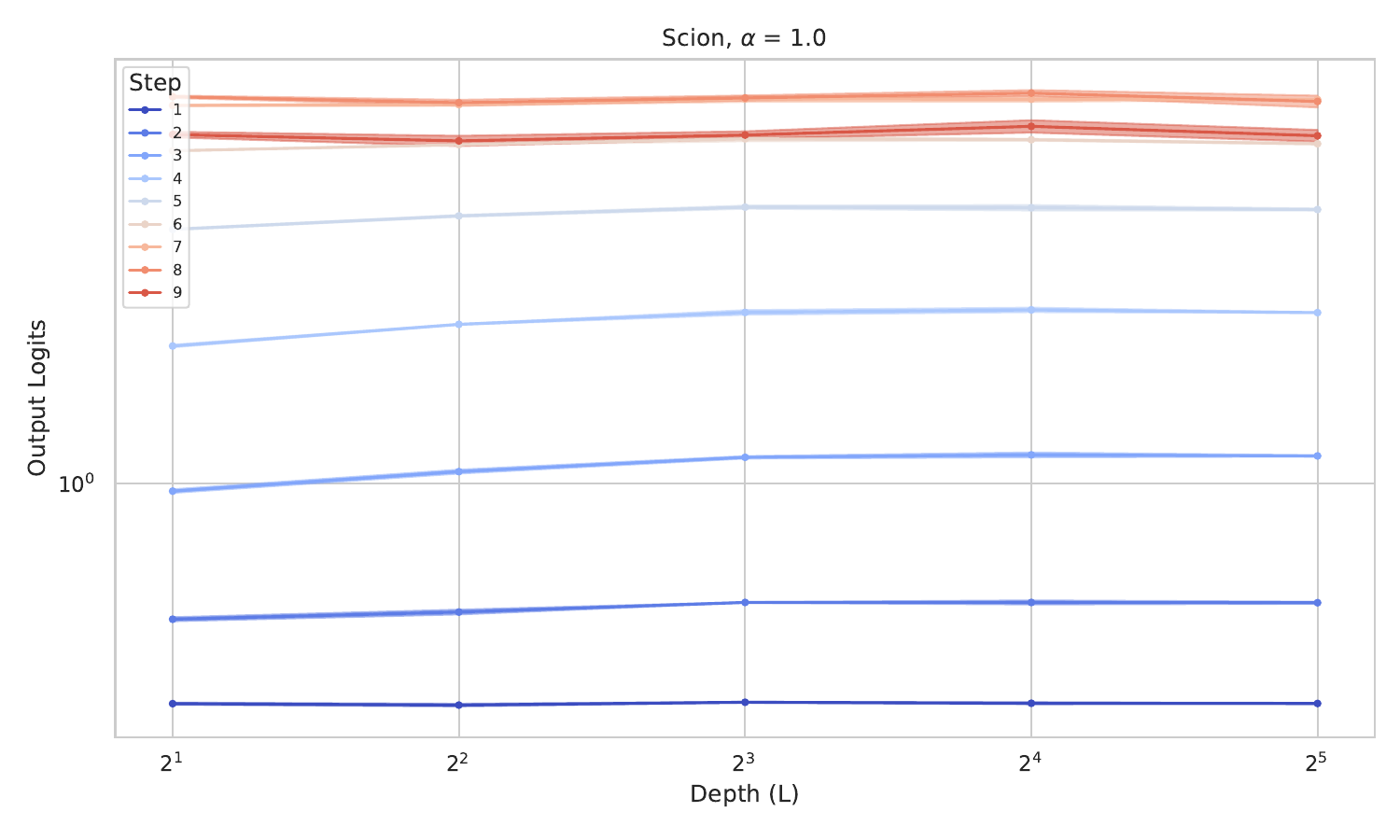}
        \endminipage
        \caption{Coordinate checks for \Scion, on nanoGPT trained on FineWeb, with $\alpha=1.0$, and learning rate $\gamma_{\text{base}}=2^{-11}$.
        We plot the change at the final residual representation $\lVert h^{(L)}_t - h^{(L)}_1 \rVert_2$ (left plot) and at the output logits $\lVert f_t - f_1 \rVert_2$ (right plot). More details on configuration can be found in Table \ref{tab:NanoGPT_hp} in Appendix \ref{appdx_sec:exp_details}.
         }
        \label{fig:nanogpt_coord_check_alpha1}
    \end{figure}

    In Figure \ref{fig:nanogpt_coord_check_alpha1}, we empirically verify the feature learning condition by performing a coordinate check \citep{yang2023TPIVFeatureLearningDepth}. As we scale the depth $L$, we measure the magnitude changes at the final residual representation $h^{(L)}$ and at the output logits $f$. We observe that these magnitudes remain stable across scales, confirming successful feature learning.

    \subsection{Complete Feature Learning}
    \label{sec:complete_feature_learning}
        This criterion is not necessary for hyperparameter transfer, but it distinguishes $\alpha=1$ from the other values in $[\nicefrac{1}{2}, 1]$ as the only value for which the dynamics stay non-linear in its parameter, for every scale. 
        
        To present this criterion, we need two definitions described below.

        \paragraph{Linearization}
        The linearization $h^{\text{lin}, \theta}$ of the function $h(\theta)$ with respect to $\theta$ about $\theta_0$ is
        \begin{equation}         
            \begin{aligned}
            h^{\text{lin}, \theta} (\theta, \theta_0) = h(\theta_0) + \langle \nabla_\theta h(\theta)|_{\theta_0}, \theta- \theta_0 \rangle.
            \end{aligned}         
        \end{equation}
        We say $h$ is \emph{linear} in $\theta$ if $h=h^{\text{lin}, \theta}$.

        \paragraph{Lazy-regime}
        We say a layer $\phi(h^{(\ell)})$ is \emph{lazy} with respect to a subset of parameters $\theta \subset \{\theta_j\}_{j<\ell}$ if at finite depth and width, $\phi(h^{(\ell)})$ is not linear in $\theta$ and the change $\Delta_\theta \phi(h^{(\ell)})$ at initialization from updating only $\theta$ (i.e. $\theta \mapsto \theta + \Delta \theta$)  is asymptotically the same as the change to the linearization of $h$:
        \begin{equation}
            \frac{|\Delta_\theta h^{(\ell)} - \Delta_\theta h^{\text{lin}, \theta}|}{|\Delta_\theta h^{\text{lin}, \theta}|} = o (1) \quad \text{as } d_\ell, L \rightarrow \infty. 
        \end{equation}
         
\begin{figure}[!ht]
        \includegraphics[width=\linewidth]{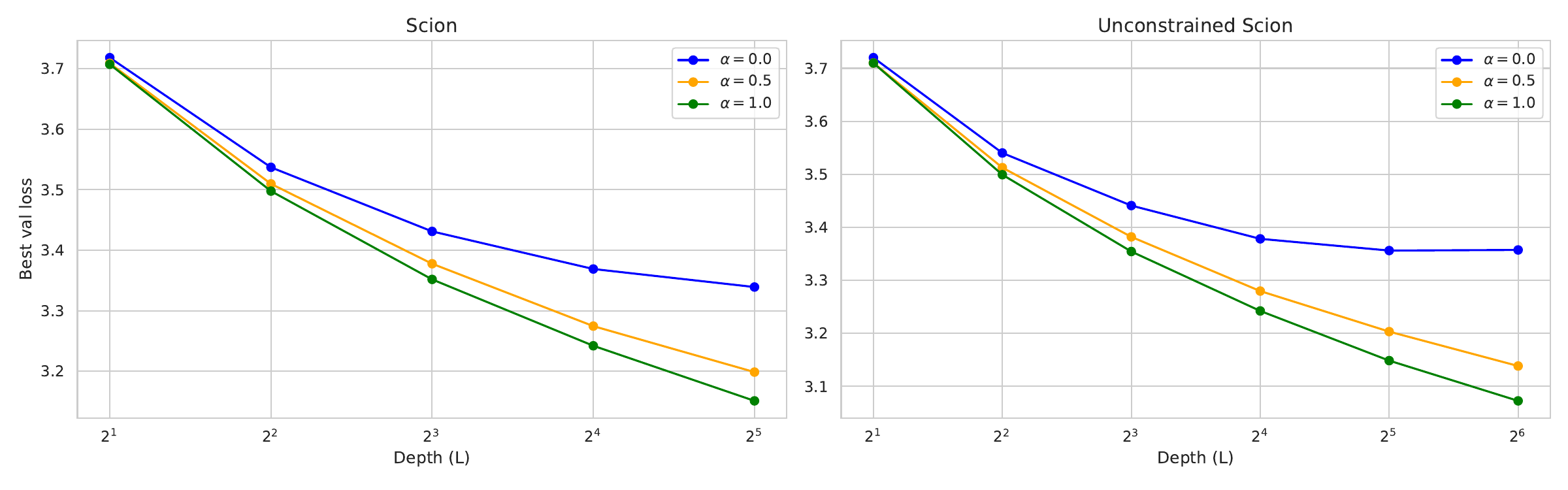}
        \caption{Best loss per learning rate in function of the depth $L$ for $\alpha \in \{0, \nicefrac{1}{2}, 1\}$, with \Scion and \uScion, on nanoGPT, trained on FineWeb, on $1$ seed. We observe that the choice $\alpha=1$ provides the best performances overall.
         Note that the case $\alpha=0$ is analogous to the SP/$\muP$ setup for Adam \citep{yang2021TPIVFeatureLeanringWidth,dey2026dontlazycompletepenables}.
        }
         \label{fig:best_loss_per_lr_gpt}
    \end{figure}

        Empirically, better generalization is observed when the network parametrization satisfies complete
        feature learning, i.e. when neither the hidden layers $\{\phi(h^{(\ell)})\}_{\ell \in \{1,\dots,L\}}$, nor the output model $f$ are lazy with respect to any subset of model parameters.
        We reproduce the experiment from \cite{dey2026dontlazycompletepenables} for the \Scion and \uScion optimizers in Figure~\ref{fig:best_loss_per_lr_gpt}, which reports the best validation loss achieved across learning rates as a function of depth $L$, for $\alpha \in \{0, \nicefrac{1}{2}, 1\}$. Consistently with the complete feature learning criterion, $\alpha = 1$ is the only value that yields systematically lower loss as depth increases, for both \Scion and \uScion.

    \subsubsection{Non-lazy regime for \bScion}
    \label{sec:nonlazy_for_scion}
    
    Another way to measure whether or not a parametrization results in the lazy regime is to measure the relative distance from initialization compared to the linearization.
    This gives a way to empirically verify  which values of $\alpha$ enable a training in a non-lazy regime.

\begin{figure}[!ht]
        \includegraphics[width=\linewidth]{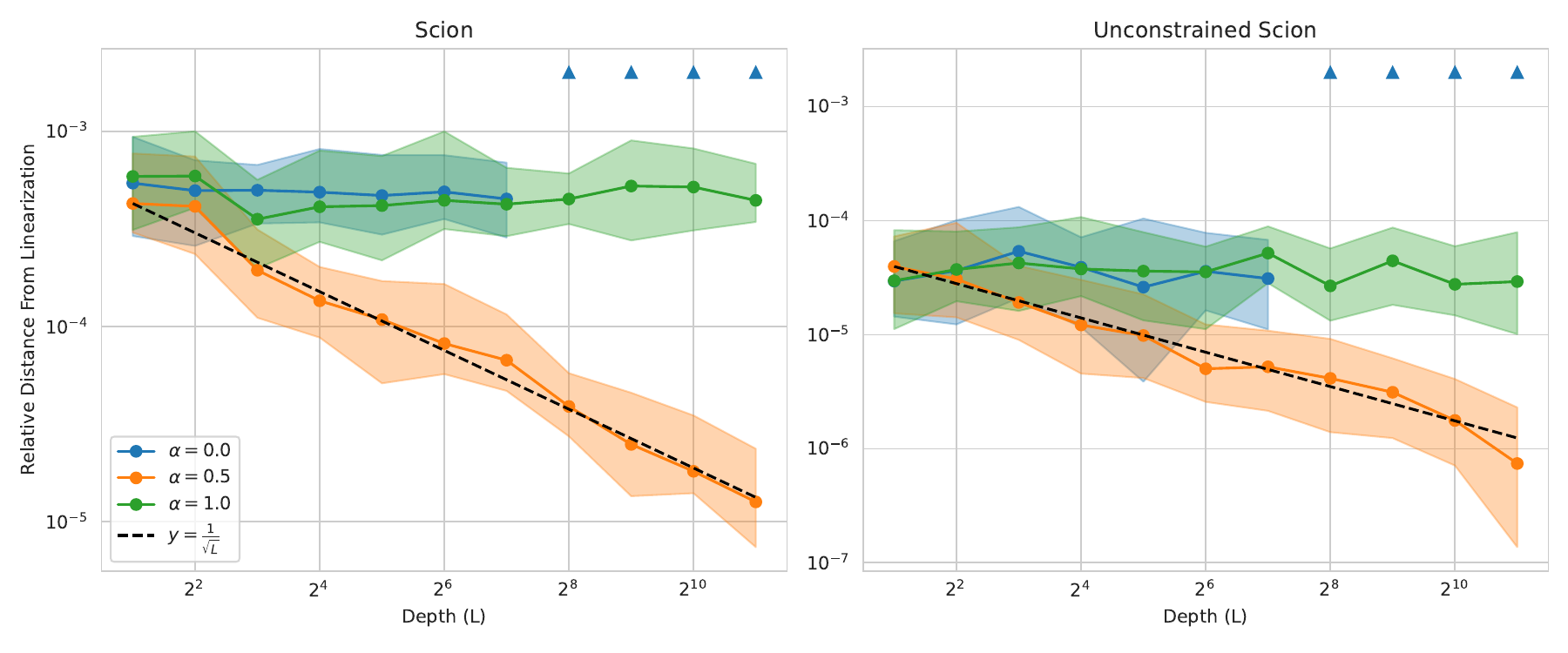}
        \caption{Relative distance from linearization for \Scion and \uScion on a 2-layer-per-block residual network, trained on MNIST, on $50$ seeds. The y-axis represents \Eqref{eq:linearization_nanogpt}. 
        The blue triangles represent the divergence of the case $\alpha=0$ at large depth.
        Only $\alpha=1$ results in complete and stable feature learning.
        $\alpha=\nicefrac{1}{2}$ approaches a linearized regime, at a rate of $\frac{1}{\sqrt{L}}$.
        Experimental details can be found in Table \ref{tab:resnet_hp}  in Appendix \ref{appdx_sec:exp_details}. }   
        \label{fig:relative_distance_linearization}
    \end{figure}

     We reproduce the experiment of \cite{dey2026dontlazycompletepenables} for \Scion and \uScion in Figure \ref{fig:relative_distance_linearization}, where we plot:
     \begin{equation}
     \label{eq:linearization_nanogpt}
            \frac{|\Delta_\theta h^{(\ell)} - \Delta_\theta h^{\text{lin}, \theta}|}{|\Delta_\theta h^{\text{lin}, \theta}|}
    \end{equation}
    for multiple depths and $\alpha \in \{0, \nicefrac{1}{2}, 1\}$, and we verify empirically that \Scion and \uScion are not lazy for $\alpha=1$ and $\alpha=0$.
    
\subsection{Experiments}
The architectures considered in the following experiments have no learnable $Z^{(\ell)}$ matrices, which are fixed to the identity. The hidden-layer width is constant, and training is restricted to the $W^{(\ell)}$ operators.
We additionally conducted experiments with the learnable linear operators $Z^{(\ell)}$ on the residual stream in Figure \ref{fig:moddednanogpt_with_Z} in Appendix \ref{sec:nanogpt_with_Z}.

\subsubsection{NanoGPT on FineWeb}
We ran experiments on nanoGPT that we adapt from the code in \cite{pethick2025trainingdeeplearningmodels}, which is based on modded-nanoGPT snapshot of \cite{modded_nanogpt_2024}, itself based on the original nanoGPT of \cite{karpathy2023nanogpt}. 

More experimental details can be found in Appendix \ref{appdx_sec:exp_details}.

The architecture employs transformer blocks to implement the residual function $\mathcal{F}_\ell (h^{(\ell - 1)} ; \theta_{\mathcal{F}}^{(\ell)})$ for all hidden layers $\ell \in \{1, \dots, L\}$. We perform orthogonalization for the query, key, value $(Q, K, V)$, and MLP weight matrices separately, rather than applying the transformation to their concatenation.

\begin{figure}[H]
        \includegraphics[width=\linewidth]{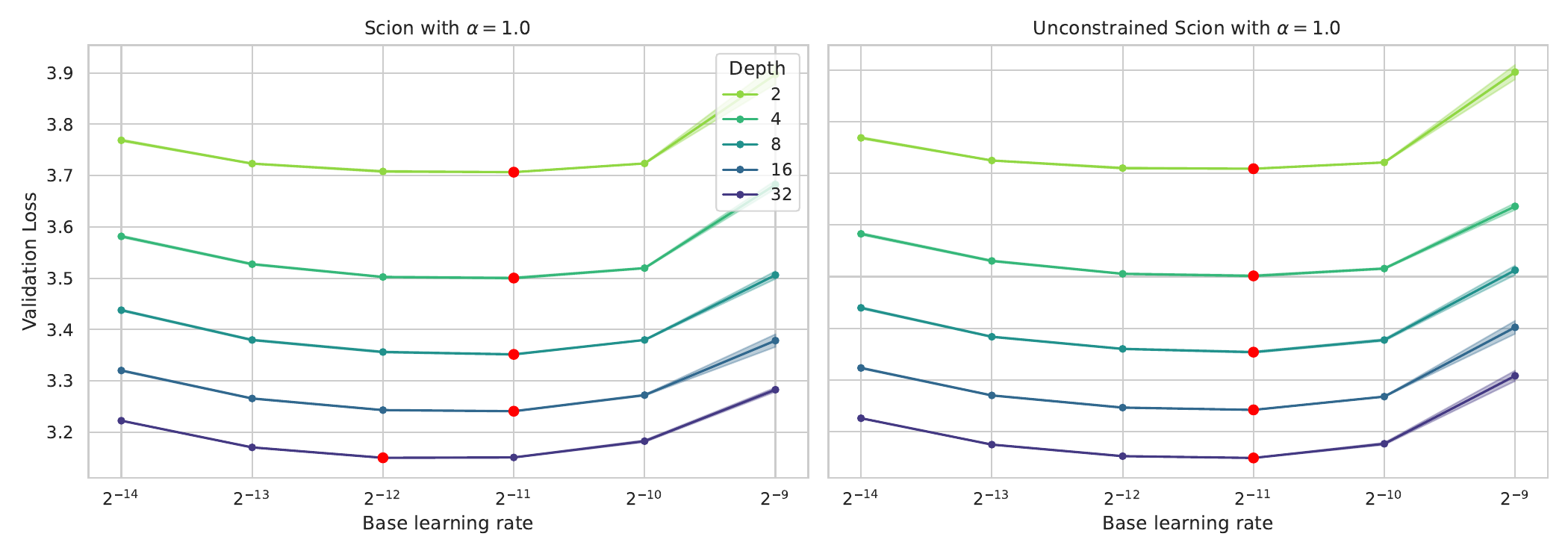}
        \caption{
        \Scion and \uScion on nanoGPT trained on FineWeb, with $\alpha=1$, on 3 seeds.
        Depth-aligned and parallel loss curves indicate that the optimal base learning rate remains invariant with depth. When the minimizers of the curves do not exactly coincide, the loss values remain very close in this flat region.
        For the \Scion case (left plot), the relative difference in loss value between base learning rate $2^{-12}$ and $2^{-11}$ for the deepest model is $0.024\%$.}
        \label{fig:nanogpt_alpha_1}
    \end{figure}

\begin{figure}[H]
       \includegraphics[width=\linewidth]{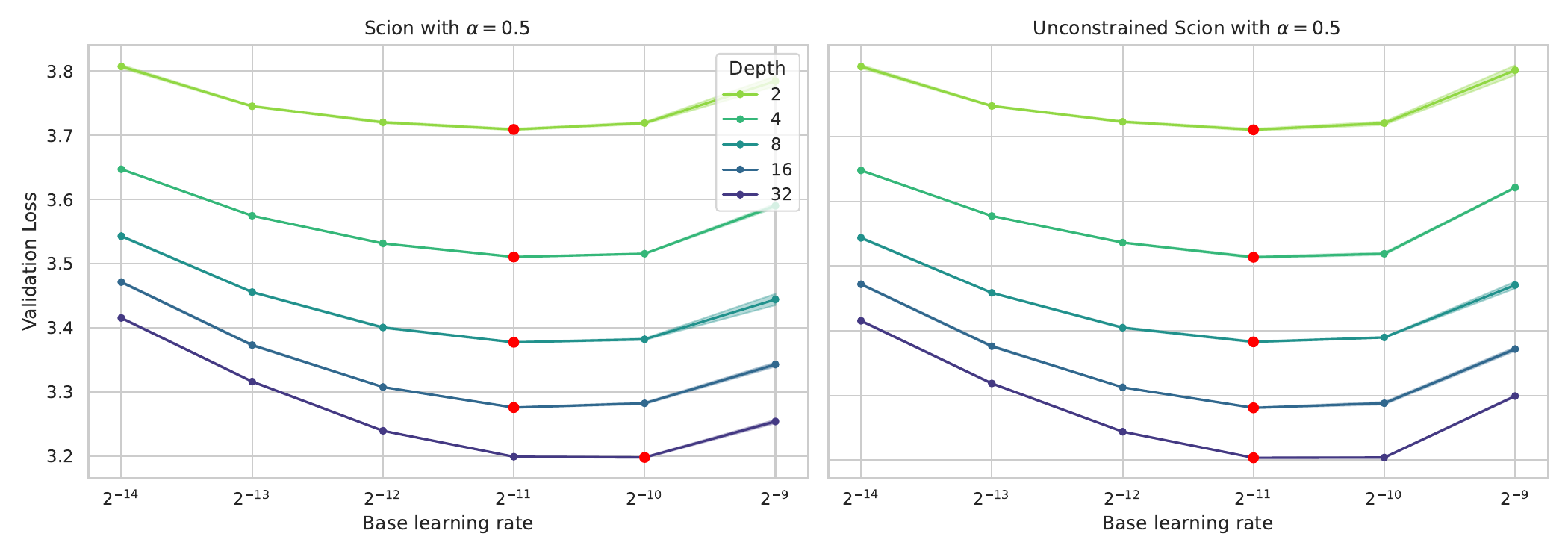}
        \caption{
        \Scion and \uScion on nanoGPT trained on FineWeb, with $\alpha=0.5$, on 3 seeds.
        The value $\alpha=0.5$ with the appropriate learning rate scaling also exhibits hyperparameter transfer with respect to the depth, as shown by the depth-aligned loss curves. For the \Scion case (left plot), the minimizers of the curves do not exactly coincide in the flat region, and the relative difference in loss value between base learning rate $2^{-11}$ and $2^{-10}$ for the deepest model is $0.037\%$.}
        \label{fig:nanogpt_alpha_05}
    \end{figure}

Figures \ref{fig:nanogpt_alpha_1} and \ref{fig:nanogpt_alpha_05} show the depth-wise hyperparameter transfer for \Scion and \uScion, with the choice $\alpha=1$ and $\alpha=\nicefrac{1}{2}$ respectively.

\subsubsection{MLP-Mixer on  CIFAR-10}
    We also used \Scion on the MLP-Mixer architecture for image classification tasks \citep{tolstikhin2021mlpmixerallmlparchitecturevision}, which offers the advantage of relying solely on MLP layers rather than convolutions, closely matching the modeling assumptions we made.
    Experimental details can be found in Appendix \ref{appdx_sec:exp_details}.

\begin{figure}[H]
        \centering
       \includegraphics[width=\linewidth]{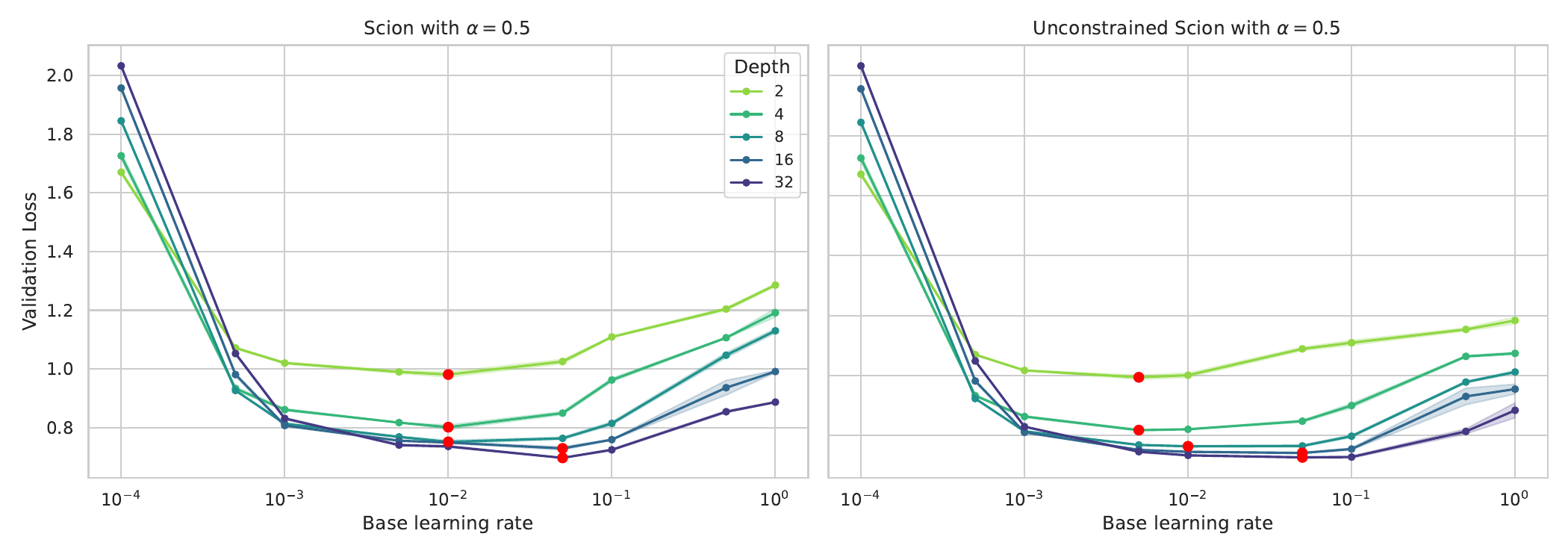}
        \caption{
        \Scion and \uScion on MLP-Mixer trained on CIFAR-10, with $\alpha=0.5$, on 3 seeds.
        The curves display different behaviors from the nearly parallel trends observed in Figures \ref{fig:nanogpt_alpha_1} and \ref{fig:nanogpt_alpha_05}. Despite these differences, the minimizers occur at similar locations, and the corresponding minimal values remain closely matched.
        For \Scion (left plot), the relative difference in loss values for base learning rates $10^{-2}$ and $5\times10^{-2}$ for the deepest model is approximately $6.3\%$. For \uScion (right plot), the relative difference is approximately $2.7\%$.
        }
         \label{fig:mlp_mixer_alpha05}
    \end{figure}

    Figures \ref{fig:mlp_mixer_alpha1} and \ref{fig:mlp_mixer_alpha05} show the depth-wise hyperparameter transfer for \Scion and \uScion, with the choice $\alpha=1$ and $\alpha=\nicefrac{1}{2}$ respectively.

    Again, we observe in Figures \ref{fig:mlp_mixer_alpha1} and \ref{fig:mlp_mixer_alpha05} that the minimizers of the curves are mostly aligned, and, when small deviations occur,  the minimal values remain very close. 

    \begin{figure}[H]
        \includegraphics[width=\linewidth]{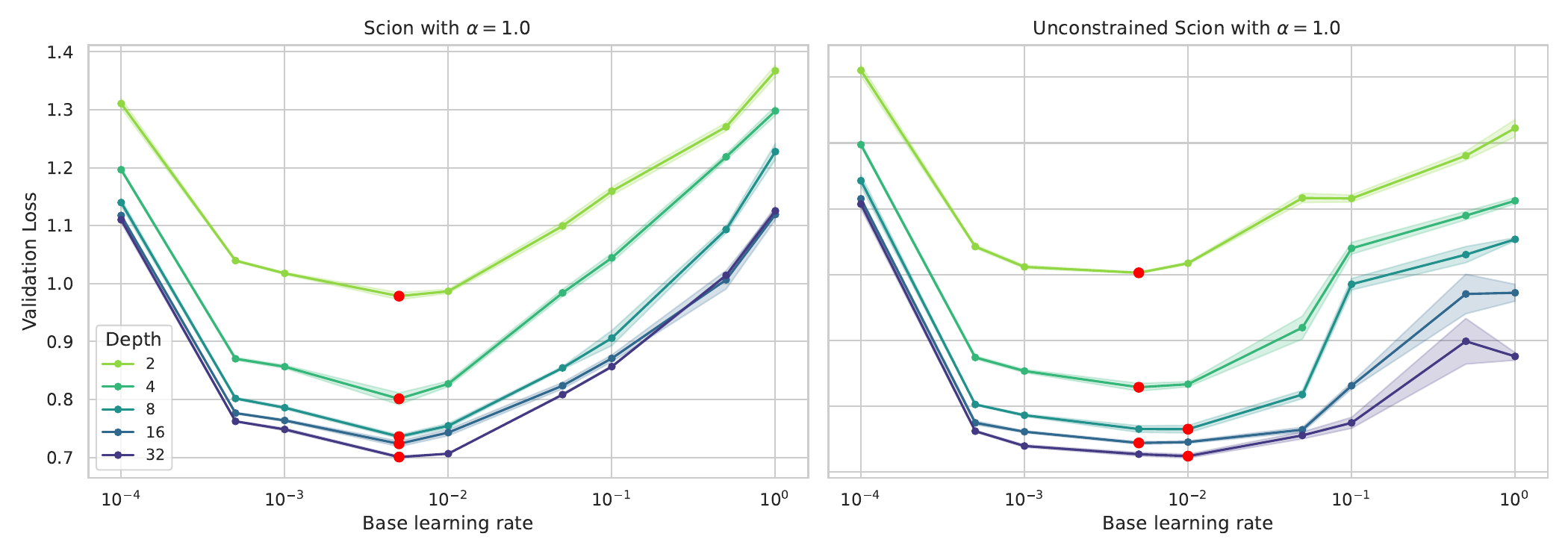}
        \caption{
        \Scion and \uScion on MLP-Mixer trained on CIFAR-10, with $\alpha=1$, on 3 seeds.
        The curves exhibit different shapes and are not perfectly parallel, in contrast to the behavior observed in Figures \ref{fig:nanogpt_alpha_1} and \ref{fig:nanogpt_alpha_05}. Nevertheless, the locations of the minimizers are largely aligned, and when deviations occur, the attained minimal values remain very close.
        For \uScion (right plot), the relative difference in loss values for base learning rates $5\times10^{-3}$ and $10^{-2}$ for the deepest model is approximately $0.62\%$.
        }
        \label{fig:mlp_mixer_alpha1}
    \end{figure}

\subsubsection{DeiT on ImageNet}
\begin{figure}[H]
    \centering
    \includegraphics[width=0.7\linewidth]{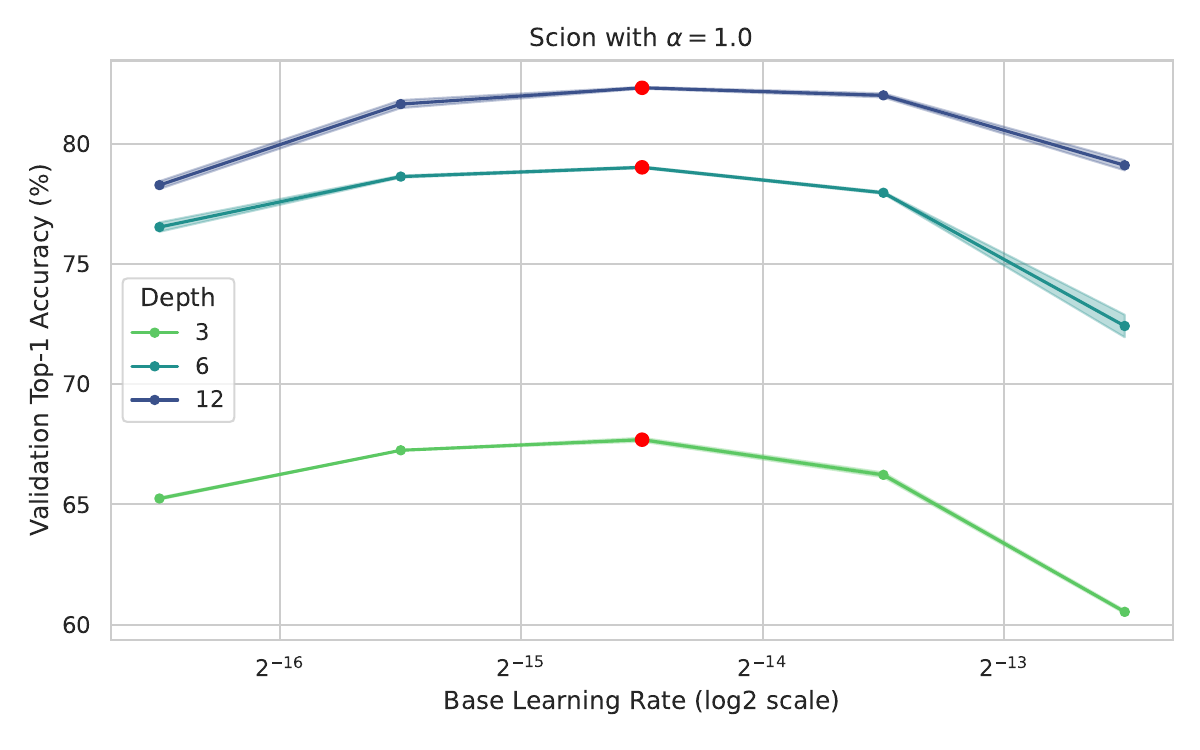}
    \caption{Scion  on DeiT trained on ImageNet, with $\alpha=1$, on $2$ seeds. The similar trends and aligned optima of the top‑1 accuracy curves across depths show that the optimal base learning rate is invariant with respect to depth. The best accuracy, achieved with depth $L=12$, is $82.3\%$.}
    \label{fig:deit_scion_alpha1}
\end{figure}

We trained a DeiT-base model adapted from \cite{pethick2025trainingdeeplearningmodels}, which builds upon the original DeiT codebase of \cite{pmlr-v139-touvron21a}. 
For reference, the 12-layer baseline architecture by \cite{pmlr-v139-touvron21a} achieves a top-1 test accuracy of $81.8\%$ on ImageNet when trained using AdamW for 300 epochs. With our training setup, we similarly achieve an accuracy of $82.3\%$ in 200 epochs, matching the reported results in \cite{pethick2025trainingdeeplearningmodels} (see Figure \ref{fig:deit_scion_alpha1}).

\section{Width-wise Transfer with Spectral Constraints}
\label{sec:width_wise_hp_transfer}

\subsection{Spectral conditions for feature learning}

For this section, we will simplify the setup and focus our derivations on MLPs. 
In particular, we desire that the features and the updates have $\Theta(1)$ element size with respect to the width.

\subsubsection{Problem setup}
    We consider an MLP with a fixed depth $L=\Theta(1)$, defined as
    \begin{equation}
        \begin{aligned}
        h^{(0)} &= W^{(0)} x \\
        h^{(\ell)} &= W^{(\ell)} \phi(h^{(\ell - 1)}) \quad (\forall \ell \in \{1, \dots, L\})\\
        f &= W^{(L+1)} \phi(h^{(L)}),
        \end{aligned}
    \end{equation}   
    where $x \in \mathbb{R}^{d_{\text{in}}}$ are the input data, $h^{(\ell)} \in \mathbb{R}^{d_\ell}, \ell \in \{0, \dots, L\}$ are the hidden states, and $f \in \mathbb{R}^{d_{\text{out}}}$ is the output after nonlinear activation $\phi$. The embedding and un-embedding matrices are respectively $W^{(0)} \in \mathbb{R}^{d_0 \times d_{\text{in}}}$ and $W^{(L+1)} \in \mathbb{R}^{d_{\text{out}} \times d_L}$. 

\subsubsection{Spectral Scaling Condition}
To ensure width-wise transfer, we adopt the conditions established by \cite{Yang2023ASpectralCondtion} for MLPs, which generalize to a broader range of architectures, including residual networks.

 \begin{restatable}[Feature Learning]
    {des}{RestatableDesFeatureLearning}
 \label{des:3}
    The hidden layers and their updates should be stable with the width:
    \begin{equation}
        \begin{aligned}
            \lVert h^{(\ell)} \rVert_2^2 = \Theta(d_{\ell}) \text{ and } \lVert \Delta h^{(\ell)} \rVert_2^2 = \Theta(d_\ell)  \quad \forall \ell=1, \dots, L.
        \end{aligned}
    \end{equation}
 \end{restatable}

Feature learning can be ensured by the spectral scaling condition on the weight matrices and their updates:
    \begin{equation}
        \begin{aligned}
            \lVert W^{(\ell)} \rVert_{S_\infty}^2 = \Theta\left(\frac{d_\ell}{d_{\ell-1}}\right) \text{ and } \lVert \Delta W^{(\ell)} \rVert_{S_\infty}^2 = \Theta\left(\frac{d_\ell}{d_{\ell-1}}\right) \quad \forall \ell=1, \dots, L.
        \end{aligned}
    \end{equation}

    \paragraph{Non-cancellation assumptions}
    \label{par:non_cancellation_assumptions_width_transfer_scion}
    We make the following assumptions on the alignment between updates and the current parameters. For any layer $\ell$, and training step $t$:
    \begin{enumerate}
        \item The current quantities and the $\lmo$ do not perfectly cancel
        \begin{equation}
        \label{eq:weight_lmo_spectral_cancellation_scion}
        \lVert
            -\gamma W^{(\ell)} + \gamma \lmo(M_t^{(\ell)})
            \rVert_{\mathcal{S}_\infty}^2
        = \Theta \left(
            \lVert
            \gamma W^{(\ell)}
            \rVert_{\mathcal{S}_\infty}^2
            +
            \lVert
            \gamma \lmo(M_t^{(\ell)})
            \rVert_{\mathcal{S}_\infty}^2
        \right).
        \end{equation}
        \item The different updates of the layer do not perfectly cancel.
         \begin{equation}
        \label{eq:alignment_sum_update_width_transfer}
        \lVert \Delta h^{(\ell)} \rVert_2^2 
        = \Theta \left( 
        \lVert \Delta W^{(\ell)} \phi(h^{(\ell-1)}) \lVert_2^2
        + \lVert W^{(\ell)} \Delta \phi(h^{(\ell-1)}) \lVert_2^2 
        + \lVert \Delta W^{(\ell)} \Delta \phi(h^{(\ell-1)}) \lVert_2^2 
         \right).
        \end{equation}
    \end{enumerate}
    These assumptions are empirically verified across widths and optimization steps in Figure \ref{fig:alignment_sum_update_with_scion_width}.
    
\subsection{Width Transfer for \bScion}
\label{sec:width_wise_hp_transfer_for_scion}

    We provide an analysis of the width-transfer property of \Scion, which was only demonstrated for \uScion in \cite{pethick2025trainingdeeplearningmodels}.     

    \begin{restatable}[\Scion and \uScion verify Desideratum \ref{des:3}]{proposition}{ScionWidth}
    \label{prop:width_scion}
     Under the assumptions defined in the subsections \ref{sec:assumptions_tools} and \ref{par:non_cancellation_assumptions_width_transfer_scion}, the updates of the features after a step of \Scion or \uScion, remain stable when the width $d_\ell$ becomes increasingly large. 
    \end{restatable}
    
\begin{figure}[!ht]
    \centering
        \includegraphics[width=\linewidth]{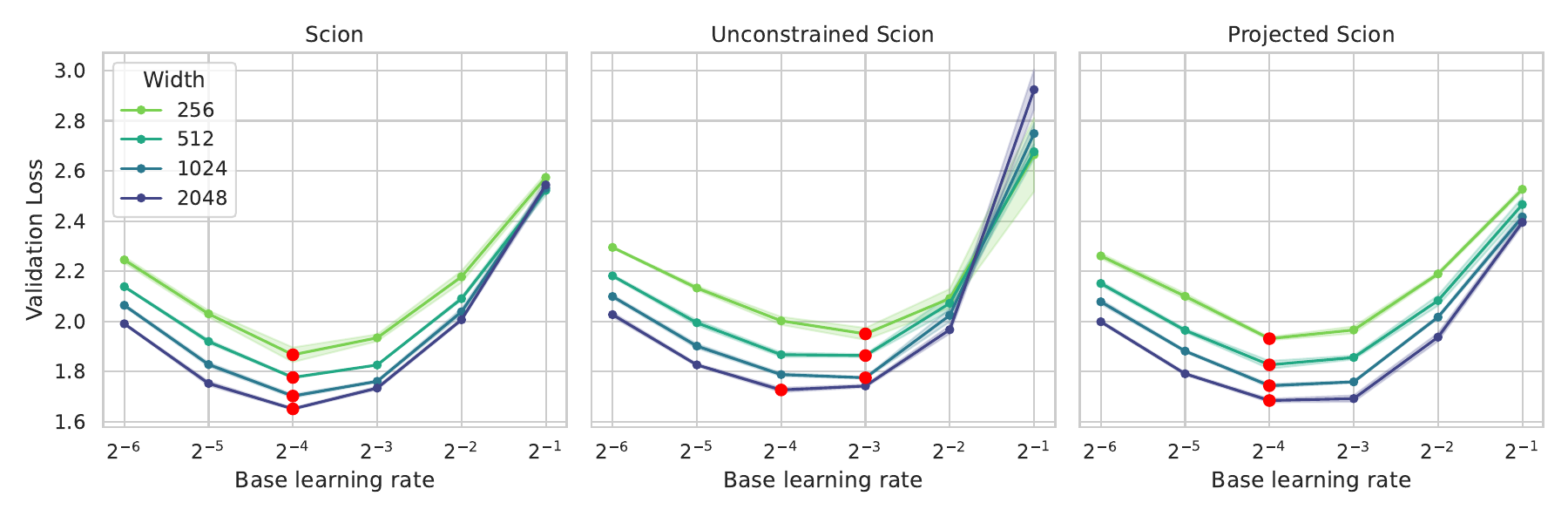}
    \caption{Width Transfer with \Scion, \uScion and \pScion on Shallow nanoGPT trained on Shakespeare-Char dataset, with depth 3, on 3 seeds.
    We observe that the minimizers of the curves, highlighted by the red dots, are generally well aligned. In the case where they differ slightly, the minimal values are nearly identical, indicating a flat region around the optimum.
    For \uScion (middle plot), the relative difference in loss values for base learning rates $2^{-4}$ and $2^{-3}$ for the largest model is approximately $2.9\%$.
    Experimental details can be found in Table \ref{tab:ShallowGPT_hp} in Appendix \ref{appdx_sec:exp_details}.
    }
    \label{fig:width_transfer_shakespeare_char}
    \end{figure}

    A demonstration of Proposition \ref{prop:width_scion} is available in Appendix \ref{appdx-proof:width_transfer_scion}, and an illustration of the width-wise transfer of the learning rate is given in Figure \ref{fig:width_transfer_shakespeare_char}.

\subsection{\pbScion}
\label{sec:projected_scion}
    In this section, we consider an algorithm that enforces the constraints through projections, by directly clipping the singular values of the weight matrices to lie within a certain range. Doing so provides control of the singular values of the weights and their updates, which can help stabilize training \citep{newhouse2025training}. We provide the definition of spectral clipping below.

    \begin{definition}[Spectral Clipping]
    \label{def:spectralclipping}
        Let $W\in \R^{d \times n}$, and $W=U \Sigma V^\top$ be its singular value decomposition, where $\Sigma=\text{diag}(\sigma_1, \dots, \sigma_{min(d,n)})$ is the diagonal matrix composed of the singular values of $W$, and $U, V$ are orthogonal matrices. Let $a,b \in \R_+$ with $a<b$. The spectral clipping $\spectralclip_{[a, b]}: \mathbb{R}^{d \times n} \rightarrow \mathbb{R}^{d \times n}$ is defined as:
        \begin{equation}
            \spectralclip_{[a, b]} (W) = U  \clip_{[a,b]}(\Sigma)V^\top
        \end{equation}
        where $\clip_{[a,b]}(\Sigma)$ is the projection mapping of the singular values on $[a,b]$.
    \end{definition}

    Since computing the SVD is expensive in practice, especially for deep learning models with large weight matrices, we can reconstruct the clipping functions using the matrix sign function \citep{cesista2025spectralclipping} which can be approximated via the Newton-Schulz iterations \citep{kovarik1970some,bjorck1971iterative}. 

    We can therefore study a variant of \Scion, given in Algorithm \ref{alg:projected_SCG}, based on a projected version of SCG, where the update can be written as a composition of the $\spectralclip$ function and the uSCG update.

    \begin{figure}[ht!]
    \centering
        \begin{minipage}{0.55\linewidth}
        \begin{algorithm}[H]
        \caption{Projected SCG}
        \label{alg:projected_SCG}
        \textbf{Input:} Horizon $n$, initialization $\mathbf{\Theta}_1 \in \mathcal D$, $M_0 = 0$, 
        momentum $\beta_t\in [0,1)$, spectral constraints $a$ and $b$ with $a<b$, and stepsize $\gamma_t \in (0,1)$
        \begin{algorithmic}[1]
            \For{$t = 1, \dots, n$}
                \State Sample data $x \sim \mathcal P$
                \For{$\theta^{(\ell)} \in \mathbf{\Theta}$}
                \State $M^{(\ell)}_{t} \gets (1-\beta_{t}) \nabla_{\theta^{(\ell)}_t} \mathcal{L} + \beta_t M^{(\ell)}_{t-1}$
                \State $\theta^{(\ell)}_{t+1} \gets \spectralclip_{[a,b]} (\theta^{(\ell)}_t + \gamma_t \lmo(M^{(\ell)}_t))$
                \EndFor
            \EndFor
        \end{algorithmic}
        \end{algorithm}
    \end{minipage}
    \end{figure}

    We also observe Algorithm \ref{alg:projected_SCG} to exhibit width-wise hyperparameter-transfer in experiments, as detailed in the next subsection. 

\subsubsection{Width transfer for \pbScion}
\label{sec:width_wise_hp_transfer_for_projected_scion}
   
\pScion makes it possible to establish an upper bound on the size of the feature updates that is independent of width. This provides a theoretical
stability guarantee.

\begin{restatable}[Width-wise stability of \pScion]
{proposition}{ProjectedScionWidth}
\label{prop:width_projected_scion}
Under the assumptions defined in Sections \ref{sec:assumptions_tools} and \ref{par:non_cancellation_assumptions_width_transfer_scion}, the feature updates after a step of \pScion admit an upper bound that is independent of the width $d_\ell$. In particular, for each layer $\ell$,
\begin{equation}
    \lVert \Delta h^{(\ell)} \rVert_2^2 = \mathcal{O}(d_\ell)
\end{equation} 
with respect to the width $d_\ell$.
\end{restatable}

Proposition \ref{prop:width_projected_scion} establishes the stability part
of the width-wise scaling but it does not establish the matching lower bound to get $\lVert \Delta h^{(\ell)} \rVert_2^2=\Theta(d_\ell)$. For the methods without projection considered in Section \ref{sec:width_wise_hp_transfer_for_scion}, the lower bound was obtainable by ruling out cancellations between the current parameters and the update. For \pScion, the projection introduces an additional nonlinearity that can result in systematic cancellation for small $\gamma$; details are provided in Appendix \ref{appdx_proof:width_transfer_projected_scion_interval} which includes an experiment that confirms this cancellation happening. Establishing a matching lower bound for general spectral clipping therefore remains an open point in our analysis.

Figure \ref{fig:width_transfer_shakespeare_char} illustrates the empirically observed width-wise transfer property of \pScion. 
    
\section{Radius Transfer}
\label{sec:radius_transfer}

    Besides the learning rate, a remaining challenge lies in tuning the \emph{scales} $\{\rho^{(\ell)}\}_{\ell\in\{0, \dots, L+1\}}$ of \Scion and \uScion.
    These radii appear explicitly in the $\lmo$ expressions scaling the layer-wise magnitude of the updates, and therefore play a crucial role in the optimization process.
    
    In this section, we provide preliminary empirical evidence for radius transfer for \Scion and \uScion across both width and depth.
    We formalize the radius transfer pipeline into the reproducible Protocol below.

    \begin{tcolorbox}[colback=BlueGreen!5,colframe=BlueGreen!50,title=\textbf{Protocol: Practical Learning Rate and Radius Transfer}]
    \label{protocol}
    \begin{enumerate}
        \item \textbf{Apply Scaling Rules:} Apply the parameter and learning rate scaling rules defined in Section \ref{sec:complete_scion} and parametrize the radii. For instance, for an \textit{(embedding $\rightarrow$ hidden $\rightarrow$ un-embedding)} architecture with tied embedding and un-embedding weights, parameterize the layer-wise radii configuration as $(a,1,a)$, using a shared scalar factor $a>0$.
        \item \textbf{Low-Dimensional Sweep:} Define a small, computationally inexpensive version of the model (i.e. with small dimensions) and run a hyperparameter sweep to find the optimal learning rate and optimal $a$: $(\gamma^*, a^*)$.
        \item \textbf{Target Scale Deployment:} Train the large-scale target model directly using the optimal $(\gamma^*, a^*)$, bypassing expensive tuning at scale.
    \end{enumerate}
    \end{tcolorbox}

    We emphasize that this section is empirical and should be interpreted as an initial exploration of the phenomenon and of the practical protocol provided.

    \subsection{Radius and Depth Transfer}
    \cite{large2024scalable} define the modular norm recursively based on the model's architecture, which in particular allows normalization within this norm using a user-specified mass. Under this modular norm framework, the authors explore mass allocation and show that both the optimal mass ratio between hidden and extreme layers and the optimal learning rate transfer from small model to larger ones, both in depth and width, when training with Adam.

\begin{figure*}[!ht]
    \centering
    \begin{subfigure}[b]{0.505\linewidth}
        \centering
        \includegraphics[width=\linewidth]{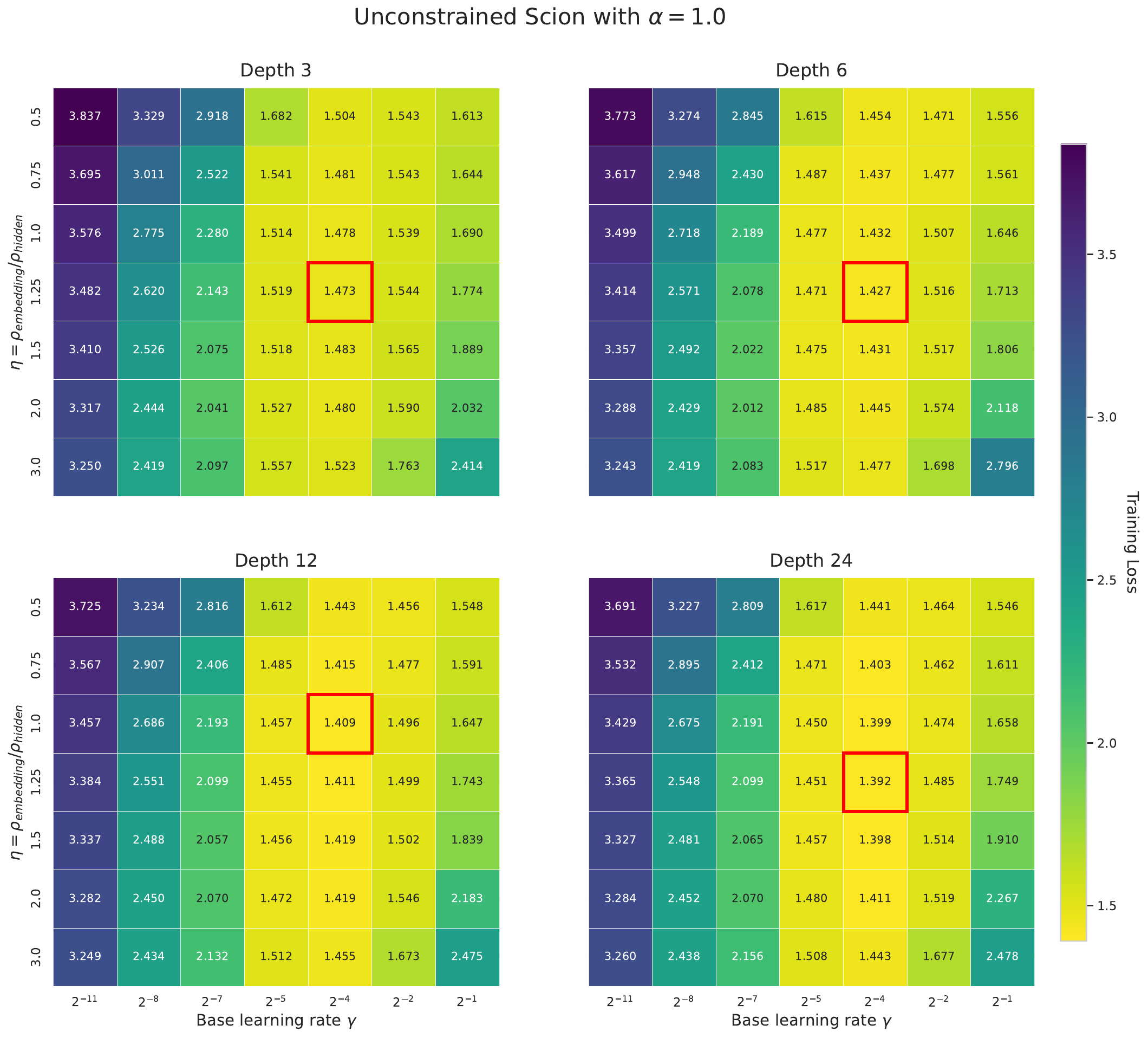}
        \caption{Shakespeare-Char dataset.}
        \label{fig:radius_transfer_shakespeare}
    \end{subfigure}
    \hfill
    \begin{subfigure}[b]{0.485\linewidth}
        \centering
        \includegraphics[width=\linewidth]{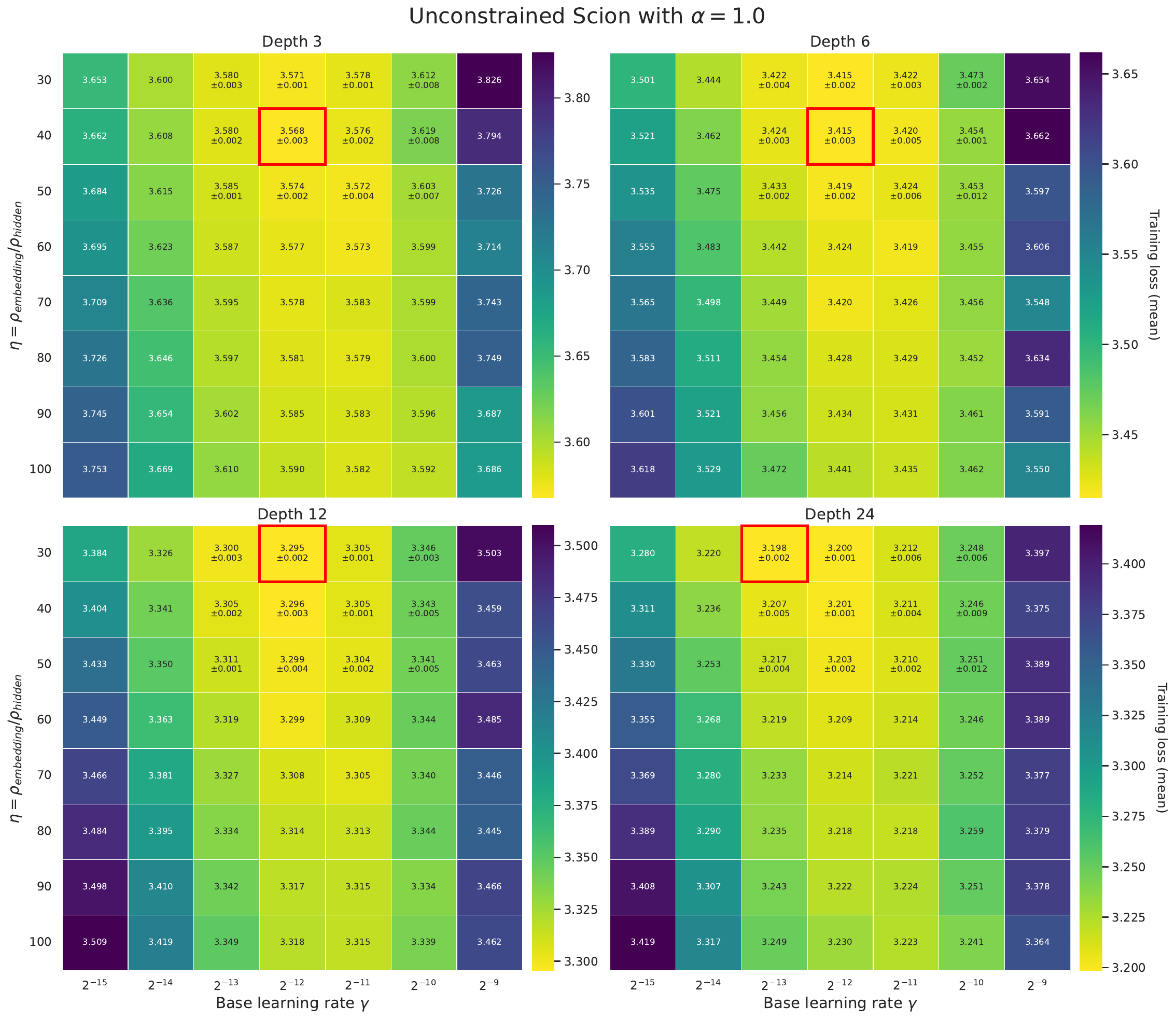}
        \caption{FineWeb dataset.}
        \label{fig:radius_transfer_fineweb}
    \end{subfigure}
    \caption{Preliminary empirical evidence for transfer of the ratio of the radii with \uScion\ ($\alpha=1$). The optimal combination of radii ratio and base learning rate is nearly identical across depths or lies within a flat region of the loss landscape, suggesting that this landscape remains invariant with respect to depth in the setting considered. 
    \textit{Left} (a): Trained on Shallow nanoGPT, for 1 seed. For depth $12$, the relative difference in loss between the cell $(\eta,\text{base learning rate})=(1.25, 2^{-4})$ and the cell $(1.0 , 2^{-4})$ is $0.1\%$. Experimental details are in Table \ref{tab:ShallowGPT_hp} in Appendix \ref{appdx_sec:exp_details}. 
    \textit{Right} (b): Trained on nanoGPT. The cells with std values were trained on 3 seeds. For depth $24$, the relative difference in loss between the cell $(\eta,\text{base learning rate})=(40, 2^{-12})$ and the cell $(30, 2^{-13})$ is $0.06\%$. Experimental details are in Table \ref{tab:NanoGPT_hp} in Appendix \ref{appdx_sec:exp_details}.}
    \label{fig:radius_transfer_combined}
\end{figure*}

    We conducted a similar experiment on Shallow nanoGPT on the Shakespeare-Char dataset, and on nanoGPT on FineWeb.
    Details of the experiments can be found in Table \ref{tab:ShallowGPT_hp} and Table \ref{tab:NanoGPT_hp} respectively, in Appendix \ref{appdx_sec:exp_details}.
    We trained multiple models of several depths, learning rates and ratios of radii $\eta=\frac{\rho_{\text{embedding}}}{\rho_{\text{hidden}}}$ where $\rho_{\text{embedding}}$ is the radius of the embedding and un-embedding layers and $\rho_{\text{hidden}}$ is the radius of the hidden layers. 
    The results are displayed in Figure \ref{fig:radius_transfer_shakespeare} and Figure \ref{fig:radius_transfer_fineweb}.

    Taken together, these experiments provide encouraging preliminary empirical evidence that the transfer of both the optimal learning rate and radii may occur. See also \cite[Section 3.4]{filatov2026optimalscalingneedsoptimal} for a similar experiment showing radius transfer across training horizons.

\subsection{Radius and Width \& Depth Transfer}
\begin{figure}[!ht]
    \centering
    \includegraphics[width=\linewidth]{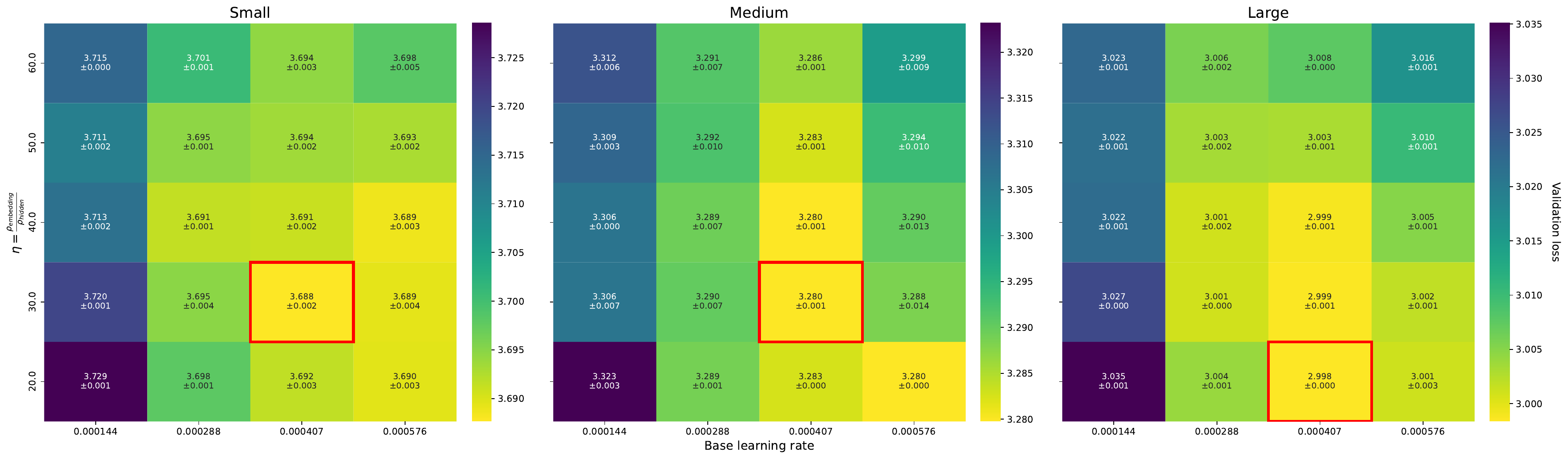}
    \caption{
    Preliminary empirical evidence for
    width and depth transfer with Scion, on nanoGPT, trained on FineWeb, with $\alpha=1$, on $3$ seeds. The Small model has depth 6 and width 384, the Medium one has depth 12 and width 768 and the Large one has depth 24 and width 1536.
    For the Large model, the relative difference in loss between the cell $(\eta,\text{base learning rate})=(30, 0.000407)$ and the cell $(20, 0.000407)$
    is $0.03\%$.
    Experimental details are in Table \ref{tab:NanoGPT_hp} in Appendix \ref{appdx_sec:exp_details}.
    }
    \label{fig:width_and_depth_transfer}
\end{figure}

In this section, we provide preliminary empirical evidence on the learning rate transfer of \Scion across both the width and the depth of the model.
This observation is empirical rather than theoretically established in our framework, since the width-and-depth-transfer analyses are derived under different architectural assumptions.
Figure \ref{fig:width_and_depth_transfer} illustrates this empirical behavior.

The three models were trained for $5,100$ iterations. According to the Chinchilla scaling laws \citep{hoffmann2022training}, a compute-optimal training run requires approximately $20$ tokens per parameter. Given our training configuration, where we consider a batch size of $512$ and a sequence length of $1024$, each training step processes $512 \times 1024 = 524,288$ tokens. 

%% file: sec-conclusion.tex
\section{Conclusion}

In this work, we investigated the hyperparameter-transfer behavior of \Scion and its unconstrained variant, and we demonstrated that this optimizer supports both width and depth-wise transfer. Consistent with the findings of CompleteP \citep{dey2026dontlazycompletepenables}, we observed that the setting $\alpha=1$ yields the strongest performance and best transfer. 

Our main contribution is to establish these transfer phenomena for the spectral  optimizer \Scion, in both its constrained and unconstrained variants, extending prior analyses of hyperparameter transfer. 

Through a diverse set of experiments, we further showed that \Scion can be parameterized to guarantee feature learning, operation in the non-lazy regime, and movement from initialization at the maximum stable learning rate. 

Our experiments covered several deep neural network architectures and datasets, including nanoGPT trained on FineWeb, DeiT on ImageNet, and MLP-Mixer on CIFAR-10.

We finish by remarking that these results can also apply to Muon, provided that the components relying on Adam are scaled appropriately so that it remains compatible with the transfer regime.

\subsection{Perspectives}
Given the promising preliminary results observed in Section \ref{sec:radius_transfer}, it would be interesting to theoretically investigate the radius transfer property and derive a rigorous proof. 
Leveraging this hyperparameter property could prove highly beneficial for several downstream tasks. In particular, transferring the radius of a constrained set has direct implications for improving robustness against adversarial attacks \citep{goodfellow2015explainingharnessingadversarialexamples,sokolicrobustlargemargin,miyato2018spectralnormalizationgenerativeadversarial}, generalization bounds \citep{GenBound17}, or control on the Lipschitz continuity of neural networks \citep{large2024scalable,miyato2018spectralnormalizationgenerativeadversarial}. Extending to include other parametrizations of weight decay than the Frank-Wolfe step remains an interesting direction too \citep{chou2025correction}.

%% file: sec-appendix.tex
\section{Additional figures}
\subsection{Additional plots from the NanoGPT experiments}    
\begin{figure}[!ht]
    \centering
        \includegraphics[width=0.7\linewidth]{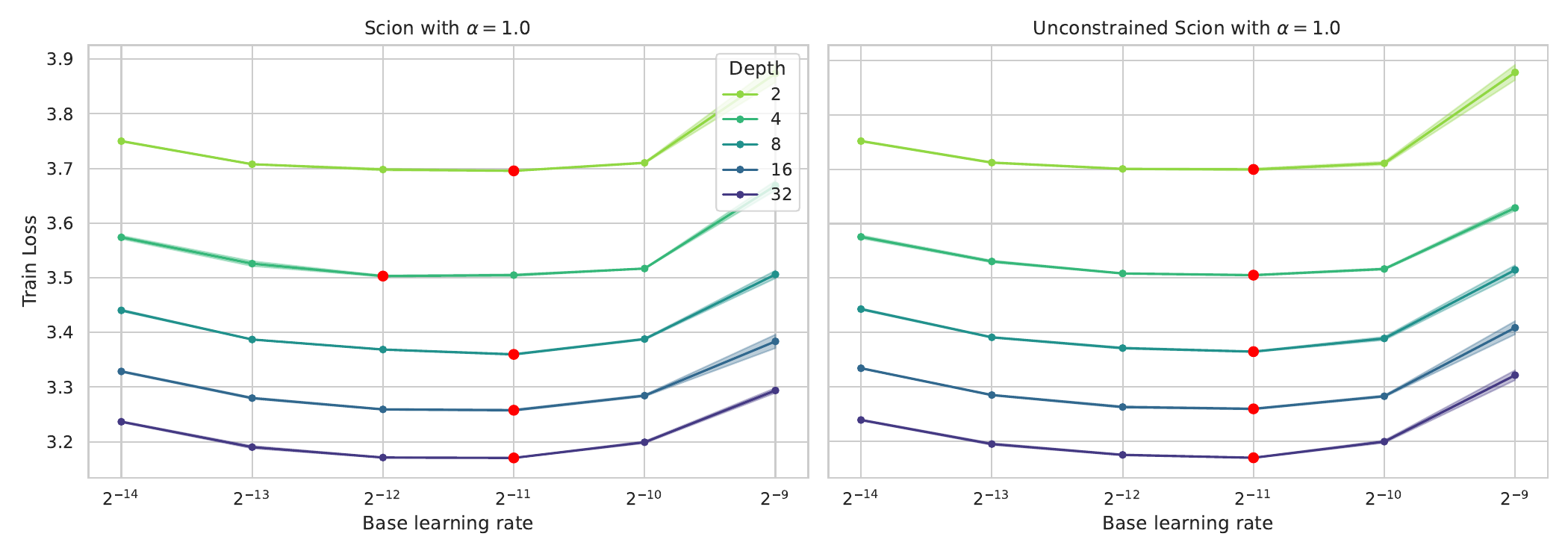}
        \includegraphics[width=0.7\linewidth]{moddednanogpt_finewebtext/GPT_fineweb_outdepth_transfer_scion_and_uscion_alpha_1.0_val_loss_5100_epochs_3_seeds.pdf}
        \includegraphics[width=0.7\linewidth]{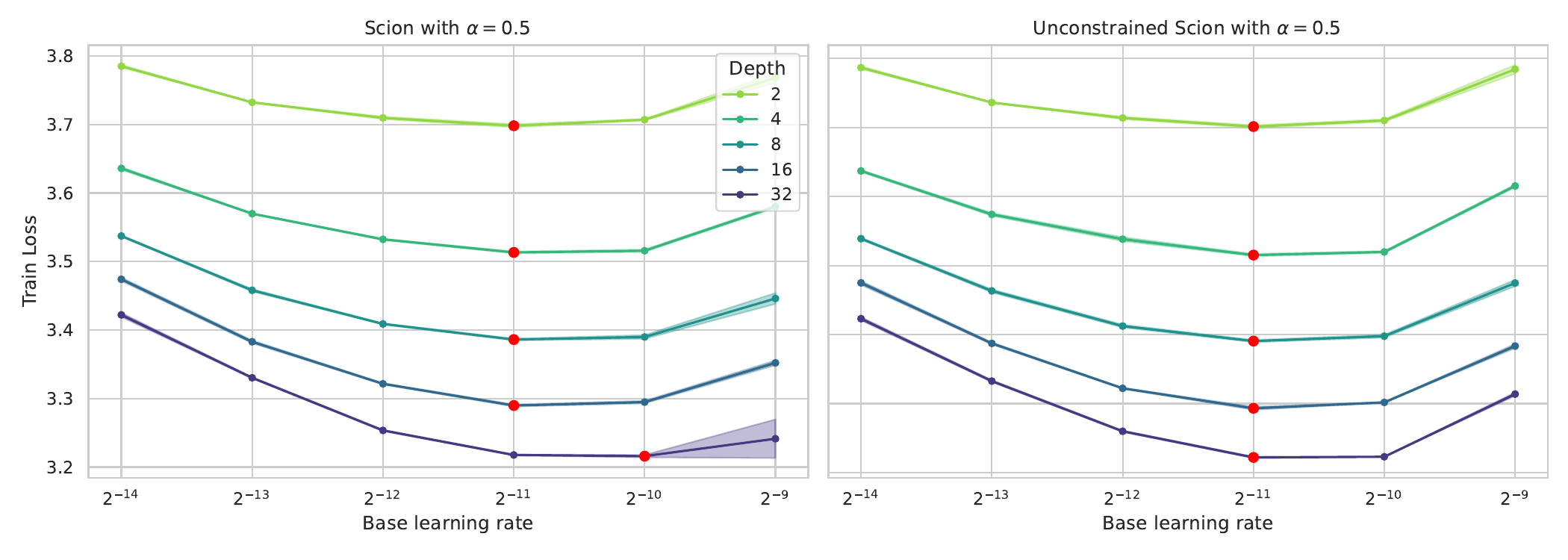}
        \includegraphics[width=0.7\linewidth]{moddednanogpt_finewebtext/GPT_fineweb_outdepth_transfer_scion_and_uscion_alpha_0.5_val_loss_5100_epochs_3_seeds.pdf}

    \caption{
    \Scion (left column) and \uScion (right column) on nanoGPT trained on FineWeb, on 3 seeds.
    The top rows correspond to $\alpha = 1.0$ and the bottom rows to $\alpha = 0.5$.
    The minimizers of the different curves, indicated by red dots, are strongly aligned across depths.
    When they do not exactly coincide, the corresponding loss values remain very close, indicating a flat loss region around the optimum.
    The validation loss relative errors are reported in Figures \ref{fig:nanogpt_alpha_1} and \ref{fig:nanogpt_alpha_05}. The training loss relative error with \Scion is approximately $0.050\%$ with $\alpha=1$ and depth $L=4$, and $0.052\%$ with $\alpha=0.5$ and depth $L=32$.
    }
    \label{appdx_fig:nanogpt_scion_depth_transfer_alpha_1_and_05}
    \end{figure}

    We trained multiple models where we vary the depth and the learning rate. Figure \ref{appdx_fig:nanogpt_scion_depth_transfer_alpha_1_and_05} illustrates the depth-wise hyperparameter transfer with $\alpha=1.0$ and $\alpha=0.5$, for \Scion and \uScion.

\subsection{Additional plots from the MLP-Mixer experiments}
\begin{figure}[!ht]
    \centering
        \includegraphics[width=0.7\linewidth]{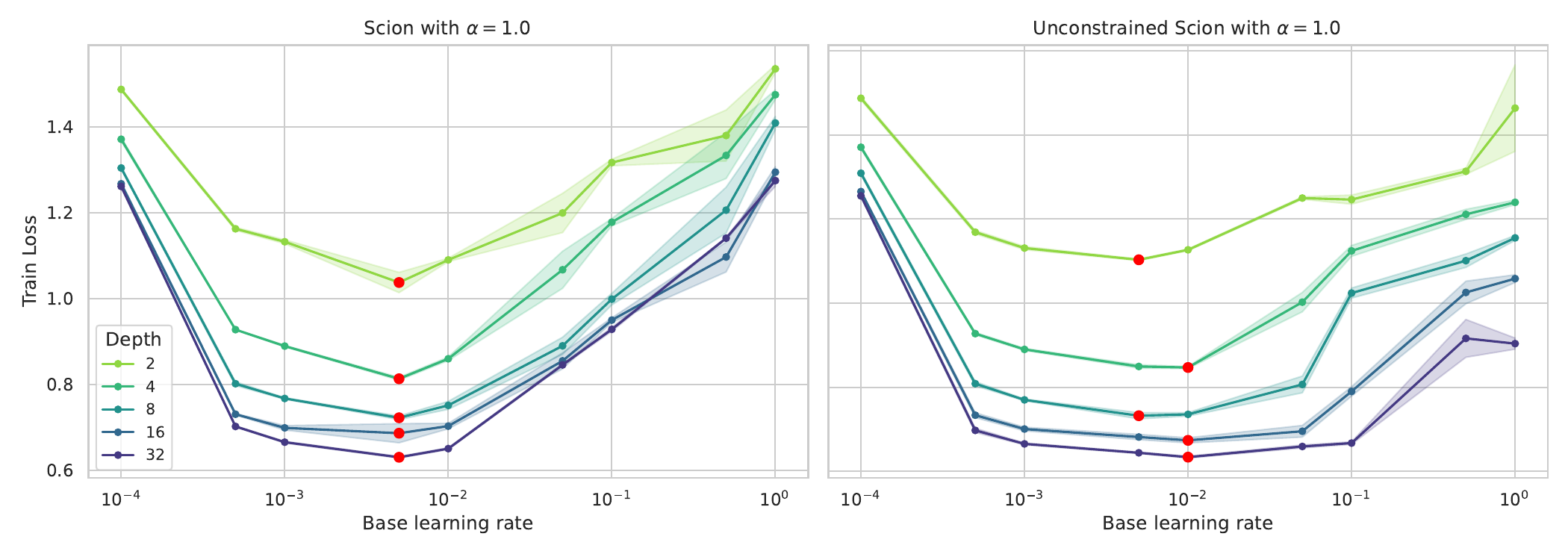}
        \includegraphics[width=0.7\linewidth]{mlp_mixer_cifar/in_between_points/MLPMixer_scion_and_uscion_val_1.0_loss_with_inbetween_points.pdf}
        \includegraphics[width=0.7\linewidth]{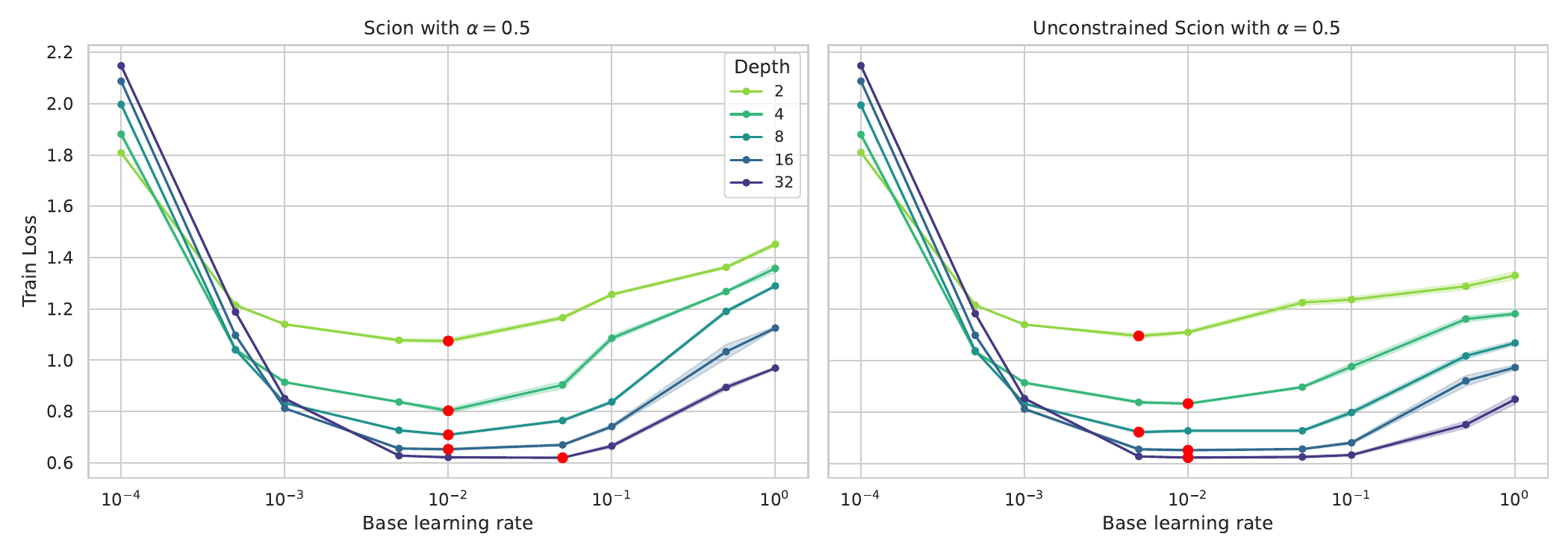}
        \includegraphics[width=0.7\linewidth]{mlp_mixer_cifar/in_between_points/MLPMixer_scion_and_uscion_val_0.5_loss_with_inbetween_points.pdf}

    \caption{
    \Scion (left columns) and \uScion (right columns) on MLP-Mixer trained on CIFAR-10, on 3 seeds.
    The top rows correspond to $\alpha = 1.0$ and the bottom rows to $\alpha = 0.5$.
    Compared to the FineWeb experiments, the loss curves are less parallel across depths, indicating greater heterogeneity in optimization dynamics.
    Nevertheless, the minimizers of the curves, indicated by red dots, remain largely aligned.
    When deviations occur, they lie within a flat region of the loss landscape, where the achieved loss values remain very close.
    The validation loss relative errors are reported in Figures \ref{fig:mlp_mixer_alpha1} and \ref{fig:mlp_mixer_alpha05}. The training loss relative error with \uScion with $\alpha=1$ for depth $L=32$ is approximately $1.4\%$. With $\alpha=0.5$, the relative errors for the deepest models are $1.5\%$ and $0.67\%$ with \Scion and \uScion respectively.
    }
    \label{appdx_fig:mlp_mixer_scion_depth_transfer_alpha_1_and_05}
\end{figure}

    Figure \ref{appdx_fig:mlp_mixer_scion_depth_transfer_alpha_1_and_05} illustrates the hyperparameter transfer in depth with $\alpha=1.0$ and $\alpha=0.5$ for \Scion and \uScion. 

\clearpage
\section{Experimental details}
\label{appdx_sec:exp_details}

    \paragraph{NanoGPT}
    We ran experiments on a snapshot of the speed-running model modded-nanoGPT of \cite{modded_nanogpt_2024}, which is an evolution of the original nanoGPT of \cite{karpathy2023nanogpt}. 
    Specifically, it replaces positional embeddings with rotary embeddings, uses RMSNorm in place of LayerNorm, uses a linear learning rate decay scheduler and employs a $\text{ScaledReLU}^2$ activation function. As this architecture ties the embedding and un-embedding weights, and following Recommendation $3.1$ of \cite{pethick2025trainingdeeplearningmodels}, we configured \Scion and \uScion using the Sign norm for the embedding and un-embedding layers and the Spectral norm for the hidden layers: $\texttt{Sign} \rightarrow \texttt{Spectral} \rightarrow \texttt{Sign}$. The models are trained on the FineWeb dataset \citep{penedo2024the}, for 5100 iterations with a batch size of $512$ sequences of length $1024$, totaling approximately $2.67\times 10^9$ tokens. According to Chinchilla scaling laws \citep{hoffmann2022training}, this data budget is compute optimal for a model with roughly $133\times10^6$ parameters, which corresponds to a depth of $19$ layers. Our evaluation across depths 2, 4, 8, 16, and 32 thus allows us to analyze model performance in both the under-parameterized regime and the over-parameterized regime, with depth 16 serving as our closest near-optimal baseline.
    
    The details on the hyperparameters can be found in Table \ref{tab:NanoGPT_hp}. 
    The experiments were conducted on four H100 GPUs.

    \paragraph{Shallow nanoGPT}
    We also consider a nanoGPT model \citep{karpathy2023nanogpt} where we additionally remove the weight sharing between the first and last layer. Following \cite{pethick2025trainingdeeplearningmodels}, we configured \Scion and \uScion with a $\texttt{Spectral} \rightarrow \texttt{Spectral} \rightarrow \texttt{Sign}$ norm configuration across the embedding, hidden and un-embedding layers.
    The hyperparameters can be found in Table \ref{tab:ShallowGPT_hp}.
    The experiments were performed using four H100 GPUs.

    \paragraph{MLP-Mixer}
    MLP-Mixer is an alternative architecture to transformers and is based entirely on MLPs. It is reported to achieve performance comparable to that of Convolutional Neural Networks and Transformers.  
    Each Mixer layer consists of a token-mixing MLP, and a channel-mixing MLP, with skip-connections. Each of these MLP has two layers and a GELU activation function (Gaussian Error Linear Units function, defined as
    $GELU(x) = x\,\Phi (x)$ where $\Phi$ is the Cumulative Distribution Function for the Gaussian distribution). 
    We replaced the Layer norm from \cite{tolstikhin2021mlpmixerallmlparchitecturevision} with RMSNorm.

    We used random cropping and data augmentation, by generating mixed samples on CIFAR-10. The training was performed on $300$ epochs, with a batch size of $2048$. We used a cosine learning rate scheduler with a linear warm-up. 
    The embedding dimension is set to $128$, the hidden widths of the token-mixing MLP and the channel-mixing MLP are $64$ and $512$ respectively.

    \textit{Why not convolutions?}
    We can note here that \Scion does not provide spectral norm constraint guarantees to convolutional kernels, which also motivates the choice of this architecture.
    Indeed, the current \Scion algorithm applies its constraints to the reshaped kernel tensor, rather than to the linear convolution operator, which corresponds to a large block-Toeplitz matrix. Since the spectral norm of the reshaped kernel is not the spectral norm of the induced convolution operator, the constraint enforced by the $\lmo$ is not applicable to the operator norm. Therefore, \Scion cannot provide reliable spectral constraints for convolutional layers in practice.

    Following Recommendation 3.1 of \cite{pethick2025trainingdeeplearningmodels} for computer vision tasks, we used the Spectral norm for the first and hidden layers and Sign for the last layer: $\texttt{Spectral} \rightarrow \texttt{Spectral} \rightarrow \texttt{Sign}$.

    The details on the hyperparameters can be found in Table \ref{tab:mlpmixer_hp}. The experiments were conducted on four V100 GPUs.   

    \paragraph{DeiT} The DeiT-base model utilized in this work was adapted from the codebase of \cite{pethick2025trainingdeeplearningmodels}, which originates from the official DeiT implementation developed by \cite{pmlr-v139-touvron21a}. Following \cite{pethick2025trainingdeeplearningmodels}, we used the norm configuration $\texttt{Spectral} \rightarrow \texttt{Spectral} \rightarrow \texttt{Sign}$. Additionally, the architecture substituted LayerNorm with RMSNorm, scaled the GELU activation function by a factor of $\sqrt{2}$, and uses bias parameters, which slightly differs from the setup of our other experiments. See Table \ref{tab:DeiT_hp} for details on the hyperparameters. The experiments were conducted on $4 \times4 $ H100 GPUs.

    \begin{table}[!ht]
        \centering
    \caption{NanoGPT hyperparameters.}
    \label{tab:NanoGPT_hp}
        \begin{tabular}{|c|c|} 
            \hline 
            Hyperparameter & \uScion and \Scion \\
            \hline \hline 
            Width &  768   \\
            Head dim &  128   \\
            Activation function & ScaledReLU$^2$   \\
            Vocabulary size &  50304   \\
            Dataset & FineWeb   \\
            Batch size &  512   \\
            Block size &  1024   \\
            Iterations $n$ & 5100 \\
            Warmdown & 28.5\% \\
            Stepsize schedule & Constant then linear decay 
            \\
            Averaging parameter $\beta$ & 0.9\\
            Radii & 3000/50/3000 \\
            \hline
        \end{tabular}
    \end{table}

    \begin{table}[!ht]
        \centering
    \caption{DeiT hyperparameters.}
    \label{tab:DeiT_hp}
        \begin{tabular}{|c|c|} 
            \hline 
            Hyperparameter & \Scion \\
            \hline \hline 
            Head dim &  64   \\
            Activation function & $\sqrt{2}\cdot$ GELU  \\
            Normalization function & RMSNorm \\
            Sequence length &  197   \\
            Dataset & ImageNet-1k   \\
            Batch size &  4096   \\
            Epochs & 200 \\
            Stepsize schedule & Constant then cosine decay \\
            Averaging parameter $\beta$ & 0.9\\
            Radii & 25/25/500 \\
            \hline
        \end{tabular}
    \end{table}

    \begin{table}[!ht]
    \centering
    \caption{Shallow nanoGPT hyperparameters.}
    \label{tab:ShallowGPT_hp}
        \begin{tabular}{|c|c|}
            \hline 
            \makecell{Hyperparameter} & \Gape[1pt]{\makecell{\uScion, \Scion \\ and \pScion}} \\
            \hline \hline 
            Depth (in the width-transfer experiments) & 3 \\
            Width (in the depth-transfer experiments) & 256 \\
            Head dim & 64   \\
            Activation function &  $\sqrt{2}\cdot$ GELU
            \\
            Vocabulary size &  65   \\
            Dataset &  Shakespeare-Char   \\
            Batch size &  32   \\
            Block size &  1024   \\
            Iterations $n$ & 122 \\
            Stepsize schedule & Linear decay 
            \\
            Averaging parameter $\beta$ &  0.9 \\
            Radii (in the width-transfer experiments) & 1/3/10 \\
            \hline
        \end{tabular}
    \end{table}

    \begin{table}[!ht]
    \centering
    \caption{MLP-Mixer hyperparameters.}
    \label{tab:mlpmixer_hp}
        \begin{tabular}{|c|c|}
            \hline 
            Hyperparameter & \uScion and \Scion \\
            \hline \hline 
            Dataset & CIFAR-10 \\
            Embedding dimension & 128 \\
            Token-mixing MLP hidden width & 64 \\
            Channel-mixing MLP hidden width & 512 \\
            Activation & GELU \\
            Batch size & 2048 \\
            Learning rate schedule & Cosine decay with linear warm-up \\
            Training epochs $n$ & 300 \\
            Averaging parameter $\beta$ &  0.9 \\
            Radii & 25/25/500 \\
            \hline
        \end{tabular}
    \end{table}

    \begin{table}[!ht]
    \centering
    \caption{Residual Network hyperparameters (for the relative distance from linearization analysis in Figure \ref{fig:relative_distance_linearization}).}
    \label{tab:resnet_hp}
        \begin{tabular}{|c|c|}
            \hline 
            Hyperparameter & \uScion and \Scion \\
            \hline \hline 
            Dataset & MNIST \\
            Layers per branch & 2 \\
            Activation & ReLU \\
            Batch size & 128 \\
            Iterations $n$ & 1 \\
            Averaging parameter $\beta$ &  0.9 \\
            \hline
        \end{tabular}
    \end{table}
    
    \begin{table}[!ht]
    \centering
    \caption{Residual Network hyperparameters (for the weight alignment experiments in Figures \ref{fig:weight_alignment_W_and_Z_with_phi_h}, \ref{fig:alignment_sum_res_block}, \ref{fig:alignment_sum_update_res_block}, and \ref{fig:alignment_sum_update_res_block_with_scion}).}
    \label{tab:resnet_hp_alignment}
        \begin{tabular}{|c|c|}
            \hline 
            Hyperparameter & \uScion and \Scion \\
            \hline \hline 
            Dataset & CIFAR-10 \\
            Layers per branch & 1 \\
            Width & 300 \\
            Activation & ReLU \\
            Batch size & 1024 \\
            Learning rates $\gamma$ and $\gamma_Z$ & $10^{-2}$ \\
            Averaging parameter $\beta$ & 0.9 \\
            Depth factor $\alpha$ & 1 \\
            Radii & 1/1/1 \\
            \hline
        \end{tabular}
    \end{table}

    \begin{table}[!ht]
    \centering
    \caption{
    MLP hyperparameters (for the weight alignment experiments in Figure
    \ref{fig:alignment_sum_update_with_scion_width}).}
    \label{tab:mlp_hp_alignment}
        \begin{tabular}{|c|c|}
            \hline 
            Hyperparameter & \Scion and \pScion \\
            \hline \hline 
            Dataset & CIFAR-10 \\
            Depth & 3 \\
            Activation & ReLU \\
            Batch size & 1024 \\
            Learning rate $\gamma$ & $10^{-2}$ \\
            Averaging parameter $\beta$ &  0.9 \\
            Radii & 1/1/1 \\
            Interval $[a,b]$ for \pScion & [0, 1] \\
            \hline
        \end{tabular}
    \end{table}   

\clearpage
\section{Proofs of Complete Scion from Section \ref{sec:derivations_depth_transfer_scion_uscion}}

We recall Desiderata 1 and 2 and demonstrate in the following that they hold for \Scion and \uScion.

\subsection{Depth transfer desiderata}
    \RestatableDesStableInit*
            
    \RestatableDesStableUpdate*
    
    In the following, we will use the following argument: $\forall a, b >0, (a+b)^2 = \Theta(a^2+b^2)$. Indeed:
            \begin{equation}
            \label{appdx_eq:a_plus_b_squared}
                 \begin{aligned}
                (a+b)^2 &= \mathcal{O}(a^2+b^2) \text{ since } (a+b)^2=a^2+2ab+b^2\leq 2(a^2+b^2).\\
                \text{and} \quad (a+b)^2 &= \Omega(a^2+b^2) \text{ since } (a+b)^2=a^2+2ab+b^2\geq a^2+b^2 \text{ with } a, b >0,\\
                \text{therefore : } (a+b)^2 &= \Theta(a^2+b^2), \forall a, b >0.
            \end{aligned}
            \end{equation}

\subsection{Assumptions and tools}
    \label{appdx_sec:assumptions_tools}

    Throughout this section, we maintain the same assumptions introduced in the main body of the paper in Section \ref{sec:assumptions_tools}. For brevity, these foundational assumptions are not repeated here.

\paragraph{Update expression of $W^{(\ell)}$}
Since we consider a residual network, we need to compute, for every layer $\ell$
 
\begin{alignat}{2}
    \Delta W^{(\ell)} &= \gamma \operatorname{lmo}(\nabla_{W^{(\ell)}}\mathcal{L}) && \quad \text{with \uScion{}} \label{appdx_eq:delta_W_uscion} \\
    \Delta W^{(\ell)} &= -\gamma W^{(\ell)} + \gamma \operatorname{lmo}(\nabla_{W^{(\ell)}}\mathcal{L}) && \quad \text{with \Scion{}} \label{appdx_eq:delta_W_scion}
\end{alignat}

    We compute the expression for $\rm{lmo}(\nabla_{W^{(\ell)}}\mathcal{L})$.
   
    \begin{equation*}         
        \begin{aligned}
        \nabla_{W^{(\ell)}}\mathcal{L} = \frac{\partial \mathcal{L}}{\partial W^{(\ell)}} = \frac{\partial \mathcal{L}}{\partial h^{(\ell)}} \frac{\partial h^{(\ell)}}{\partial W^{(\ell)}} =  \frac{\partial \mathcal{L}}{\partial h^{(\ell)}}  \frac{1}{L^{\alpha}} \phi(h^{(\ell - 1)})^\top.
        \end{aligned} 
    \end{equation*}
     
   According to lemma C.5 of \cite{pethick2025trainingdeeplearningmodels}, the $\lmo$ of the rank one gradient with batch size one can be written as:
    \begin{equation}
    \label{appdx_eq:hyp_lemma_C5_lmo_expression}
        \begin{aligned}
        \lmo\left(\frac{\partial \mathcal{L}}{\partial h^{(\ell)}} {\phi(h^{(\ell - 1)})}^\top\right)
        = - \rho^{(\ell)} \sqrt{ \frac{d_{\ell}}{d_{\ell-1}}}
        \frac{\frac{\partial \mathcal{L}}{\partial h^{(\ell)}}}
        {\lVert \frac{\partial \mathcal{L}}{\partial h^{(\ell)}} \rVert_2} 
        \frac{ {\phi (h^{(\ell - 1)})}^\top}{\lVert \phi (h^{(\ell - 1)}) \rVert_2}.
        \end{aligned}
    \end{equation}
    
    Therefore, in the case of our residual model, and by using the scale-invariance property of the $\lmo$, we have 

    \begin{equation}
    \label{appdx_eq:lmo_W_l}
        \lmo(\nabla_{W^{(\ell)}}\mathcal{L})= - \rho^{(\ell)} \sqrt{ \frac{d_{\ell}}{d_{\ell-1}}}
        \frac{\frac{\partial \mathcal{L}}{\partial h^{(\ell)}}}
        {\lVert \frac{\partial \mathcal{L}}{\partial h^{(\ell)}} \rVert_2} 
        \frac{ {\phi (h^{(\ell - 1)})}^\top}
        {\lVert \phi (h^{(\ell - 1)}) \rVert_2} \quad (\forall \ell \in \{1, \dots, L\}),
    \end{equation}
    
    \begin{equation}
    \label{appdx_eq:lmo_W_0}
        \lmo(\nabla_{W^{(0)}}\mathcal{L}) = - \rho^{(0)} \sqrt{ \frac{d_{0}}{d_{\text{in}}}}
        \frac{\frac{\partial \mathcal{L}}{\partial h^{(0)}}}
        {\lVert \frac{\partial \mathcal{L}}{\partial h^{(0)}} \rVert_2} 
        \frac{ {x}^\top}
        {\lVert x \rVert_2}, 
    \end{equation}
    
    \begin{equation}
    \label{appdx_appdx_eq:lmo_W_Lp1}
        \lmo(\nabla_{W^{(L+1)}}\mathcal{L}) = - \rho^{(L+1)} 
        \sqrt{ \frac{d_{\text{out}}}{d_{L}}}
        \frac{\frac{\partial \mathcal{L}}{\partial f}}
        {\lVert \frac{\partial \mathcal{L}}{\partial f} \rVert_2} 
        \frac{ {\phi (h^{(L)})}^\top}
        {\lVert \phi (h^{(L)}) \rVert_2}.
    \end{equation}

\paragraph{Update expression of $Z^{(\ell)}$} Similarly, let us suppose we learn the $\{Z^{(\ell)}\}_{\ell \in \{1, \dots, L\}}$ with the associated learning rate $\gamma_Z$ and radii $\{{\rho_Z}_\ell\}_{\ell \in \{1, \dots, L\}}$. Then, we have
\begin{equation*}
    \nabla_{Z^{(\ell)}} \mathcal{L}=\frac{\partial \mathcal{L}}{\partial h^{(\ell)}} \frac{\partial h^{(\ell)}}{\partial Z^{(\ell)}}=\frac{\partial \mathcal{L}}{\partial h^{(\ell)}} {h^{(\ell-1)}}^\top.
\end{equation*}

Therefore, 
\begin{equation}
\label{appdx_eq:lmo_Z_l}
    \Delta Z^{(\ell)}= \gamma_Z \lmo(\nabla_{Z^{(\ell)}} \mathcal{L})= - \gamma_Z {\rho_Z}_\ell \sqrt{\frac{d_\ell}{d_{\ell-1}}} \frac{\frac{\partial \mathcal{L}}{\partial h^{(\ell)}}}{\lVert \frac{\partial \mathcal{L}}{\partial h^{(\ell)}} \rVert_2} \frac{{h^{(\ell-1)}}^\top}{\lVert h^{(\ell-1)} \rVert_2} \quad (\forall \ell \in \{1, \dots, L\}).
\end{equation}      
\subsection{Proof that Desideratum \ref{des:1} holds for \bScion and \ubScion}
\label{appdx-proof:cond1}
    
    \ScionFirstDes*
    
    We provide here the complete demonstration that the Desideratum \ref{des:1} is satisfied for \Scion and \uScion.

    \begin{proof}
        Let $\ell \in \{1, \dots, L\}$.
        Using the alignment of the weights (\Eqref{eq:hyp_alignment}) and the scale of the activation function $\phi$ assumptions (\Eqref{eq:hyp_scale_activation_function}), we have
         \begin{equation*}
            \begin{aligned}
             \lVert h^{(\ell)} \rVert_2^2 &= \Theta( \lVert Z^{(\ell)} h^{(\ell - 1)} \rVert_2^2 + \frac{1}{L^{2\alpha}} \lVert W^{(\ell)} \phi (h^{(\ell - 1)}) \rVert_2^2) 
             \qquad \text{from \Eqref{eq:alignment_sum_res_block}}
             \\
             & = \Theta \left(\lVert Z^{(\ell)} \rVert_{\mathcal{S}_\infty}^2 \lVert h^{(\ell - 1)} \rVert_2^2 + \frac{1}{L^{2\alpha}} \lVert W^{(\ell)}   \rVert_{\mathcal{S}_\infty}^2 \lVert h^{(\ell - 1)} \rVert_2^2 \right) \qquad \text{from \Eqref{eq:hyp_alignment} and \Eqref{eq:hyp_scale_activation_function}} \\
             & = \Theta \left((\lVert Z^{(\ell)} \rVert_{\mathcal{S}_\infty}^2 + \frac{1}{L^{2\alpha}} \lVert W^{(\ell)}  \rVert_{\mathcal{S}_\infty}^2) \lVert h^{(\ell - 1)} \rVert_2^2 \right)\\
             & = \Theta \left(\prod_{k=1}^{\ell} (\lVert Z^{(k)} \rVert_{\mathcal{S}_\infty}^2 + \frac{1}{L^{2\alpha}} \lVert W^{(k)}  \rVert_{\mathcal{S}_\infty}^2) \lVert h^{(0)} \rVert_2^2 \right). 
            \end{aligned}
        \end{equation*}

        \textbf{Remark:} In the above derivation, the assumptions we use are invoked once per layer and composed over $L$ layers. \textit{A priori}, constants hidden by $\Theta(\cdot)$ that are not uniform in $\ell$ or $L$ could compound into a factor growing or collapsing with $L$. We emphasize that our assumptions are stated as holding with constants uniform in $\ell$ and $L$, empirically validated by Figure \ref{fig:alignment_sum_res_block} across multiple depths.

        Moreover, we can control these equivalence constants as follows, to prevent their accumulation.
        Let $r_\ell=\frac{\lVert h^{(\ell)}\rVert_2^2}{\lVert h^{(\ell-1)}\rVert_2^2}=1+2L^{-\alpha}c_\ell + L^{-2\alpha}s_\ell$
        with $c_\ell=\frac{\langle h^{(\ell-1)}, W^{(\ell)} \phi(h^{(\ell-1)})\rangle}{\lVert h^{(\ell-1)}\rVert_2^2}$
        and $s_\ell=\frac{ \lVert W^{(\ell)} \phi(h^{(\ell-1)})\rVert_2^2}{\lVert h^{(\ell-1)}\rVert_2^2}$.
        By Cauchy Schwarz, we have that $|c_\ell| \leq \sqrt{(s_\ell)}$.
        By Lipschitz continuity of $\phi$ and using \Eqref{eq:hyp_init_and_update_weight},
        we have that $s_\ell \leq L_\phi^2 \lVert W^{(\ell)}\rVert_{\mathcal{S}_\infty}^2 \leq S$ with $S$ being a constant uniform in $\ell$ and $L$.
        Then, we have
        \begin{equation}
            \begin{array}{ccc}
                (1-L^{-\alpha}\sqrt{S})^2 \leq &  r_\ell \leq & (1+L^{-\alpha}\sqrt{S})^2\\
            \end{array}
        \end{equation}
        With $\alpha=1$, we have
        \begin{equation}
            \begin{array}{ccccc}
                (1-L^{-1}\sqrt{S})^{2L} & \leq &  \prod_{\ell=1}^L r_\ell = \frac{\lVert h^{(L)}\rVert_2^2}{\lVert h^{(0)}\rVert_2^2} & \leq & (1+L^{-1}\sqrt{S})^{2L}\\
                e^{-2\sqrt{S}}(1+o(1)) & \leq &  \frac{\lVert h^{(L)}\rVert_2^2}{\lVert h^{(0)}\rVert_2^2} & \leq &  e^{2\sqrt{S}}.
            \end{array}
        \end{equation}
        Therefore, both directions hold with constants independent of $L$ and $\ell$.
        For $\alpha=\frac{1}{2}$, we have $e^{-2\sqrt{SL}} \leq \frac{\lVert h^{(L)}\rVert_2^2}{\lVert h^{(0)}\rVert_2^2} \leq e^{2\sqrt{SL}}$
        which is not enough. 
        Using Taylor expansion, we have $\log r_\ell = 2L^{-1/2}c_\ell + L^{-1} s_\ell - \frac{1}{2}\left( 2L^{-1/2} c_\ell + L^{-1} s_\ell \right)^2 + \mathcal{O}(L^{-3/2})=2L^{-1/2}c_\ell + L^{-1}(s_\ell - 2c_\ell^2) + \mathcal{O}(L^{-3/2})$.
        Since $s_\ell$ and $c_\ell$ are bounded by constant, we have $\log r_\ell = 2L^{-1/2}c_\ell + \mathcal{O}(L^{-1})$.
        Then, $\log \frac{\lVert h^{(L)} \rVert_2^2}{\lVert h^{(0)} \rVert_2^2}=\sum_{\ell=1}^L \log r_\ell = 2L^{-1/2}\sum_{\ell=1}^L c_\ell + \mathcal{O}(1)$. 
        Since $c_\ell=\frac{\langle h^{(\ell-1)}, W^{(\ell)} \phi(h^{(\ell-1)})\rangle}{\lVert h^{(\ell-1)}\rVert_2^2}$ form a martingale difference sequence in $\ell$ and $W^{(\ell)}$ is generated from a sign-symmetric distribution with variance $\sigma^2/d_\ell$, we have that, conditionally on the previous state $h^{(\ell-1)}$, $c_\ell$ is a random variable with mean $\mathbb{E}[c_\ell|h^{(\ell-1)}]=0$ and variance $\text{Var}(c_\ell|h^{(\ell-1)}) \leq \frac{L_\phi^2 \sigma^2}{d_\ell}$.
        Hence $\sum_{\ell=1}^L c_\ell = \mathcal{O}_{\mathbb{P}}(\sqrt{L/d})$ and $2L^{-1/2}\sum_{\ell=1}^L c_\ell = \mathcal{O}_{\mathbb{P}}(d^{-1/2})$ uniformly in $L$. 
        And $\frac{\lVert h^{(L)}\rVert_2^2}{\lVert h^{(0)}\rVert_2^2} = e^{\mathcal{O}(1)}e^{\mathcal{O}_{\mathbb{P}}(d^{-1/2})}=e^{\mathcal{O}(1)}(1+o_{\mathbb{P}}(1))$.
        By orthogonality of martingale increments, with $d_{\min}=\min_\ell d_\ell$, we have
        $\mathbb{E}\Big[\Big(\sum_{\ell=1}^L c_\ell\Big)^2\Big]\le L_\phi^2\sigma^2\sum_{\ell=1}^L\frac{1}{d_\ell}\le\frac{L_\phi^2\sigma^2 L}{d_{\min}}$.
        Hence, $2L^{-1/2}\sum_{\ell=1}^L c_\ell=\mathcal{O}_{\mathbb{P}}(d_{\min}^{-1/2})$ uniformly in $L$.
        Together with the $\mathcal{O}(1)$ term, this gives
        $\frac{\lVert h^{(L)}\rVert_2^2}{\lVert h^{(0)}\rVert_2^2}=e^{\mathcal{O}(1)}\,e^{\mathcal{O}_{\mathbb{P}}(d_{\min}^{-1/2})}$,
        which is $e^{\mathcal{O}(1)}(1+o_{\mathbb{P}}(1))$ as soon as $d_{\min}\to\infty$.
        Since $\lVert h^{(0)}\rVert_2^2=\Theta(d_0)$, we obtain with high probability
        $\lVert h^{(L)}\rVert_2^2=\Theta(d_0)$ uniformly in $L$ (and $=\Theta(d)$ if all widths satisfy $d_\ell=\Theta(d)$).

        From the alignment of the weights assumption (\Eqref{eq:hyp_alignment}), combined with the initialization of the weights (\Eqref{eq:hyp_init_and_update_weight}), we have $ \lVert h^{(0)} \rVert_2^2 = \Theta( \lVert W^{(0)} \rVert_{\mathcal{S}_\infty}^2 \lVert x \rVert_2^2) = \Theta \left( \frac{d_0}{d_{\text{in}}} d_{\text{in}} \right) = \Theta(d_0)$.
        Then, 
         \begin{equation*}
            \begin{aligned}
            \lVert h^{(\ell)} \rVert_2^2
             & = \Theta \left(\prod_{k=1}^{\ell} (\lVert Z^{(k)} \rVert_{\mathcal{S}_\infty}^2 + \frac{1}{L^{2\alpha}} \lVert W^{(k)}  \rVert_{\mathcal{S}_\infty}^2) d_0 \right) \\
              & = \Theta \left(\prod_{k=1}^{\ell} (\frac{d_k}{d_{k-1}} + \frac{1}{L^{2\alpha}} \frac{d_k}{d_{k-1}} ) d_0 \right)\\
              & = \Theta \left(\prod_{k=1}^{\ell} (\frac{d_k}{d_{k-1}} \times(1 + \frac{1}{L^{2\alpha}})) d_0 \right)\\
              & = \Theta \left( \frac{d_\ell}{d_{0}} \prod_{k=1}^{\ell} (1 + \frac{1}{L^{2\alpha}}) d_0 \right) \\
               & = \Theta \left( d_\ell \prod_{k=1}^{\ell} (1 + \frac{1}{L^{2\alpha}}) \right).
             \end{aligned}
        \end{equation*}
        
        Then, $\frac{1}{d_L} \lVert h^{(L)} \rVert_2^2 = \Theta  \left( (1 + \frac{1}{L^{2\alpha}} )^L  \right)$ converges when $L$ goes to infinity if and only if $\alpha \geq \frac{1}{2}$.
        Indeed, if we denote $y=\left(1+\frac{1}{L^{2\alpha}} \right)^L$, then, by considering Taylor expansion, we have $\ln(y)=L \ln \left( 1+ \frac{1}{L^{2\alpha}} \right)=L \left( \frac{1}{L^{2\alpha}} - \frac{1}{2L^{4\alpha}} + \mathcal{O}(\frac{1}{L^{6\alpha}}) \right)=L^{1-2\alpha} - \frac{1}{2}L^{1-4\alpha} + \mathcal{O}(L^{1-6\alpha})$. Therefore,
        \begin{equation*}
            \begin{aligned}
                \left\{
                    \begin{array}{cc}
                        \ln(y) \rightarrow \infty \text{ and } y \rightarrow \infty  & \text{if } \alpha<\frac{1}{2}  \\
                        \ln(y) \rightarrow 1 \text{ and } y \rightarrow e  & \text{if } \alpha=\frac{1}{2}  \\
                         \ln(y) \rightarrow 0 \text{ and } y \rightarrow 1  & \text{if } \alpha>\frac{1}{2}.
                    \end{array}
                \right.
            \end{aligned}   
        \end{equation*}
    
         Therefore,
         \begin{equation}   
         \label{appdx_eq:condition_1_h_l}
            \begin{aligned}
            \frac{1}{d_\ell} \lVert h^{(\ell)} \rVert_2^2 & = \Theta (1) \quad \forall \ell \in \{1, \dots, L\} \text{ for } \alpha \geq \frac{1}{2}.\\
            \end{aligned}
        \end{equation}
         
         For $f = W^{(L+1)} \phi(h^{(L)})$, using the alignment of the weights and the scale of the activation function $\phi$ assumptions, we have
         \begin{equation*}
            \begin{aligned}
             \lVert f \rVert_2^2 &=\Theta( \lVert W^{(L+1)} \rVert_{\mathcal{S}_\infty}^2  \lVert h^{(L)}\rVert_2^2 ) 
             \qquad \text{from \Eqref{eq:hyp_alignment} and \Eqref{eq:hyp_scale_activation_function}}.
             \end{aligned}
        \end{equation*}

        Then, using the initialization of the weights and the previous derivations of $\lVert h^{(L)}\rVert_2^2$, we have
        \begin{equation*}
            \begin{aligned}
            \lVert f \rVert_2^2
             &= \Theta \left( \frac{d_{\text{out}}}{d_L} \times d_L \right)
             \qquad \text{from \Eqref{eq:hyp_init_and_update_weight} and \Eqref{appdx_eq:condition_1_h_l}} \\
             &= \Theta (1)\\
             &= \mathcal{O}(1).
            \end{aligned}
        \end{equation*}
         
        These derivations are the same for \Scion, since they only relate to the initialization of the model.
    \end{proof}

\subsection{Proof that Desideratum \ref{des:2} holds for \bScion and \ubScion}
    \ScionSecondDes*
    
    We provide here the complete demonstration that the Desideratum \ref{des:2} is satisfied for \uScion and \Scion.
    
    \subsubsection{\ubScion case}
    \label{appdx-proof:cond2_uScion}
     
     \begin{proof}
        Let $\ell \in \{1, \dots, L\}$.
        For \uScion, we use the expressions of the updates $\Delta W^{(\ell)}$ and $\Delta Z^{(\ell)}$ as in \Eqref{appdx_eq:delta_W_uscion}, that are written as
            \begin{equation}
            \label{appdx_eq:deltaW_uscion}
                \Delta W^{(\ell)} = \gamma \lmo(\nabla_{W^{(\ell)}} \mathcal{L}) 
            \end{equation}

            \begin{equation}
            \label{appdx_eq:deltaZ_uscion}
                \Delta Z^{(\ell)} = \gamma_Z \lmo(\nabla_{Z^{(\ell)}} \mathcal{L}).
            \end{equation}
            
             Writing the expression of the layer $h^{(\ell)}$ in function of the previous $h^{(\ell-1)}$, we have
        
            \begin{equation*}
            \begin{aligned}
                \Delta_{W^{(\ell)}, Z^{(\ell)}}h^{(\ell)} &= h^{(\ell)}_{t+1} - h^{(\ell)}_{t}\\
                &= Z^{(\ell)}_{t+1} h^{(\ell-1)}_{t+1} + L^{-\alpha}W^{(\ell)}_{t+1} \phi(h^{(\ell-1)}_{t+1}) - Z^{(\ell)}_{t} h^{(\ell-1)}_{t} - L^{-\alpha}W^{(\ell)}_{t} \phi(h^{(\ell-1)}_{t}).
            \end{aligned}
            \end{equation*}

            Since we are only looking at the change with the update of $W^{(\ell)}$ and $Z^{(\ell)}$, we have $h^{(\ell - 1)}_{t+1}=h^{(\ell-1)}_{t}$. Therefore,
            \begin{equation*}
            \begin{aligned}
                \Delta_{W^{(\ell)}, Z^{(\ell)}}h^{(\ell)}
                &=Z^{(\ell)}_{t+1} h^{(\ell-1)}_{t}+ L^{-\alpha}W^{(\ell)}_{t+1} \phi(h^{(\ell-1)}_{t}) - Z^{(\ell)}_{t} h^{(\ell-1)}_{t} - L^{-\alpha}W^{(\ell)}_{t} \phi(h^{(\ell-1)}_{t}) \\
                &= \Delta Z^{(\ell)}h_t^{(\ell - 1)} + L^{-\alpha}W^{(\ell)}_{t+1} \phi(h^{(\ell-1)}_{t}) - L^{-\alpha}W^{(\ell)}_{t} \phi(h^{(\ell-1)}_{t})\\
                &= \Delta Z^{(\ell)}h_t^{(\ell - 1)} + L^{-\alpha} (W^{(\ell)}_{t} + \Delta W^{(\ell)}) \phi(h^{(\ell-1)}_{t}) - L^{-\alpha}W^{(\ell)}_{t} \phi(h^{(\ell-1)}_{t})\\
                &= \Delta Z^{(\ell)}h_t^{(\ell - 1)} + L^{-\alpha} \Delta W^{(\ell)} \phi(h^{(\ell-1)}_{t}).\\
                \end{aligned}
            \end{equation*}

            By using the expressions of $\Delta W^{(\ell)}$ and $\Delta Z^{(\ell)}$ from  \Eqref{appdx_eq:deltaW_uscion} and \Eqref{appdx_eq:deltaZ_uscion} combined with the expressions of the associated LMOs in \Eqref{appdx_eq:lmo_W_l} and \Eqref{appdx_eq:lmo_Z_l}, we have 

            \begin{equation*}
            \begin{aligned}
                \lVert \Delta_{W^{(\ell)}, Z^{(\ell)}}h^{(\ell)} \rVert_2^2 
                &= \Theta \left( \lVert \Delta Z^{(\ell)}h_t^{(\ell - 1)} + L^{-\alpha} \Delta W^{(\ell)} \phi(h^{(\ell-1)}_{t}) \rVert_2^2 \right) \\
                &= \Theta \left( 
                \left \lVert 
                \sqrt{\frac{d_{\ell}}{d_{\ell-1}}}  
                \gamma_Z 
                \frac
                {\frac{\partial \mathcal{L}}{\partial h^{(\ell)}}}
                {\lVert \frac{\partial \mathcal{L}}{\partial h^{(\ell)}} \rVert_2}
                \lVert (h^{(\ell-1)}_{t}) \rVert_2
                + L^{-\alpha}
                \sqrt{\frac{d_{\ell}}{d_{\ell-1}}}  
                \gamma 
                \frac
                {\frac{\partial \mathcal{L}}{\partial h^{(\ell)}}}
                {\lVert \frac{\partial \mathcal{L}}{\partial h^{(\ell)}} \rVert_2}
                \lVert \phi(h^{(\ell-1)}_{t}) \rVert_2
                \right \rVert_2^2 \right). 
            \end{aligned}
            \end{equation*}
            We note that $\Delta Z^{(\ell)} h^{(\ell-1)}$ and $L^{-\alpha} \Delta W^{(\ell)} \phi(h^{(\ell-1)})$ are both non-positive multiples of $\frac{\partial \mathcal{L}}{\partial h^{(\ell)}}$, which allows us to write

            \begin{equation*}
            \begin{aligned}
                \lVert \Delta_{W^{(\ell)}, Z^{(\ell)}}h^{(\ell)} \rVert_2^2 
                &= \Theta \left( 
                \left \lVert 
                \sqrt{\frac{d_{\ell}}{d_{\ell-1}}}  
                \gamma_Z 
                \frac
                {\frac{\partial \mathcal{L}}{\partial h^{(\ell)}}}
                {\lVert \frac{\partial \mathcal{L}}{\partial h^{(\ell)}} \rVert_2}
                \lVert (h^{(\ell-1)}_{t}) \rVert_2
                \right \rVert_2^2 
                + L^{-2\alpha}
                \left \lVert 
                \sqrt{\frac{d_{\ell}}{d_{\ell-1}}}  
                \gamma 
                \frac
                {\frac{\partial \mathcal{L}}{\partial h^{(\ell)}}}
                {\lVert \frac{\partial \mathcal{L}}{\partial h^{(\ell)}} \rVert_2}
                \lVert \phi(h^{(\ell-1)}_{t}) \rVert_2
                \right \rVert_2^2 \right) \\
                &= \Theta \left( \frac{d_{\ell}}{d_{\ell-1}} 
                \gamma_Z^2 \lVert h^{(\ell-1)}_{t} \rVert^2_2
                +
                L^{-2\alpha} 
                \frac{d_{\ell}}{d_{\ell-1}} 
                \gamma^2 \lVert \phi(h^{(\ell-1)}_{t}) \rVert^2_2 \right).
            \end{aligned}
            \end{equation*}

            Using the scale of the activation function assumption (\Eqref{eq:hyp_scale_activation_function}), and the Desideratum 1 (\Eqref{appdx_eq:condition_1_h_l}), we have
            \begin{equation*}
                \begin{aligned}
                \lVert \Delta_{W^{(\ell)}, Z^{(\ell)}}h^{(\ell)} \rVert_2^2 &= \Theta \left( 
                \frac{d_{\ell}}{d_{\ell-1}} 
                \gamma_Z^2 \lVert h^{(\ell-1)}_{t} \rVert^2_2
                +
                L^{-2\alpha} 
                \frac{d_{\ell}}{d_{\ell-1}} 
                \gamma^2 \lVert h^{(\ell-1)}_{t} \rVert^2_2 
                \right)
                \\
                &= \Theta \left( 
                d_{\ell}
                \gamma_Z^2
                +
                L^{-2\alpha} 
                d_{\ell} 
                \gamma^2 
                \right)\\
                &= \Theta \left( 
                d_{\ell}( \gamma_Z^2 + L^{-2\alpha} \gamma^2 )
                \right).\\
                \end{aligned}
            \end{equation*}
            
            So, $\frac{1}{d_\ell} \lVert \Delta_{W^{(\ell)}, Z^{(\ell)}}h^{(\ell)} \rVert_2^2 = \Theta( \gamma_Z^2 + L^{-2\alpha} \gamma^2 )$.
            
            Therefore, if we want the entries of $\Delta h^{(\ell)}$ to be in $\Theta(1/L)$, we need to have 
            \begin{equation}
                \gamma = \gamma_{\text{base}} L^{\alpha-1}, \text{ and } \gamma_Z = {\gamma_Z}_{\text{base}}L^{-1}.
            \end{equation}

        Now, let us consider the embedding layer $h^{(0)}=W^{(0)}x$.
        
        Using the update expression $\Delta W^{(0)}$, from \Eqref{appdx_eq:deltaW_uscion}, with the expression of the associated $\lmo$ in \Eqref{appdx_eq:lmo_W_0}, we have
        
        \begin{equation*}
        \begin{aligned}
            \lVert \Delta W^{(0)} x\rVert_2^2 &= \Theta \left( \gamma^2 \frac{d_0}{d_{\text{in}}} \lVert x \rVert_2^2 \right) = \Theta \left(\gamma^2 d_0 \right) = \gamma^2 \Theta(d_0).
        \end{aligned}
        \end{equation*}

        Therefore, we do not need to scale the learning rate of the first weight $W^{(0)}$ in order to respect Desideratum 2, and
         \begin{equation}
                \gamma_{W^{(0)}} = {\gamma_{W^{(0)}}}_{\text{base}}.
            \end{equation}

        Finally, for the output layer $f=W^{(L+1)} \phi(h^{(L)})$, using  the update expression $\Delta W^{(L+1)}$ from \Eqref{appdx_eq:deltaW_uscion} with the associated $\lmo$ expressions in \Eqref{appdx_appdx_eq:lmo_W_Lp1}, and Desideratum 1 (\Eqref{appdx_eq:condition_1_h_l}) applied to $h^{(L)}$, we have
        
       \begin{equation*}
        \begin{aligned}
            \lVert \Delta W^{(L+1)} h^{(L)} \rVert_2^2 &= 
            \Theta \left( \gamma^2 \frac{d_{\mathrm{out}}}{d_L} \frac{\lVert \frac{\partial \mathcal{L}}{\partial f}\rVert_2^2}{\lVert \frac{\partial \mathcal{L}}{\partial f} \rVert_2^2}
            \frac{\lVert \phi(h^{(L)}) \rVert_2^2}{\lVert \phi(h^{(L)}) \rVert_2^2} \lVert h^{(L)}\rVert_2^2 \right) \\
            &= \Theta \left(  \gamma^2 \frac{d_{\mathrm{out}}}{d_L} \lVert h^{(L)}\rVert_2^2 \right)\\
            &= \gamma^2 \Theta \left( d_{\mathrm{out}} \right).
        \end{aligned}
        \end{equation*}

        Therefore, we do not need to scale the learning rate of the last weight $W^{(L+1)}$ in order to respect Desideratum 2.
         \begin{equation}
                \gamma_{W^{(L+1)}} = {\gamma_{W^{(L+1)}}}_{\text{base}}.
            \end{equation}
            
    \end{proof}

    \subsubsection{\bScion case}
    \label{appdx-proof:cond2_Scion}

    \begin{proof}
        Let $\ell \in \{1, \dots, L\}$.
            For \Scion, we use the expressions of the updates $\Delta W^{(\ell)}$ and $\Delta Z^{(\ell)}$ as in \Eqref{appdx_eq:delta_W_scion}, that are written as
            \begin{equation}
            \label{appdx_eq:deltaW_scion}
                \Delta W^{(\ell)} = -\gamma W^{(\ell)} + \gamma \lmo(\nabla_{W^{(\ell)}} \mathcal{L})
            \end{equation}

            \begin{equation}
             \label{appdx_eq:deltaZ_scion}
                \Delta Z^{(\ell)} = -\gamma_Z Z^{(\ell)} + \gamma_Z \lmo(\nabla_{Z^{(\ell)}} \mathcal{L}).
            \end{equation}

            Then,
            \begin{equation*}
             \begin{aligned}
            \lVert \Delta_{W^{(\ell)}, Z^{(\ell)}}h^{(\ell)} \rVert_2^2 
            &= \lVert \Delta Z^{(\ell)}h_t^{(\ell - 1)} + L^{-\alpha} \Delta W^{(\ell)} \phi(h^{(\ell-1)}_{t}) \rVert_2^2 \\
            &= \Theta \left( \lVert \Delta Z^{(\ell)}h_t^{(\ell - 1)} \rVert_2^2 + \lVert L^{-\alpha} \Delta W^{(\ell)} \phi(h^{(\ell-1)}_{t}) \rVert_2^2 \right) 
            \qquad \text{from \Eqref{eq:alignment_sum_update_res_block}}
            \\
            &= \Theta \left( 
            \lVert (-\gamma_Z Z^{(\ell)} + \gamma_Z \lmo(\nabla_{Z^{(\ell)}} \mathcal{L})) h^{(\ell-1)}_{t} \rVert_2^2
            \right.\\
            & \quad \left.
            +
            \lVert L^{-\alpha} (-\gamma W^{(\ell)} + \gamma \lmo(\nabla_{W^{(\ell)}} \mathcal{L})) \phi(h^{(\ell-1)}_{t}) \rVert_2^2 \right) \\
            &= \Theta \left(
            \lVert \gamma_Z Z^{(\ell)}  h^{(\ell-1)}_{t} \rVert_2^2 
            +
            \lVert \gamma_Z \lmo(\nabla_{Z^{(\ell)}} \mathcal{L}) h^{(\ell-1)}_{t}
            \rVert_2^2 \right.\\
            & \quad \left. +
            \lVert L^{-\alpha} \gamma W^{(\ell)}  \phi(h^{(\ell-1)}_{t}) \rVert_2^2 
            +
            \lVert L^{-\alpha} \gamma \lmo(\nabla_{W^{(\ell)}} \mathcal{L}) \phi(h^{(\ell-1)}_{t}) \rVert_2^2 
            \right).
            \quad \text{from \Eqref{eq:alignment_sum_update_res_block_with_scion}}
            \\
            \end{aligned}
            \end{equation*}
                
            We saw in the unconstrained case that the second and fourth terms are in $\Theta (d_{\ell}( \gamma_Z^2 + L^{-2\alpha} \gamma^2 ))$. 
            
            For the third term, using the alignment of the weights (\Eqref{eq:hyp_alignment}) and the scale of the activation function $\phi$ (\Eqref{eq:hyp_scale_activation_function}) assumptions, we have
            \begin{equation*}
                \begin{aligned}
                \lVert L^{-\alpha} \gamma W^{(\ell)}  \phi(h^{(\ell-1)}_{t}) \rVert_2^2 
                &= \Theta \left( L^{-2\alpha} \gamma^2 \lVert W^{(\ell)} \rVert_{\mathcal{S}_\infty}^2 \lVert \phi(h^{(\ell-1)}_{t}) \rVert_2^2 \right)
                \\
                &=  \Theta \left( L^{-2\alpha} \gamma^2 \lVert W^{(\ell)} \rVert_{\mathcal{S}_\infty}^2 \lVert h^{(\ell-1)}_{t} \rVert_2^2 \right).
                \end{aligned}
            \end{equation*}

            Then, using the initialization of the weights in \Eqref{eq:hyp_init_and_update_weight} and the Desideratum 1 in \Eqref{appdx_eq:condition_1_h_l}, we have 

            \begin{equation*}
                \begin{aligned}
                \lVert L^{-\alpha} \gamma W^{(\ell)}  \phi(h^{(\ell-1)}_{t}) \rVert_2^2 
                &=  \Theta \left( L^{-2\alpha} \gamma^2 \frac{d_{\ell}}{d_{\ell-1}}  d_{\ell-1} \right) 
                \\
                &=  \Theta \left( L^{-2\alpha} \gamma^2 d_{\ell} \right).
                \\
                \end{aligned}
            \end{equation*}
            
            Similarly, for the first term, we have
            \begin{equation*}
                \begin{aligned}
                \lVert \gamma_Z Z^{(\ell)}  h^{(\ell-1)}_{t} \rVert_2^2 
                &= \Theta \left( \gamma_Z^2 \lVert Z^{(\ell)} \rVert_{\mathcal{S}_\infty}^2 \lVert h^{(\ell-1)}_{t} \rVert_2^2 \right)
                \\
                &=  \Theta \left( \gamma_Z^2 \frac{d_{\ell}}{d_{\ell-1}}  d_{\ell-1} \right) 
                \\
                &=  \Theta \left( \gamma_Z^2 d_{\ell} \right).
                \\
                \end{aligned}
            \end{equation*}
                
            Finally, we have
             \begin{equation*}
                \begin{aligned}
                 \lVert \Delta_{W^{(\ell)}, Z^{(\ell)}}h^{(\ell)} \rVert_2^2 = \Theta \left(d_{\ell}( \gamma_Z^2 + L^{-2\alpha} \gamma^2 ) \right). 
                 \end{aligned}
            \end{equation*}
            We can conclude in the same way as in the unconstrained case.

    Now, let us consider the embedding layer $h^{(0)}=W^{(0)}x$.
    With \Scion, using the update expression of $W^{(0)}$ in \Eqref{appdx_eq:deltaW_scion}, combined with the associated $\lmo$ expression in \Eqref{appdx_eq:lmo_W_0}, we have
    \begin{equation*}
    \begin{aligned}
        \Delta W^{(0)} &= -\gamma W^{(0)} + \gamma \text{lmo}(\nabla _{W^{(0)}} \mathcal{L}) \\
        & = -\gamma W^{(0)} - \gamma \rho^{(0)}\sqrt{\frac{d_0}{d_{\text{in}}}} \frac{\frac{\partial \mathcal{L}}{\partial h^{(0)}}}{\lVert \frac{\partial \mathcal{L}}{\partial h^{(0)}} \rVert_2} \frac{x^\top}{\rVert x \lVert_2}.
    \end{aligned}
    \end{equation*}
    Then, using the alignment of the weights (\Eqref{eq:hyp_alignment}) and the initialization of $W^{(0)}$ (\Eqref{eq:hyp_init_and_update_weight}) assumptions, we have
    \begin{equation*}
    \begin{aligned}
        \lVert \Delta W^{(0)} x\rVert_2^2 &= \Theta \left(
        \gamma^2 \lVert W^{(0)} x\rVert_2^2
        + \gamma^2 \frac{d_0}{d_{\text{in}}} \lVert x \rVert_2^2 
        \right) 
        \quad \text{from \Eqref{eq:alignment_sum_update_res_block_with_scion}}
        \\ 
        &= \Theta \left(
        \gamma^2 \lVert W^{(0)} \rVert_{\mathcal{S}_\infty}^2 \lVert x \rVert_2^2
        + \gamma^2 \frac{d_0}{d_{\text{in}}} \lVert x \rVert_2^2 
        \right) \quad \text{(Claim 1 in \cite{Yang2023ASpectralCondtion})} \\
        &= \Theta \left(
        \gamma^2 \frac{d_0}{d_{\text{in}}} d_{\text{in}}
        + \gamma^2 \frac{d_0}{d_{\text{in}}} d_{\text{in}}
        \right)\\
        &= \Theta \left(
        2 \gamma^2 d_0
        \right) \\
        &= \gamma^2 \Theta \left( d_0 \right). 
    \end{aligned}
    \end{equation*}

    We can conclude similarly to the unconstrained case.

    Finally, for the output layer $f=W^{(L+1)} \phi(h^{(L)})$, if we use the update expression of $W^{(L+1)}$ in \Eqref{appdx_eq:deltaW_scion}, combined with the associated $\lmo$ expression in \Eqref{appdx_appdx_eq:lmo_W_Lp1},we have
    \begin{equation*}
    \begin{aligned}
        \Delta W^{(L+1)}
        &= -\gamma W^{(L+1)} - \gamma \rho^{(L+1)} \sqrt{\frac{d_{\mathrm{out}}}{d_L}} \frac{\frac{\partial \mathcal{L}}{\partial f}}{\lVert \frac{\partial \mathcal{L}}{\partial f} \rVert_2} \frac{\phi(h^{(L)})^\top}{\lVert  \phi(h^{(L)}) \rVert_2}.
        \end{aligned}
    \end{equation*}
    Then, using the alignment of the weights (\Eqref{eq:hyp_alignment}), the initialization of $W^{(L+1)}$ (\Eqref{eq:hyp_init_and_update_weight}) and the Desideratum 1 (\Eqref{appdx_eq:condition_1_h_l}) applied to $h^{(L)}$, we have
    \begin{equation*}
    \begin{aligned}
        \lVert \Delta W^{(L+1)} h^{(L)} \rVert_2^2
        &= \Theta \left(
        \gamma^2 \lVert W^{(L+1)} h^{(L)} \rVert_2^2
        + \gamma^2 \lVert \text{lmo}(\nabla_{W^{(L+1)}} \mathcal{L}) h^{(L)} \rVert_2^2
        \right) 
        \quad \text{(from \Eqref{eq:alignment_sum_update_res_block_with_scion})}\\
        &= \Theta \left(
        \gamma^2 \lVert W^{(L+1)}\rVert_{\mathcal{S}_\infty}^2 \lVert h^{(L)} \rVert_2^2
        + \gamma^2 d_{\mathrm{out}}
        \right)
        \\
        &= \Theta \left(
        \gamma^2 \frac{d_{\mathrm{out}}}{d_L} d_L 
        + \gamma^2 d_{\mathrm{out}}
        \right)
        \\
        &= \Theta \left(
        \gamma^2 d_{\mathrm{out}}
        \right).
    \end{aligned}
    \end{equation*}

    And we can conclude the same way as in the unconstrained case.
    
    \end{proof}

    \subsection{Proof that Desiderata \ref{des:1} and \ref{des:2} hold for any step $t$}
    \label{appdx-proof:cond1_2_induction}
    The proof of Desideratum \ref{prop:des2} relies on the results in Desideratum \ref{des:1}, which is stated only for initialization. In other words, in Appendix \ref{appdx-proof:cond2_uScion} and \ref{appdx-proof:cond2_Scion}, we only proved Desideratum \ref{des:2} for the step $t=1$. 
   
    In this section, we demonstrate that Desideratum \ref{des:1} holds for any step $t$ by induction, which implied that Desideratum \ref{des:2} holds for any step $t$ as well.
    \begin{proof}
        Let us show by induction that, for all steps $t$ we have $\lVert h_t^{(\ell)}\rVert_2^2 = \Theta(d_\ell)$, $\lVert W_{t}^{(\ell)} \rVert_{\mathcal{S}_\infty}^2= \Theta(\frac{d_\ell}{d_{\ell-1}})$, and $\lVert Z_{t}^{(\ell)} \rVert_{\mathcal{S}_\infty}^2= \Theta(\frac{d_\ell}{d_{\ell-1}})$.
        The initialization step is already verified by Desideratum \ref{des:1}.
        Now, let us assume that $\lVert h_t^{(\ell)}\rVert_2^2 = \Theta(d_\ell)$ holds at step $t$.
        Then,
        \begin{equation}
        \begin{aligned}
        \lVert h_{t+1}^{(\ell)} \rVert_2^2 
        &= \lVert Z_{t+1}^{(\ell)} h_{t+1}^{(\ell-1)} + \frac{1}{L^\alpha} W_{t+1}^{(\ell)} \phi(h_{t+1}^{(\ell-1)}) \rVert_2^2 \\
        &= \Theta \left(
        \lVert Z_{t+1}^{(\ell)} h_{t+1}^{(\ell-1)} \rVert_2^2 + \frac{1}{L^{2 \alpha}} \lVert W_{t+1}^{(\ell)} \phi(h_{t+1}^{(\ell-1)}) \rVert_2^2
        \right) 
        \qquad \text{from \Eqref{eq:alignment_sum_res_block}}\\
        &= \Theta \left(
        \lVert Z_{t+1}^{(\ell)} \rVert_{\mathcal{S}_\infty}^2  \lVert h_{t+1}^{(\ell-1)} \rVert_2^2 + \frac{1}{L^{2 \alpha}} \lVert W_{t+1} ^{(\ell)} \rVert_{\mathcal{S}_\infty}^2  \lVert h_{t+1}^{(\ell-1)} \rVert_2^2
        \right) 
        \qquad \text{from \Eqref{eq:hyp_alignment} and \Eqref{eq:hyp_scale_activation_function}}\\
        &= \Theta \left(
        \left(
        \lVert Z_{t+1}^{(\ell)}  \rVert_{\mathcal{S}_\infty}^2  + \frac{1}{L^{2 \alpha}} \lVert W_{t+1} ^{(\ell)} \rVert_{\mathcal{S}_\infty}^2  
        \right)
        \lVert h_{t+1}^{(\ell-1)} \rVert_2^2
        \right) \\
        &= \Theta \left(
        \prod_{k=1}^\ell
        \left(
        \lVert Z_{t+1}^{(k)}  \rVert_{\mathcal{S}_\infty}^2  + \frac{1}{L^{2 \alpha}} \lVert W_{t+1} ^{(k)} \rVert_{\mathcal{S}_\infty}^2  
        \right)
        \lVert h_{t+1}^{(0)} \rVert_2^2
        \right).
        \end{aligned}
        \end{equation}
        And, under the assumption that the updates do not perfectly cancel current quantities (\Eqref{eq:alignment_init_and_update}), we have, for all layers $\ell$,
        \begin{equation}
        \begin{aligned}
            \lVert Z_{t+1}^{(\ell)}  \rVert_{\mathcal{S}_\infty}^2
            &= \lVert Z_{t}^{(\ell)}  + \Delta Z^{(\ell)} \rVert_{\mathcal{S}_\infty}^2 \\
            &= \Theta \left(
            \lVert Z_{t}^{(\ell)} \rVert_{\mathcal{S}_\infty}^2 + \lVert \Delta Z^{(\ell)} \rVert_{\mathcal{S}_\infty}^2
            \right) \\
            &= \Theta \left(
            \frac{d_\ell}{d_{\ell-1}} + \frac{d_\ell}{d_{\ell -1}} 
            \right)
            \qquad \text{by induction, and using \Eqref{eq:hyp_init_and_update_weight}.}
            \\
            &= \Theta \left( \frac{d_\ell}{d_{\ell-1}} \right).
        \end{aligned}
        \end{equation}
        Similarly, we have, for all layers $\ell$,
        \begin{equation}
            \lVert W_{t+1}^{(\ell)}  \rVert_{\mathcal{S}_\infty}^2
            = \Theta \left( \frac{d_\ell}{d_{\ell-1}} \right).
        \end{equation}
        Besides,
        \begin{equation}
        \begin{aligned}
        \lVert h^{(0)}_{t+1} \rVert_2^2 = \lVert W_{t+1}^{(0)} x\rVert_2^2 =  
        \Theta \left(\lVert W_{t+1}^{(0)}  \rVert_{\mathcal{S}_\infty}^2 \lVert x\rVert_2^2 \right) = \Theta \left( \frac{d_0}{d_{\text{in}}} d_{\text{in}}\right) = \Theta(d_0).
        \end{aligned}
        \end{equation}
        Therefore,
        \begin{equation}
        \begin{aligned}
        \lVert h_{t+1}^{(\ell)} \rVert_2^2 
        &= \Theta \left(
        \prod_{k=1}^\ell
        \left(
        \lVert Z_{t+1}^{(k)}  \rVert_{\mathcal{S}_\infty}^2  + \frac{1}{L^{2 \alpha}} \lVert W_{t+1} ^{(k)} \rVert_{\mathcal{S}_\infty}^2  
        \right)
        \lVert h_{t+1}^{(0)} \rVert_2^2
        \right) \\
        &= \Theta \left(
        \prod_{k=1}^\ell
        \left(
        \frac{d_k}{d_{k-1}}  + \frac{1}{L^{2 \alpha}}  \frac{d_k}{d_{k-1}}
        \right)
        d_0
        \right)\\
        &= \Theta \left(
        d_\ell 
        \underbrace{\left(1  + \frac{1}{L^{2 \alpha}} \right)^\ell}_{\Theta(1) \text{ with } \alpha \geq \frac{1}{2}}
        \right)\\
        &=\Theta(d_\ell),
        \end{aligned}
        \end{equation}
        which ends the proof.\\
        We note that this also naturally leads to $\lVert f \rVert_2^2 = \mathcal{O}(1)$, following the same steps as in Appendix \ref{appdx-proof:cond1}.
    \end{proof}

\section{Proofs of the Width Transfer from Section \ref{sec:width_wise_hp_transfer}}

We recall Desideratum 3 and demonstrate that it holds for \Scion, extending the prior width-transfer results derived by \cite{pethick2025trainingdeeplearningmodels} for \uScion.

    \RestatableDesFeatureLearning*

    \subsection{Proof of the width transfer with \bScion, from Section \ref{sec:width_wise_hp_transfer_for_scion}}
    \label{appdx-proof:width_transfer_scion}
    
    \ScionWidth*

    \begin{proof}
        Let $\ell \in \{1, \dots, L\}$.
        With \Scion, the update of the weight matrix $W^{(\ell)}$ can be written as: 
        
        \begin{equation*}
            \Delta W^{(\ell)} = -\gamma W^{(\ell)} + \gamma \lmo(\nabla_{W_t^{(\ell)}}\mathcal{L}).
        \end{equation*}
            
      By using \Eqref{eq:alignment_sum_update_width_transfer} and \Eqref{eq:weight_lmo_spectral_cancellation_scion} combined with the expression of the lmo in \Eqref{appdx_eq:lmo_W_l}, we have
       \begin{equation}
        \label{appdx_eq:delta_h_l_scion_width}
        \begin{aligned}
             \lVert \Delta h^{(\ell)}\rVert_2^2
             & =   \Theta \left( 
            \gamma^2 \lVert W^{(\ell)} \phi(h^{(\ell-1)}) \rVert_2^2
            + \gamma^2  
            \frac{d_\ell}{d_{\ell-1}}
            \lVert \phi(h^{(\ell-1)})\rVert_2^2
            \right. \\
             & \qquad \left.
            + \lVert W^{(\ell)} \Delta \phi(h^{(\ell-1)})\rVert_2^2 
            +\lVert \Delta W^{(\ell)} \Delta \phi(h^{(\ell-1)})\rVert_2^2 
            \right).
        \end{aligned}
       \end{equation}
       
       By using the alignment of the weights (\Eqref{eq:hyp_alignment}), the initialization of the weights $W^{(\ell)}$ (\Eqref{eq:hyp_init_and_update_weight}), and the scale of the activation function $\phi$ assumption in \Eqref{eq:hyp_scale_activation_function}, we have
        \begin{equation*}
            \lVert h^{(\ell)} \rVert_2 = \Theta(\lVert W^{(\ell)} \rVert_{\mathcal{S}_\infty} \lVert h^{(\ell-1)} \rVert_2) = \Theta(\sqrt{d_\ell}).
        \end{equation*}

       With the scale of the activation function $\phi$ assumption in \Eqref{eq:hyp_scale_activation_function}, we also have:
        \begin{equation}    
        \label{appdx_eq:phi_hl_dl}
        \begin{aligned}
            \lVert \phi(h^{(\ell)}) \rVert_2^2 = \Theta(d_\ell).
       \end{aligned}         
       \end{equation}
        
        Therefore, by using the weight alignment assumption in \Eqref{eq:hyp_alignment} and the weight initialization as in \Eqref{eq:hyp_init_and_update_weight}, we have
        \begin{equation}   
        \label{appdx_eq:Wl_phi_h}
        \begin{aligned}
            \lVert W^{(\ell)} \phi(h^{(\ell-1)}) \rVert_2^2  
            & = \Theta(  \lVert W^{(\ell)} \rVert_{S_\infty}^2 \lVert \phi(h^{(\ell-1)}) \rVert_2^2  ) \\
            & = \Theta \left( 
            \frac{d_\ell}{d_{\ell-1}} 
            \lVert h^{(\ell-1)} \rVert_2^2 \right)\\
            & = \Theta \left( \frac{d_\ell}{d_{\ell-1}} d_{\ell-1}
             \right)\\
            & = \Theta \left( d_\ell \right).
       \end{aligned}         
       \end{equation}

        Likewise, 
        \begin{equation} 
        \label{appdx_eq:Wl_delta_phi}
        \begin{aligned}
            \lVert W^{(\ell)} \Delta \phi(h^{(\ell-1)}) \rVert_2^2
            &= \Theta \left( 
            \frac{d_\ell}{d_{\ell-1}}
            \lVert \Delta \phi(h^{(\ell-1)}) \rVert_2^2\right) \\
            &= \mathcal{O} \left(
            \frac{d_\ell}{d_{\ell-1}}
            \lVert \Delta h^{(\ell -1)} \rVert_2^2 \right) \\
            & \quad \text{(Lipschitz activation function)}.
       \end{aligned}         
       \end{equation}
        
        To write the term, $\Delta W^{(\ell)} \Delta \phi(h^{(\ell-1)})$, we can decompose the weight update $\Delta W^{(\ell)}$ and use the $\lmo$ expression of \Eqref{appdx_eq:lmo_W_l}, to get
        
        \begin{equation*}             
        \begin{aligned}
            \Delta W^{(\ell)} \Delta \phi(h^{(\ell-1)}) & = 
            - \gamma \rho^{(\ell)}
            \sqrt{\frac{d_\ell}{d_{\ell-1}}} 
            \frac{(\frac{\partial \mathcal{L}}{\partial h^{(\ell)}})}{\lVert \frac{\partial \mathcal{L}}{\partial h^{(\ell)}}\rVert_ 2} 
            \frac{{\phi(h^{(\ell-1)})}^\top}{\lVert \phi(h^{(\ell-1)})\rVert_2} 
            \Delta \phi(h^{(\ell-1)}) 
            \\ & \qquad 
            - \gamma W^{(\ell)} \Delta \phi(h^{(\ell-1)}).
        \end{aligned}         
       \end{equation*}

       So, 
        \begin{equation*}             
        \begin{aligned}
            \lVert \Delta W^{(\ell)} \Delta \phi(h^{(\ell-1)}) \rVert_2^2 & = \mathcal{O}
            \left(
            \gamma^2 
            \frac{d_\ell}{d_{\ell-1}}
            \left \lVert \frac{(\frac{\partial \mathcal{L}}{\partial h^{(\ell)}})}{\lVert \frac{\partial \mathcal{L}}{\partial h^{(\ell)}} \rVert_ 2} 
            \frac{{\phi(h^{(\ell-1)})}^\top}{\lVert \phi(h^{(\ell-1)})\rVert_2} 
            \Delta \phi(h^{(\ell-1)})  \right  \rVert_2^2
            \right. \\ & \quad \left.
            + \gamma^2 \lVert W^{(\ell)} \Delta \phi(h^{(\ell-1)}) \rVert_2^2
            \right) \\
            &= 
            \mathcal{O}
            \left(
            \gamma^2 
            \frac{d_\ell}{d_{\ell-1}} 
            \left \lVert \frac{(\frac{\partial \mathcal{L}}{\partial h^{(\ell)}})}{\lVert 
            \frac{\partial \mathcal{L}}{\partial h^{(\ell)}} 
            \rVert_ 2} 
            \frac{{\phi(h^{(\ell-1)})}^\top}{\lVert \phi(h^{(\ell-1)})\rVert_2} 
            \Delta \phi(h^{(\ell-1)})  \right  \rVert_2^2
            \right. \\ & \quad \left.
            + \gamma^2 
            \frac{d_\ell}{d_{\ell-1}} 
            \lVert \Delta \phi(h^{(\ell-1)}) \rVert_2^2
            \right).
         \end{aligned}         
       \end{equation*}
       
        By using the fact that the activation function $\phi$ is Lipschitz, we have
        \begin{equation*}             
        \begin{aligned}
            \lVert \Delta W^{(\ell)} \Delta \phi(h^{(\ell-1)}) \rVert_2^2
            &=  \mathcal{O} \left(
            \gamma^2 
            \frac{d_\ell}{d_{\ell-1}} 
            \left \lVert \frac{(\frac{\partial \mathcal{L}}{\partial h^{(\ell)}})}{\lVert 
            \frac{\partial \mathcal{L}}{\partial h^{(\ell)}} 
            \rVert_ 2} 
            \frac{{\phi(h^{(\ell-1)})}^\top}{\lVert \phi(h^{(\ell-1)})\rVert_2} 
            \Delta \phi(h^{(\ell-1)})  \right  \rVert_2^2
            \right. \\ & \quad \left.
            + \gamma^2 
            \frac{d_\ell}{d_{\ell-1}} 
            \lVert \Delta h^{(\ell-1)} \rVert_2^2
            \right).
       \end{aligned}         
       \end{equation*}

        We rewrite the first term and use the Lipschitz continuity of the activation function $\phi$, to get
        \begin{equation*}             
        \begin{aligned}
            \gamma^2 
            \frac{d_\ell}{d_{\ell-1}} 
            \left \lVert \frac{(\frac{\partial \mathcal{L}}{\partial h^{(\ell)}})}{\lVert \frac{\partial \mathcal{L}}{\partial h^{(\ell)}} \rVert_ 2} 
            \frac{{\phi(h^{(\ell-1)})}^\top}{\lVert \phi(h^{(\ell-1)})\rVert_2} 
            \Delta \phi(h^{(\ell-1)})  \right  \rVert_2^2 
            & = \Theta \left( \gamma^2 \frac{d_\ell}{d_{\ell-1}}
            \lVert \Delta \phi(h^{(\ell-1)}) \rVert_2^2 \right) \\
             & = \mathcal{O} \left( \gamma^2 \frac{d_\ell}{d_{\ell-1}}
             \lVert \Delta h^{(\ell -1)} \rVert_2^2 \right).
       \end{aligned}         
       \end{equation*}
        
        Therefore,
        \begin{equation*}
        \begin{aligned}
           \lVert \Delta W^{(\ell)} \Delta \phi(h^{(\ell-1)}) \rVert_2^2
            & = \mathcal{O} \left( \gamma^2 \frac{d_\ell}{d_{\ell-1}}  
            \lVert \Delta h^{(\ell -1)} \rVert_2^2 + \gamma^2 
            \frac{d_\ell}{d_{\ell-1}}  
            \lVert \Delta h^{(\ell-1)} \rVert_2^2 \right)\\
            & = \mathcal{O} \left( 2 \gamma^2 
            \frac{d_\ell}{d_{\ell-1}} 
            \lVert \Delta h^{(\ell -1)} \rVert_2^2  \right).\\
       \end{aligned}         
       \end{equation*}

        Finally, we can incorporate all these derivated terms into the expression of $\lVert \Delta h^{(\ell)}\rVert_2^2$ from \Eqref{appdx_eq:delta_h_l_scion_width}, and we have

       \begin{equation*}
        \begin{aligned}
            \lVert \Delta h^{(\ell)}\rVert_2^2
            & =   \Theta \left( 
            \gamma^2 \lVert W^{(\ell)} \phi(h^{(\ell-1)}) \rVert_2^2 
            + \gamma^2  
            \frac{d_\ell}{d_{\ell-1}}
            \lVert \phi(h^{(\ell-1)})\rVert_2^2
            \right. \\ & \qquad \left.
            + \lVert W^{(\ell)} \Delta \phi(h^{(\ell-1)})\rVert_2^2
            + \lVert \Delta W^{(\ell)} \Delta \phi(h^{(\ell-1)})\rVert_2^2 
            \right)\\
            &= \mathcal{O} \left(
            \gamma^2 d_\ell
            + \gamma^2 d_\ell 
            +
             \frac{d_\ell}{d_{\ell-1}}  \lVert \Delta h^{(\ell -1)} \rVert_2^2
            \right. \\ & \qquad \left.
            + 2 \gamma^2 
            \frac{d_\ell}{d_{\ell-1}} 
            \lVert \Delta h^{(\ell -1)} \rVert_2^2
            \right).
        \end{aligned}         
       \end{equation*}

        Thus,
        \begin{equation}
        \label{appdx_eq:recursion_delta_hl}
            \lVert \Delta h^{(\ell)}\rVert_2^2   = \mathcal{O} \left(
            2 \gamma^2 d_\ell 
            + (1+ 2\gamma^2) 
            \frac{d_\ell}{d_{\ell-1}} 
            \lVert \Delta h^{(\ell -1)} \rVert_2^2
            \right).
        \end{equation}
   
       By induction, we can show that 
        \begin{equation*}            
        \begin{aligned}
            \lVert \Delta h^{(\ell)}\rVert_2^2
            & =   \mathcal{O} \left(
            d_\ell \times (2\gamma^2) (1+\sum_{j=1}^{\ell-1} (1+2\gamma^2)^j) 
            \right)\\
            & =   \mathcal{O} \left(
            d_\ell \times (2\gamma^2) (\sum_{j=0}^{\ell-1} (1+2\gamma^2)^j) 
            \right).
       \end{aligned}         
       \end{equation*}   
        
        We provide here the details of the recursion:
                
        $\bullet$ Initialization: For $\ell=1$, we have: 
        \begin{equation*}             
        \begin{aligned}
            \lVert \Delta h^{(1)}\rVert_2^2
            &= \mathcal{O} \left(
            2 \gamma^2 d_1 
            + (1+2\gamma^2) \frac{d_1}{d_{\text{in}}} \lVert \Delta x \rVert_2^2
            \right)\\
            &= \mathcal{O} \left(
            2 \gamma^2 d_1 
            \right).\\
       \end{aligned}         
       \end{equation*}
        
        \noindent$\bullet$ Induction step: 
        Let us suppose that $\lVert \Delta h^{(\ell)} \rVert_2^2 = \mathcal{O} \left(d_\ell \times (2\gamma^2) (1+\sum_{j=1}^{\ell-1}(1+2\gamma^2)^j) \right)$.
            
            Then:
            \begin{equation*}             
            \begin{aligned}
                \lVert \Delta h^{(\ell+1)} \rVert_2^2
                &= \mathcal{O} \left(
                2 \gamma^2 d_{\ell+1} 
                + (1+2\gamma^2) \frac{d_{\ell+1}}{d_{\ell}}
                \lVert \Delta h^{(\ell)} \rVert_2^2 \right) \quad \text{(from \Eqref{appdx_eq:recursion_delta_hl})}\\
                &= \mathcal{O} \left(
                2 \gamma^2 d_{\ell+1} 
                + (1+2\gamma^2) \frac{d_{\ell+1}}{d_{\ell}} \times d_\ell \times (2\gamma^2) (1+\sum_{j=1}^{\ell-1} (1+2\gamma^2)^j)
                \right)
                \\ &
                \quad \text{(heredity hypothesis)}
                \\
                &= \mathcal{O} \left(
                2 \gamma^2 d_{\ell+1} 
                + (1+2\gamma^2) d_{\ell+1} \times (2\gamma^2) (1+\sum_{j=1}^{\ell-1} (1+2\gamma^2)^j)
                \right)
                \\
                &=  \mathcal{O} \left(
                2 \gamma^2 d_{\ell+1} \times 
                \left(
                1 + (1+2\gamma^2)\times (1+\sum_{j=1}^{\ell-1}(1+2\gamma^2)^j)
                \right)
                \right)\\
                &=  \mathcal{O} \left(
                2 \gamma^2 d_{\ell+1} \times 
                \left(
                1 + (1+2\gamma^2)\times (\sum_{j=0}^{\ell-1}(1+2\gamma^2)^j)
                \right)
                \right)\\
                &=  \mathcal{O} \left(
                2 \gamma^2 d_{\ell+1} \times 
                \left(
                1 + (\sum_{j=0}^{\ell-1}(1+2\gamma^2)^{j+1})
                \right)
                \right)\\
                 &=  \mathcal{O} \left(
                2 \gamma^2 d_{\ell+1} \times 
                (1+\sum_{j=1}^{\ell}(1+2\gamma^2)^{j})
                \right),
           \end{aligned}         
           \end{equation*}
           which completes the proof by induction.
        
        Also, since $\lVert W^{(\ell)} \Delta \phi(h^{(\ell-1)})\rVert_2^2  
            + \lVert \Delta W^{(\ell)} \Delta \phi(h^{(\ell-1)})\rVert_2^2 \geq 0$, we have
        
        \begin{equation*}             
        \begin{aligned}
             \lVert \Delta h^{(\ell)}\rVert_2^2
            & = \Omega \left(
            2 \gamma^2 d_\ell
            \right).
       \end{aligned}         
       \end{equation*}
        
        Then, we have, by symmetry
        \begin{equation*}             
        \begin{aligned}
            \Delta {h_i^{(\ell)}}^2
            &= \frac{1}{d_\ell} \lVert \Delta h^{(\ell)}\rVert_2^2
            = \mathcal{O} \left(
            2\gamma^2 \sum_{j=0}^{\ell-1} (1+2\gamma^2)^j 
            \right). 
        \end{aligned}         
       \end{equation*}
       Thus,
       
        \begin{equation*}             
        \begin{aligned}
        \Delta {h_i^{(\ell)}}^2
            &= \frac{1}{d_\ell} \lVert \Delta h^{(\ell)}\rVert_2^2
            = \Omega \left(
            2 \gamma^2
            \right).
       \end{aligned}         
       \end{equation*}
        
        Therefore, we have the maximal update condition since $\Delta {h_i^{(\ell)}}^2 = \Theta(1)$ with respect to the width of the layer $d_\ell$. Also, the optimal learning rate satisfying this condition is independent of the hidden-layer widths. 
        
        \end{proof}

 \subsection{Proof from Section \ref{sec:width_wise_hp_transfer_for_projected_scion}}
\label{appdx_proof:width_transfer_projected_scion_interval}
   Before proving Proposition \ref{prop:width_projected_scion}, we first establish the following lemma.
    
\begin{figure}[ht!]
    \centering
    \begin{subfigure}{0.32\linewidth}
        \centering
        \includegraphics[width=\linewidth]{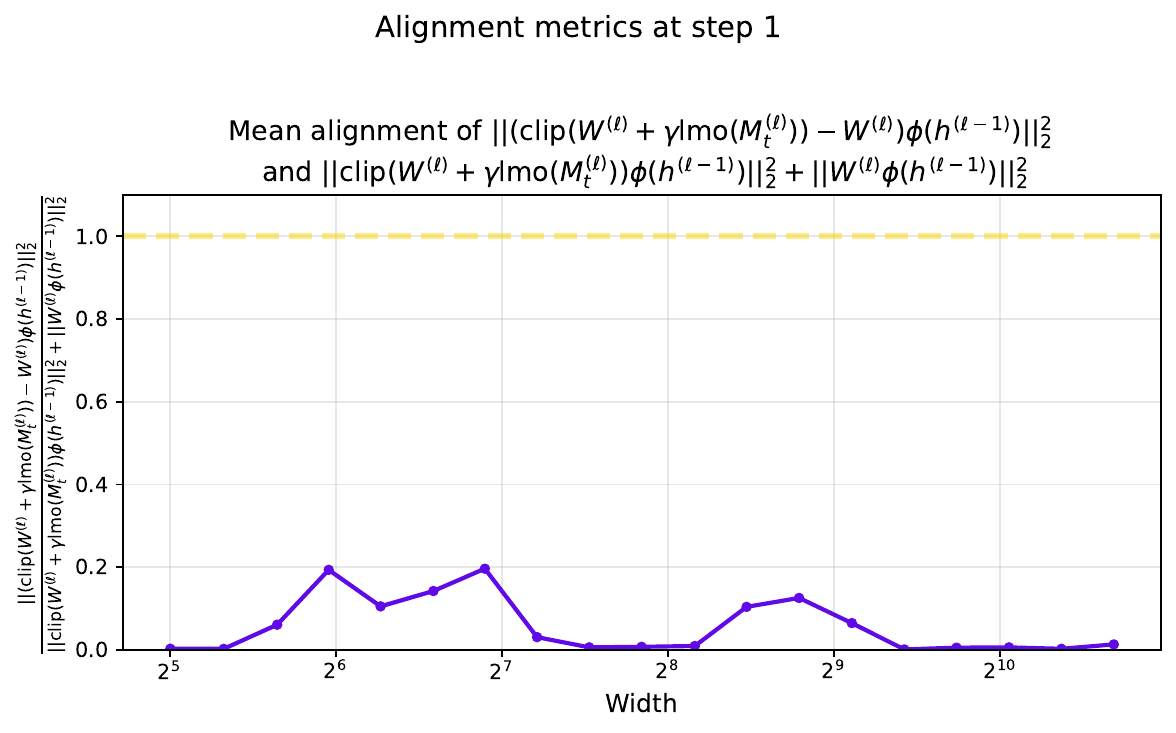}
    \end{subfigure}
    \hfill
    \begin{subfigure}{0.32\linewidth}
        \centering
        \includegraphics[width=\linewidth]{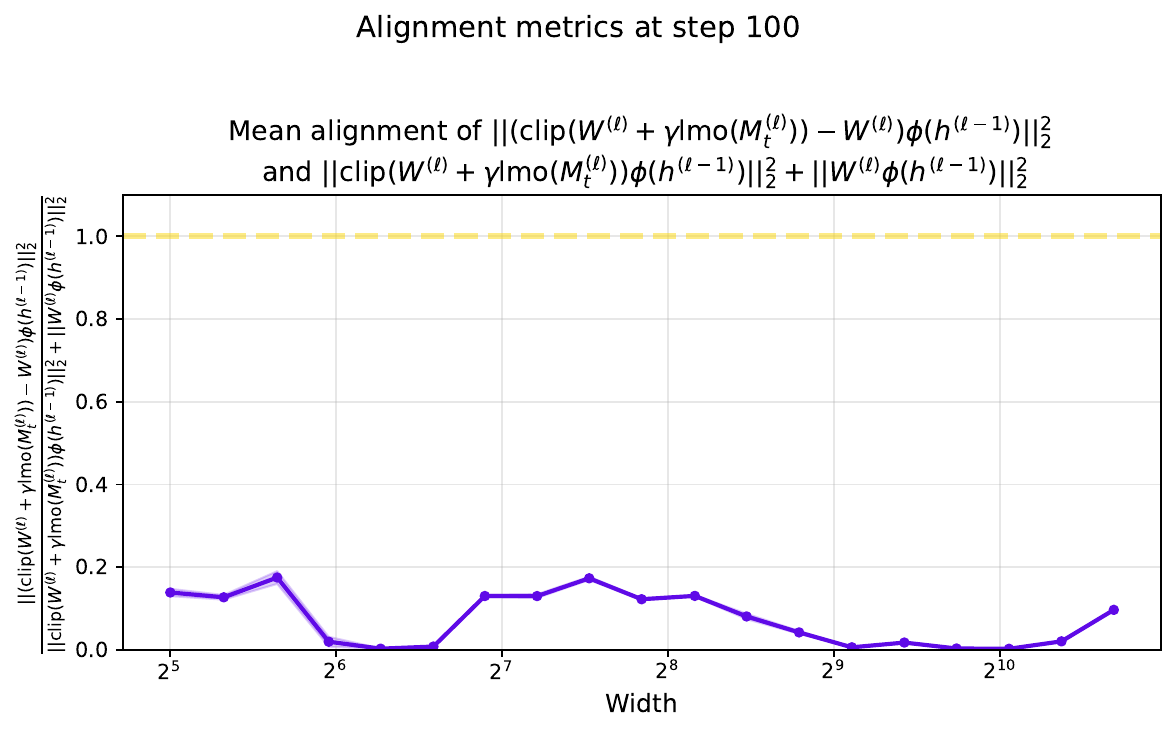}
    \end{subfigure}
    \hfill
    \begin{subfigure}{0.32\linewidth}
        \centering
        \includegraphics[width=\linewidth]{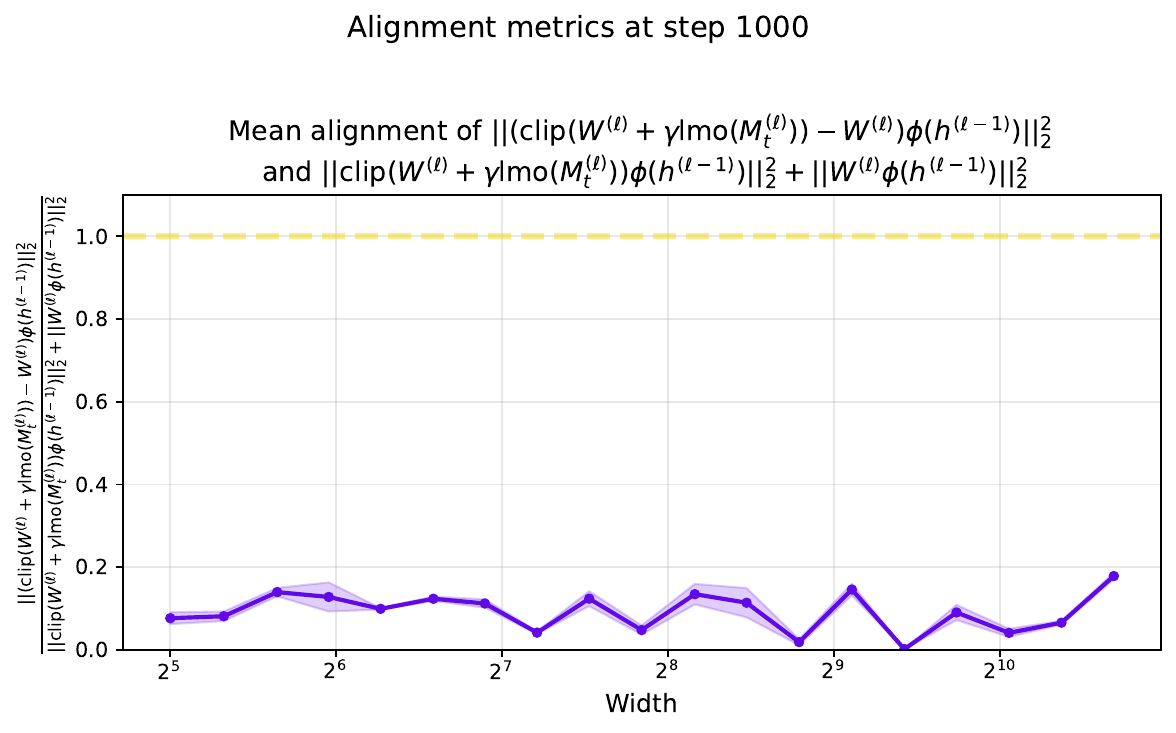}
    \end{subfigure}
    \caption{
     Cancellation of the clipping update of $W^{(\ell)}\phi(h^{(\ell-1)})$ and the current features.
    The network is an MLP trained with \pScion on CIFAR-10, over 5 trials. The layer of interest is the second hidden layer. The plots are shown at different training steps and across a wide range of network widths.
    Experimental details can be found in Table \ref{tab:mlp_hp_alignment}.}
    \label{fig:cancellation_sum_update_with_projected_scion}
\end{figure}

    \begin{lemma}[Norm control]
    \label{lemma:norm_control}
        For every vector $v$ and matrix $A$ with the clipping interval $[a,b]$ fixed independently of the width, we have
        \begin{equation}
        \label{appdx_eq:clipping_bilipshitz}
            \lVert \spectralclip_{[a,b]} (A) v \rVert_2^2 = \Theta(\lVert v \rVert_2^2).
        \end{equation}
    \end{lemma}
    \begin{proof}
         Let $M= \spectralclip_{[a,b]}(A)$. Let us denote $A=U\Sigma V^\top$ and $M=U \Lambda V^\top$ the singular value decompositions of $A$ and $M$ respectively with $U, V$ orthogonal, $\Sigma=\text{diag}(\sigma_i)$ the diagonal matrix containing the singular values of $A$ and its clipped version $\Lambda=\text{diag}(\lambda_i)$ with $\lambda_i = \text{clip}_{[a, b]} \sigma_i$. 
         Let us also denote $w=V^\top v$.
        Then, 
        \begin{equation}
            \begin{aligned}
                \lVert \Lambda w \rVert_2^2 &= \sum_i \lambda_i^2 w_i^2 \\
                \text{ and }  a^2 \lVert w \rVert_2^2 & \leq \lVert \Lambda w \rVert_2^2 \leq   b^2 \lVert w \rVert_2^2. \\
            \end{aligned}
        \end{equation}
        Since $\lVert w \rVert_2^2 = \lVert v \rVert_2^2$, we have
        $a^2 \lVert v \rVert_2^2 \leq \lVert \Lambda V^\top v \rVert_2^2 \leq b^2 \lVert v \rVert_2^2$.
        Then, by isometry of the orthogonal matrix $U$, we can conclude with
        \begin{equation}
            \begin{aligned}
                & a^2 \lVert v \rVert_2^2 \leq \lVert U \Lambda V^\top v \rVert_2^2 \leq b^2 \lVert v \rVert_2^2 \\
                \text{i.e.} \quad & a^2 \lVert v \rVert_2^2 \leq \lVert \spectralclip_{[a,b]}(A) v \rVert_2^2 \leq b^2 \lVert v \rVert_2^2. \\
            \end{aligned}
        \end{equation}
    \end{proof}
    \ProjectedScionWidth*
    \begin{proof}
    With the projected version of \Scion on an interval $[a, b]$, the update can be written as
        \begin{equation}
            \begin{aligned}
            W_{t+1}^{(\ell)} = \spectralclip_{[a,b]}(W_t^{(\ell)} + \gamma \lmo[\nabla_{W_t^{(\ell)}}\mathcal{L}]).
            \end{aligned}
        \end{equation} 
        Moreover, by definition of the spectral clipping, if $a$ and $b$ are fixed independently of the width (meaning $a,b = \Theta(1)$), we have, for any matrix $A$,
        \begin{equation}
            \lVert \spectralclip_{[a,b]}(A) \rVert_{\mathcal{S}_\infty}^2 = \Theta(1).
        \end{equation}
        
        In the following, we cannot apply this argument as it is to matrices for which we are interested in their scaling with the width. As the width scales, the spectral norm of the matrix could vanish, and, depending on the choice of $a$, we have no guarantee that the spectral norm will remain of constant order with respect to the width. Therefore, we will use:
        \begin{equation}
            \lVert \spectralclip_{[a,b]}(A) \rVert_{\mathcal{S}_\infty}^2 = \mathcal{O}(1).
        \end{equation}
     As in the derivations with the \Scion update from Appendix \ref{appdx-proof:width_transfer_scion}, we want to evaluate
        \begin{equation}
        \label{appdx_eq:delta_h_l_projeted_interval}
            \lVert \Delta h^{(\ell)}\rVert_2^2 
            = \Theta \left( \lVert \Delta W ^{(\ell)} \phi(h^{(\ell-1)})\rVert_2^2 
             + \lVert W^{(\ell)}  \Delta \phi(h^{(\ell-1)})\rVert_2^2 
             + \lVert \Delta W^{(\ell)}  \Delta \phi(h^{(\ell-1)})\rVert_2^2
             \right).
        \end{equation}
    The second term can be computed straightforwardly to end up with 
    \begin{equation}
    \label{appdx_eq:second_term_bound}
        \lVert W^{(\ell)}  \Delta \phi(h^{(\ell-1)})\rVert_2^2  
        = \mathcal{O} \left( \frac{d_\ell}{d_{\ell-1}} \lVert \Delta h^{(\ell-1)}\rVert_2^2 \right).
    \end{equation}
    For the first term, we have
    \begin{equation}
        \begin{aligned}
            \lVert \Delta W ^{(\ell)} \phi(h^{(\ell-1)})\rVert_2^2 
            &= \lVert  (\spectralclip_{[a,b]}(W^{(\ell)} + \gamma \lmo(\nabla_{W^{(\ell)}} \mathcal{L})) - W^{(\ell)} )\phi(h^{(\ell-1)})\rVert_2^2 \\
            &= \mathcal{O} \left( \lVert  \spectralclip_{[a,b]}(W^{(\ell)} + \gamma \lmo(\nabla_{W^{(\ell)}} \mathcal{L})) \phi(h^{(\ell-1)})\rVert_2^2 + \lVert W^{(\ell)} \phi(h^{(\ell-1)})\rVert_2^2 \right) 
            \\
             & \quad \text{from \Eqref{eq:init_update_big_o_width_spectral}} \\
            & = \mathcal{O} \left( \lVert \phi(h^{(\ell-1)})\rVert_2^2 + \lVert W^{(\ell)} \phi(h^{(\ell-1)})\rVert_2^2 \right) \\
            & \quad \text{(from \Eqref{appdx_eq:clipping_bilipshitz})} \\
            & = \mathcal{O} \left( \lVert h^{(\ell-1)}\rVert_2^2 + \lVert W^{(\ell)} \rVert_{\mathcal{S}_\infty}^2 \lVert\phi(h^{(\ell-1)})\rVert_2^2 \right) \\
            & \quad \text{(scale of activation function \Eqref{eq:hyp_scale_activation_function} and  alignment assumption \Eqref{eq:hyp_alignment})} \\
            & = \mathcal{O}(d_{\ell-1} + \frac{d_\ell}{d_{\ell-1}} d_{\ell-1}) \\
            & = \mathcal{O}(d_{\ell-1} + d_{\ell}).
        \end{aligned}
    \end{equation}
    For the third term, we have
    \begin{equation}         
        \begin{aligned}
        \left\lVert \Delta W^{(\ell)} \Delta \phi( h^{(\ell-1)} ) \right \rVert_2^2
        &= \left \lVert 
        \spectralclip_{[a,b]} \left(W_t^{(\ell)} 
        + \gamma \lmo(\nabla_{W_t^{(\ell)}}\mathcal{L}) \right) 
        \Delta \phi( h^{(\ell-1)} ) 
        - W_t^{(\ell)} \Delta \phi( h^{(\ell-1)} ) 
        \right \rVert_2^2 \\
        &= \mathcal{O}  \left(
        \left \lVert
        \spectralclip_{[a,b]} \left(W_t^{(\ell)} 
        + \gamma \lmo(\nabla_{W_t^{(\ell)}}\mathcal{L}) \right) 
        \Delta \phi( h^{(\ell-1)} )\right \rVert_2^2  
        + \left \lVert W_t^{(\ell)} \Delta \phi( h^{(\ell-1)} ) 
        \right \rVert_2^2 
        \right) \\
        & \quad \text{from \Eqref{eq:init_update_big_o_width_spectral}} \\
        &= \mathcal{O}  \left(
        \left \lVert
        \Delta \phi( h^{(\ell-1)} )\right \rVert_2^2  
        + \left \lVert W_t^{(\ell)} \Delta \phi( h^{(\ell-1)} ) 
        \right \rVert_2^2 
        \right) \\
        & \quad \text{(from \Eqref{appdx_eq:clipping_bilipshitz})} \\
        &= \mathcal{O}\left( 
        (1+\frac{d_\ell}{d_{\ell-1}}) \lVert \Delta h^{(\ell-1)} \rVert_2^2
        \right).\\
        & \quad \text{(from the Lipschitz continuity of $\phi$ and we re-use the bound of \Eqref{appdx_eq:second_term_bound})}.
        \end{aligned}        
    \end{equation} 
    Therefore,
    \begin{equation}
        \begin{aligned}
            \lVert \Delta h^{(\ell)} \rVert_2^2 &= \mathcal{O}\left( d_\ell + d_{\ell-1} + (1+\frac{d_\ell}{d_{\ell-1}} ) \lVert \Delta h^{(\ell-1)} \rVert_2^2 \right).\\
        \end{aligned}
    \end{equation}
    With the assumption that $d_\ell$ and $d_{\ell-1}$ are of the same order,
    and by induction, we have
    \begin{equation}
        \begin{aligned}
            \lVert \Delta h^{(\ell)} \rVert_2^2 = \mathcal{O}(d_\ell).
        \end{aligned}
    \end{equation}
       
\end{proof}
    Proposition~\ref{prop:width_projected_scion} establishes an
    upper bound on the feature updates but lacks a lower bound. The lower bound argument used for \Scion without projections relies on a non-cancellation assumption that does not hold after spectral clipping.
    In particular, at some point of the proof, we would need to have the following assumption
    \begin{equation}
    \begin{aligned}
        & \left\lVert \left(\spectralclip_{[a,b]} (W^{(\ell)} + \gamma \lmo(M_t^{(\ell)})) - W^{(\ell)}\right) \phi(h^{(\ell -1)}) \right\rVert_2^2 \\ = 
        &\quad  \Theta \left( \lVert \spectralclip_{[a,b]} (W^{(\ell)} + \gamma \lmo(M_t^{(\ell)})) \phi(h^{(\ell -1)}) \rVert_2^2 + \lVert W^{(\ell)} \phi(h^{(\ell -1)}) \rVert_2^2 \right),
     \end{aligned}
    \end{equation}
    which does not hold in practice (see for instance Figure \ref{fig:cancellation_sum_update_with_projected_scion}). Indeed, if the learning rate is too small, the left term cancels and the lower bound does not hold.
    Therefore, we only have:
    \begin{equation}
    \label{eq:init_update_big_o_width_spectral}
    \begin{aligned}
        & \left\lVert \left(\spectralclip_{[a,b]} (W^{(\ell)} + \gamma \lmo(M_t^{(\ell)})) - W^{(\ell)}\right) \phi(h^{(\ell -1)}) \right\rVert_2^2 \\ 
        = & \quad  \mathcal{O} \left( \lVert \spectralclip_{[a,b]} (W^{(\ell)} + \gamma \lmo(M_t^{(\ell)})) \phi(h^{(\ell -1)}) \rVert_2^2 + \lVert W^{(\ell)} \phi(h^{(\ell -1)}) \rVert_2^2 \right).
     \end{aligned}
    \end{equation}

\section{CompleteP style analysis}
\label{appdx_sec:completeP_style_analysis}
    In this part, we analyze the scaling behavior of the network's hidden states using the heuristics found in \cite{dey2026dontlazycompletepenables}. 
    Hence, the $Z^{(\ell)}$ matrices are not learned and are fixed as identity. The hidden-layer width is constant, and training is restricted to the $W^{(\ell)}$ operators.

    \paragraph{CompleteP assumptions}
    Following the analysis of \cite{dey2026dontlazycompletepenables}, we will assume that for every step $t$, hidden representations from different iterations have constant-order correlations, normalized gradients remain strongly aligned across layers, and gradients from different optimization steps are asymptotically uncorrelated.
    That is,
    \begin{equation}
    \label{appdx_eq:state_correlation}
        \begin{aligned}
            \phi(h_t^{(\ell)})^\top  \phi(h_{t'}^{(\ell)}) = \Theta(d_\ell) \qquad \forall \ell, \forall(t, t'),
        \end{aligned}
    \end{equation}
    and, denoting $g_t^{(\ell)}=\frac{\partial \mathcal{L}}{\partial h^{(\ell)}_t}$,
    \begin{equation}
    \label{appdx_eq:normalized_grad_correlation_complete_p}
        \begin{aligned}
            \left \langle \frac{g_t^{(\ell)}}{\lVert g_t^{(\ell)} \rVert}, \frac{g_t^{(k)}}{\lVert g_t^{(k)} \rVert} \right\rangle = \Theta(1) \qquad \forall (\ell, k), \forall t,
        \end{aligned}
    \end{equation}
    while gradients from different steps are uncorrelated.

    \begin{proposition}
        For all iterations $t > 0$, under the CompleteP correlation assumptions, the normalized squared $\ell_2$-norm change of the final layer hidden state satisfies:
        \begin{equation*}
            \frac{1}{d_L}\| \Delta h^{(L)}_t\|_2^2 = \Theta(1).
        \end{equation*}
    \end{proposition}
    
    \begin{proof}
        
    Let $t>0$ and write the definition of $h^{(L)}(t)$ to get
    \begin{equation*}
        h^{(L)}_t = h^{(0)}_t + L^{-\alpha}\sum\limits_{\ell=0}^{L-1}W^{(\ell)}_t\phi(h^{(\ell)}_t).
    \end{equation*}
    Inserting $W^{(\ell)}_t = W^{(\ell)}_0 + \sum\limits_{t< t'}\Delta W^{(\ell)}_{t'}$ (unrolling the recursion) then gives
    \begin{equation*}
        h^{(L)}_t = h^{(0)}_t + L^{-\alpha}\sum\limits_{\ell=0}^{L-1}W^{(\ell)}_0\phi(h^{(\ell)}_t) + \underbrace{L^{-\alpha}\sum\limits_{\ell=0}^{L-1}\sum\limits_{t'<t}\Delta W^{(\ell)}_{t'}\phi(h^{(\ell)}_t)}_{{\Delta h^{(L)}_{t}}}.
    \end{equation*}
    In CompleteP Desideratum 2 \citep{dey2026dontlazycompletepenables}, they aim to show that $\|\Delta h^{(L)}_t\|_{RMS}^2 = \Theta(1)$ for all $t>0$.
    Using \uScion, the update has the form 
    \begin{equation*}
        \Delta W^{(\ell)}_{t'} = W^{(\ell)}_{t'+1}-W^{(\ell)}_{t'}=\gamma \frac{g^{(\ell+1)}_{t'}}{\|g^{(\ell+1)}_{t'}\|_2}\left[\frac{\phi(h^{(\ell)}_{t'})}{\|\phi(h^{(\ell)}_{t'})\|_2}\right]^\top.
    \end{equation*}
    Plugging this into the expression for $\Delta h^{(L)}_{t}$, we find
    \begin{equation*}
        \Delta h^{(L)}_t = L^{-\alpha}\sum\limits_{\ell=0}^{L-1}\sum\limits_{t'<t}\gamma \frac{g^{(\ell+1)}_{t'}}{\|g^{(\ell+1)}_{t'}\|_2}\left[\frac{\phi(h^{(\ell)}_{t'})}{\|\phi(h^{(\ell)}_{t'})\|_2}\right]^\top\phi(h^{(\ell)}_{t}).
    \end{equation*}
    Note that, for each $\ell$, CompleteP claims that the coordinates of $\phi(h^{(\ell)}_{t})$ and $\phi(h^{(\ell)}_{t'})$ have $\Theta(1)$ correlation, so that $\phi(h^{(\ell)}_{t'})^\top\phi(h^{(\ell)}_{t}) = \Theta(d_\ell)$, while $\|\phi(h^{(\ell)}(t'))\|_2 = \Theta(\sqrt{d_\ell})$. We also make this correlation assumption here (\Eqref{appdx_eq:state_correlation}), and, by taking the norm squared and applying this yields
    \begin{equation*}
        \begin{aligned}
            \|\Delta h^{(L)}_t\|_2^2 &= L^{-2\alpha}\left\|\sum\limits_{\ell=0}^{L-1}\sum\limits_{t'<t}\gamma \frac{g^{(\ell+1)}_{t'}}{\|g^{(\ell+1)}_{t'}\|_2}\left[\frac{\phi(h^{(\ell)}_{t'})}{\|\phi(h^{(\ell)}_{t'})\|_2}\right]^\top\phi(h^{(\ell)}_t)\right\|_2^2\\
            &= L^{-2\alpha}\gamma^2d_L\left\|\sum\limits_{\ell=0}^{L-1}\sum\limits_{t'<t} \frac{g^{(\ell+1)}_{t'}}{\|g^{(\ell+1)}_{t'}\|_2}\right\|_2^2\\
            &= L^{-2\alpha}\gamma^2d_L\Theta(L^2)
        \end{aligned}
    \end{equation*}
    where the last equality follows because the layer gradients are strongly correlated across layers but uncorrelated across time steps (\Eqref{appdx_eq:normalized_grad_correlation_complete_p}).

    Hence, with $\gamma = \gamma_{\text{base}} L^{\alpha-1}$, we have $\lVert \Delta h_t^{(L)} \rVert_2^2=\Theta(d_L)$, and therefore $\frac{1}{d_L} \lVert \Delta h_t^{(L)} \rVert_2^2=\Theta(1)$.

    Note that for a sign update with the CompleteP scaling \citep{dey2026dontlazycompletepenables}, we would get $1/L^2$ for the RMS norm squared of the update.

    \end{proof}

\section{Coordinate Checks on nanoGPT}
    \begin{figure}[!ht]
        \includegraphics[width=\linewidth]{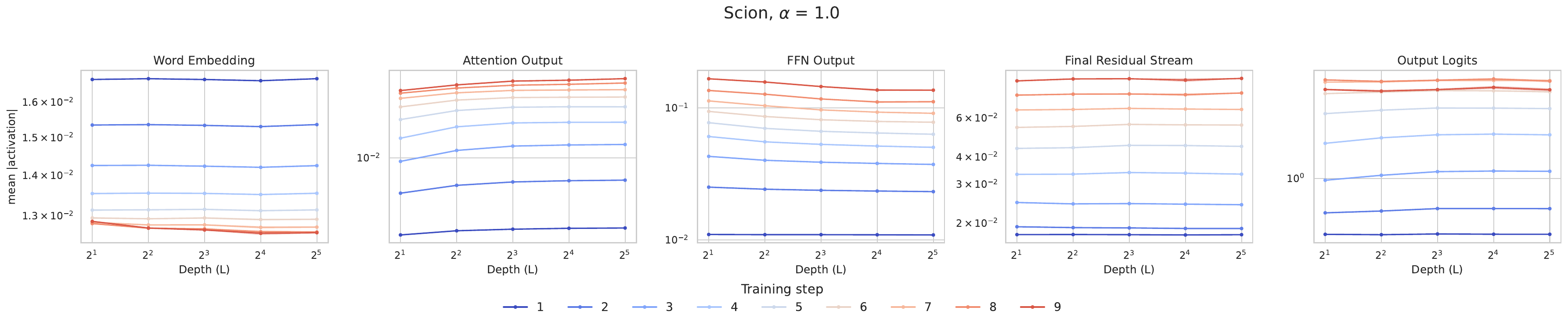}
        \caption{Coordinate check for \Scion, on FineWeb, with $\alpha=1.0$, and base learning rate $\gamma_{\text{base}}=2^{-11}$.}
        \label{fig:nanogpt_coord_check_scion_alpha1}
    \end{figure}

    \begin{figure}[!ht]
         \includegraphics[width=\linewidth]{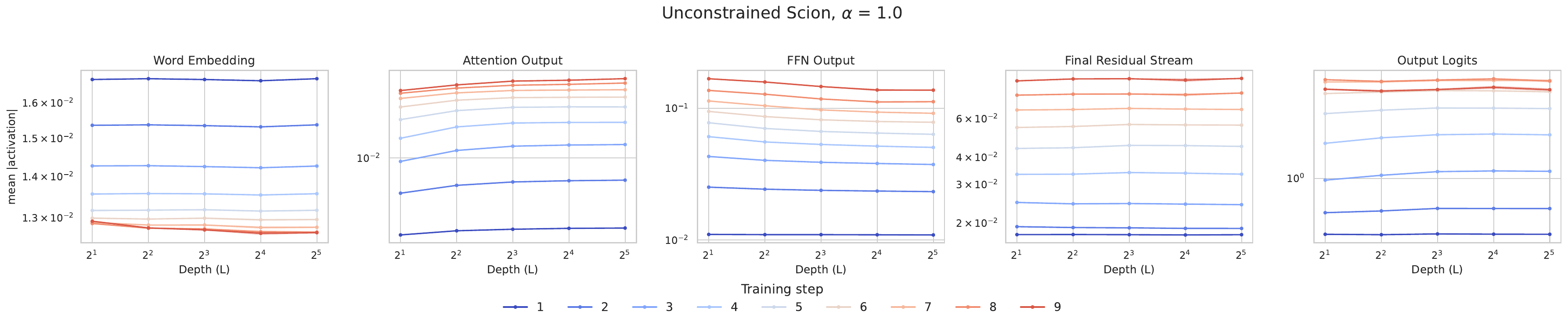}
        \caption{Coordinate check for \uScion, on FineWeb, with $\alpha=1.0$, and base learning rate $\gamma_{\text{base}}=2^{-11}$.}
        \label{fig:nanogpt_coord_check_uscion_alpha1}
    \end{figure}

    \begin{figure}[!ht]
         \includegraphics[width=\linewidth]{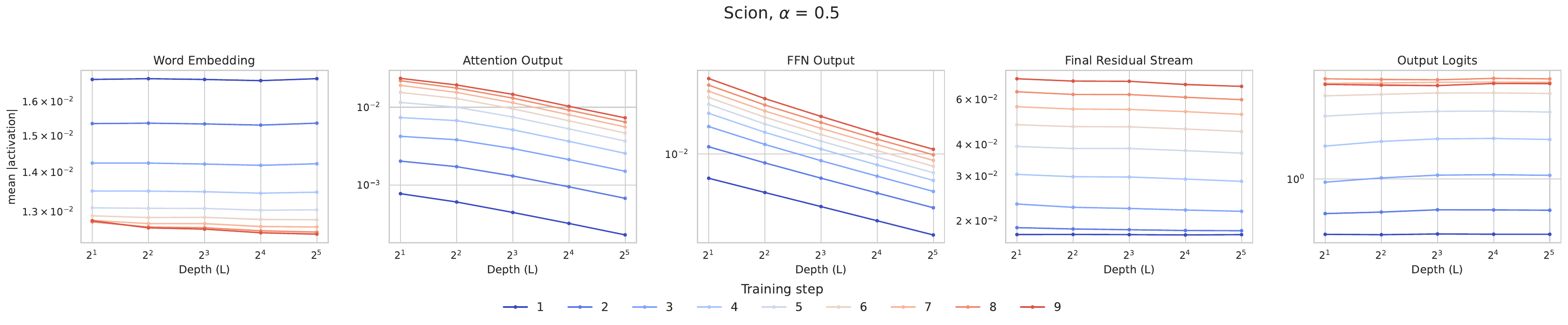}
        \caption{Coordinate check for \Scion, on FineWeb, with $\alpha=0.5$, and base learning rate $\gamma_{\text{base}}=2^{-11}$.}
        \label{fig:nanogpt_coord_check_scion_alpha05}
    \end{figure}

    \begin{figure}[!ht]
         \includegraphics[width=\linewidth]{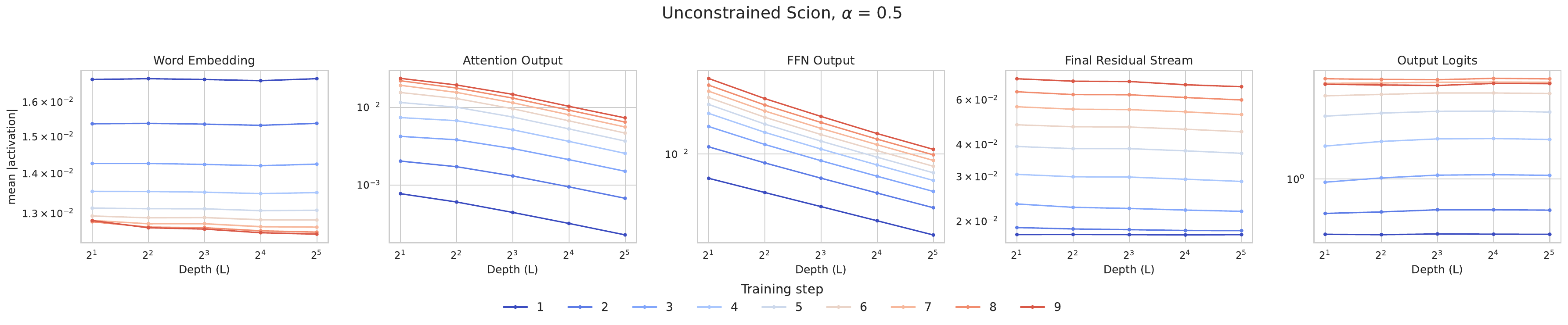}
        \caption{Coordinate check for \uScion, on FineWeb, with $\alpha=0.5$, and base learning rate $\gamma_{\text{base}}=2^{-11}$.}
        \label{fig:nanogpt_coord_check_uscion_alpha05}
    \end{figure}

\clearpage

\section{Additional Experiments}

\subsection{NanoGPT with Z}
\label{sec:nanogpt_with_Z}
\begin{figure}[ht!]
    \centering
    \includegraphics[width=\linewidth]{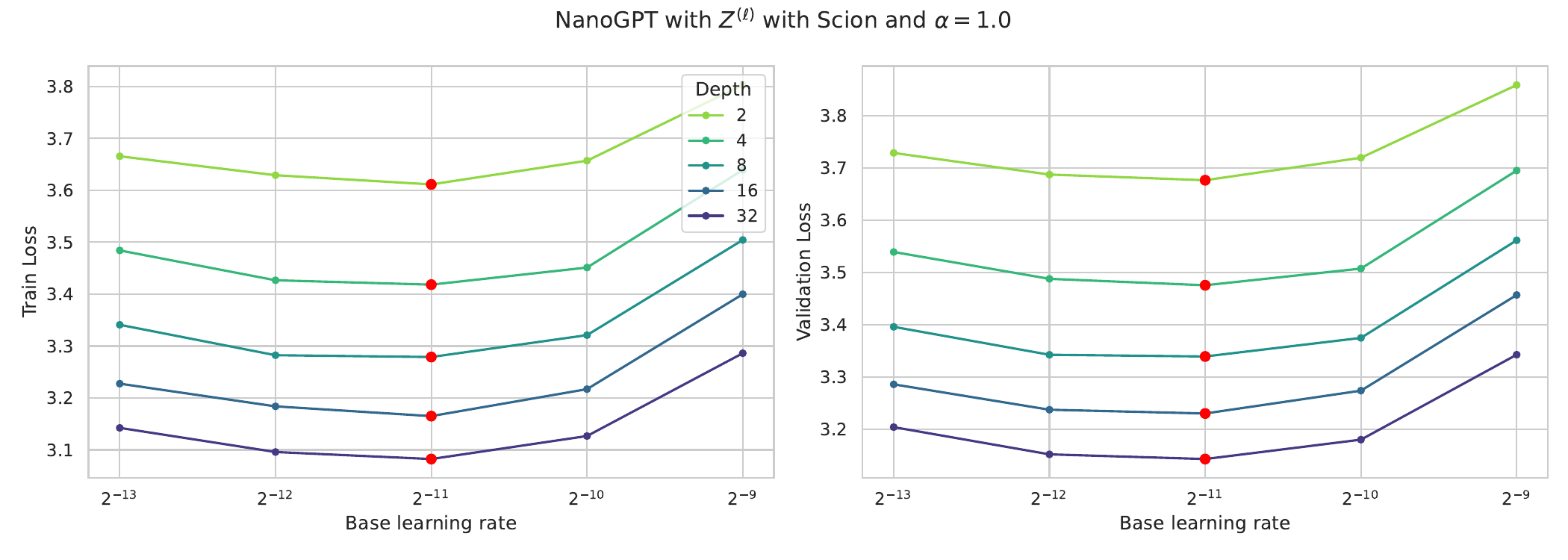}
    \caption{
    \Scion on nanoGPT with $Z^{(\ell)}$ trained on FineWeb, with $\alpha=1$, on 1 seed.
    Well depth-aligned and parallel loss curves indicate that the optimal base learning rate remains invariant with depth.
    The experimental details are the same as in Table \ref{tab:NanoGPT_hp}, 
    and the $Z^{(\ell)}$ matrices are therefore of shape $768 \times 768$ for every hidden layer $\ell$.
    }
    \label{fig:moddednanogpt_with_Z}
\end{figure}

To further evaluate the proposed depth-wise learning rate scaling developed in the analysis of Section \ref{sec:complete_scion} when the matrices $Z^{(\ell)}$ are trained, we repeat the nanoGPT experiments while learning $Z^{(\ell)}$ in every hidden layer. As shown in Figure \ref{fig:moddednanogpt_with_Z}, the loss curves remain approximately parallel and their minima occur at similar base learning rates across depths, providing empirical evidence that the proposed learning rate scaling transfers when $Z^{(\ell)}$ is learned.

\subsection{NanoGPT without any scale of the learning rates}
\begin{figure}[ht!]
    \centering
    \begin{subfigure}{0.48\linewidth}
    \includegraphics[width=\linewidth]{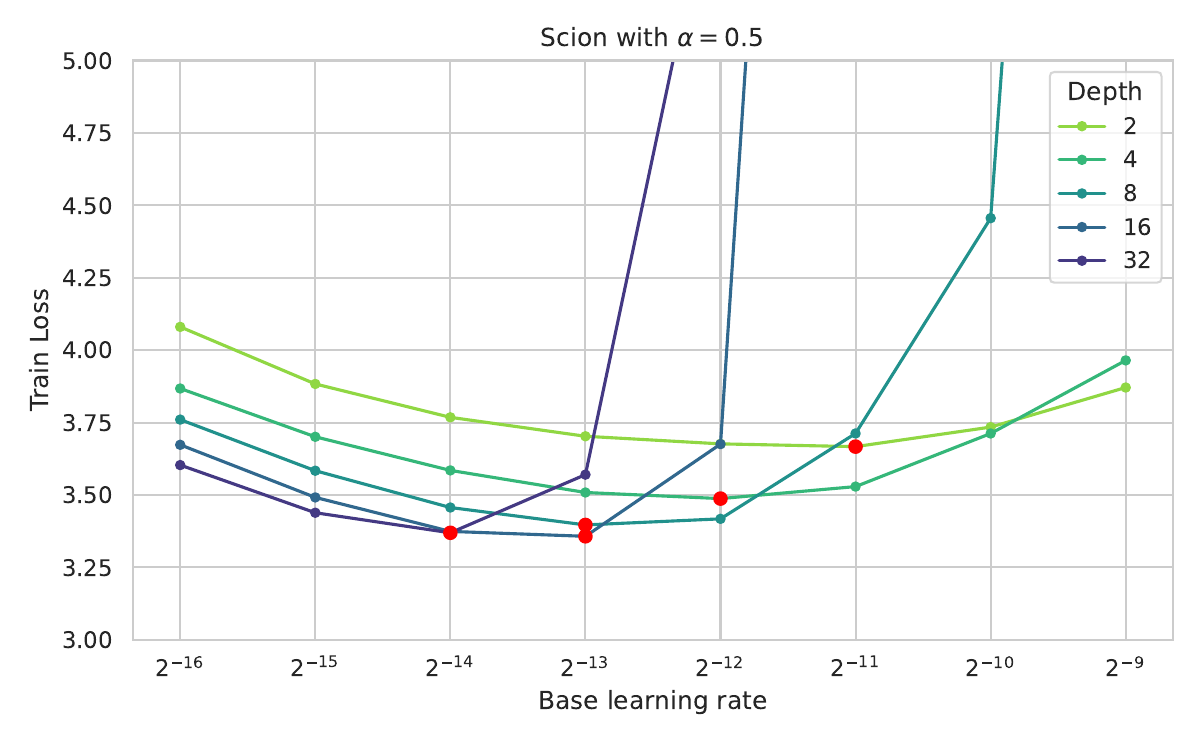}
    \end{subfigure}
    \hfill
     \begin{subfigure}{0.48\linewidth}
    \includegraphics[width=\linewidth]{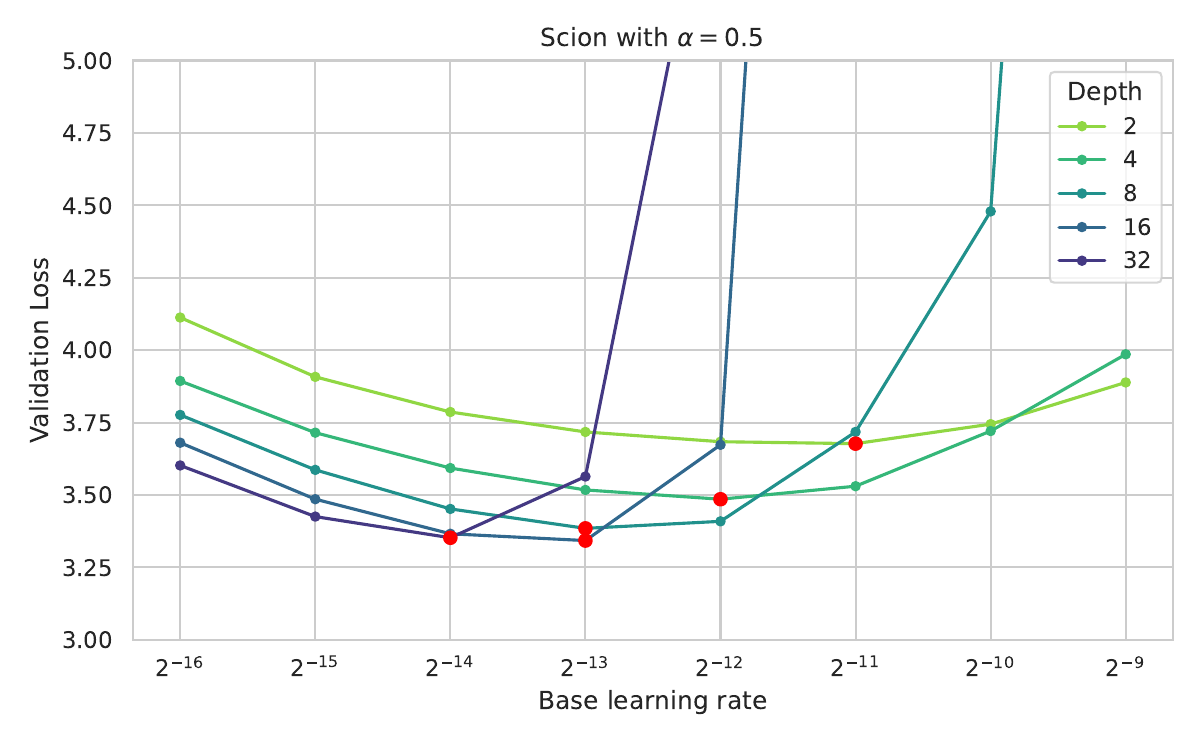}
    \end{subfigure}
    \caption{
    \Scion on nanoGPT with $Z^{(\ell)}$ trained on FineWeb, with $\alpha=0.5$, without any scaling of the learning rates ($\gamma=\gamma_{\text{base}}$ and $\gamma_Z={\gamma_Z}_{\text{base}}$), on 1 seed.
    We observe the shift of the optimal learning rate as depth increases.
    The experimental details are in Table \ref{tab:NanoGPT_hp}.
    }
    \label{fig:moddednanogpt_with_Z_no_lr_factor}
\end{figure}

To assess whether the depth-dependent learning rate scaling is necessary for transfer, we repeat the nanoGPT depth sweep without applying any depth-dependent correction. The results are displayed in Figure \ref{fig:moddednanogpt_with_Z_no_lr_factor}, where we observe that the optimal learning rate shifts as the depth increases, in contrast to the approximately depth-invariant optimum obtained with the proposed scaling. This empirically shows that removing the prescribed learning rate scalings leads to a loss of depth-wise transfer.
We test $\alpha=1/2$ since, for $\alpha=1$, the scaling factor is just $1$.

\subsection{Variability across seeds}
\begin{figure}[ht!]
    \centering
    \begin{subfigure}{0.48\linewidth}
    \includegraphics[width=\linewidth]{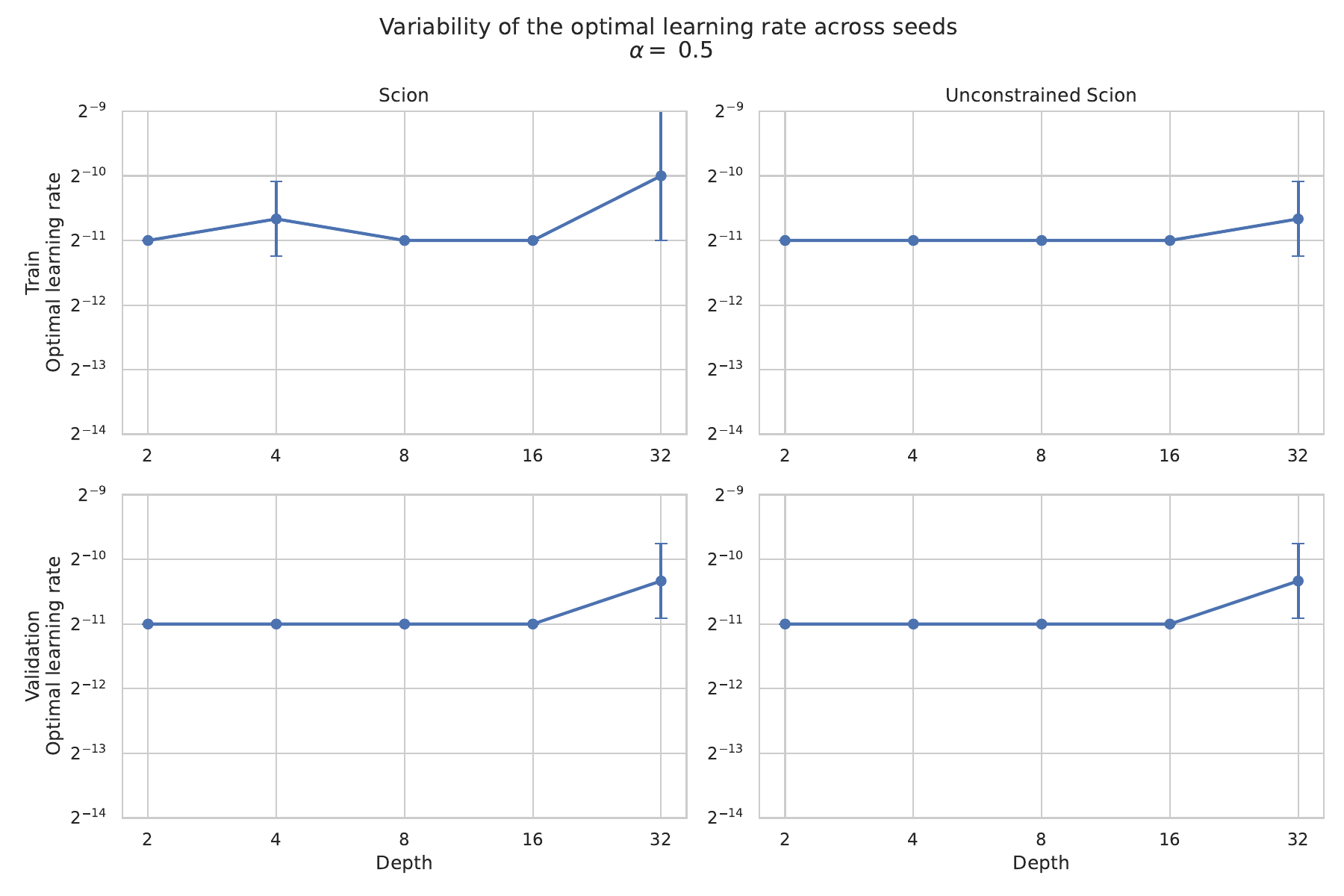}
    \end{subfigure}
    \hfill
     \begin{subfigure}{0.48\linewidth}
    \includegraphics[width=\linewidth]{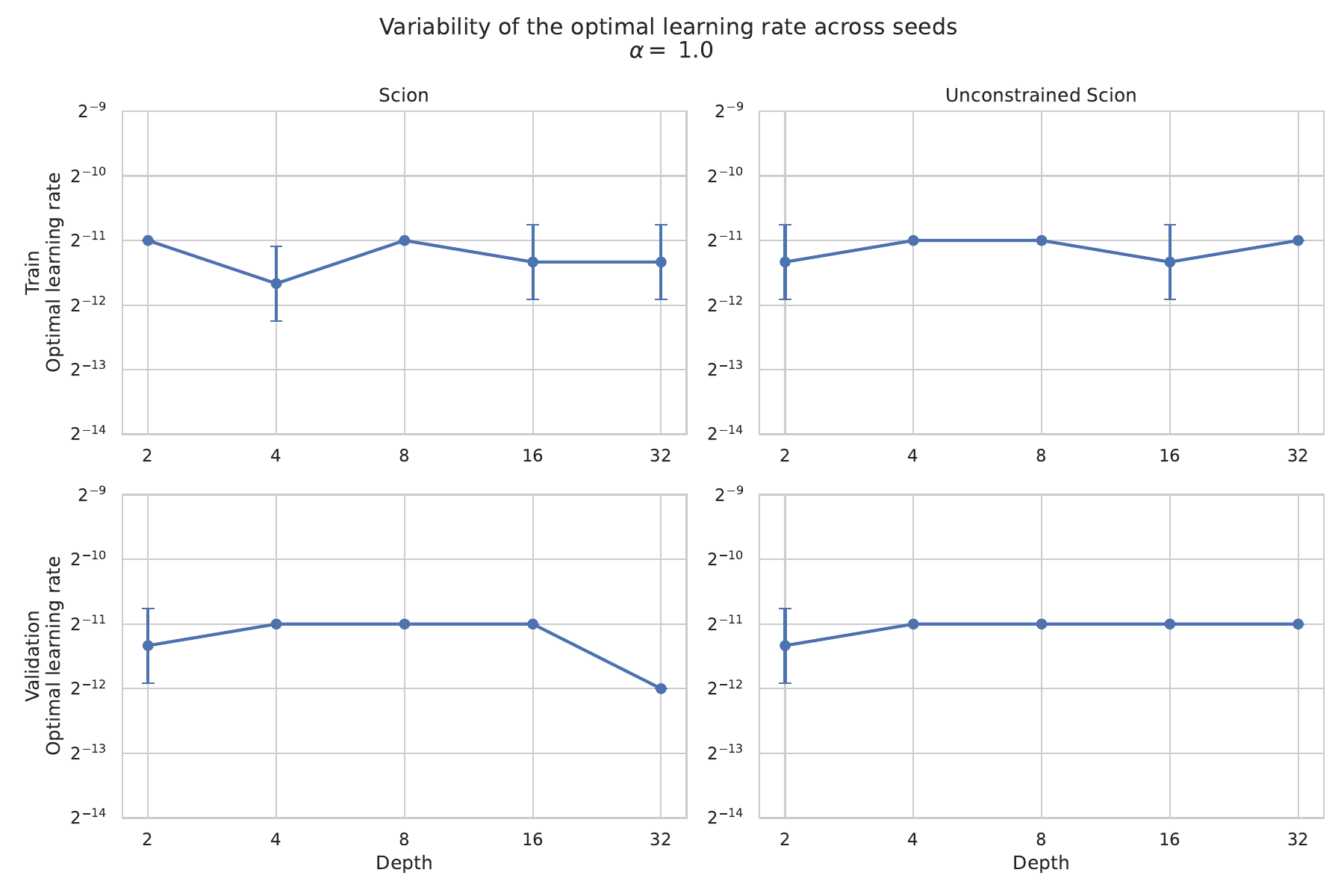}
    \end{subfigure}
    \caption{
    Variability across seeds of the optimal learning rates per depths with \Scion and \uScion for $\alpha=0.5$ (\textit{left plots}) and $\alpha=1$ (\textit{right plots}) on nanoGPT trained on FineWeb.
    The experimental details are in Table \ref{tab:NanoGPT_hp}.
    }
    \label{fig:error_bars_nanogpt}
\end{figure}

\begin{figure}[ht!]
    \centering
    \begin{subfigure}{0.48\linewidth}
    \includegraphics[width=\linewidth]{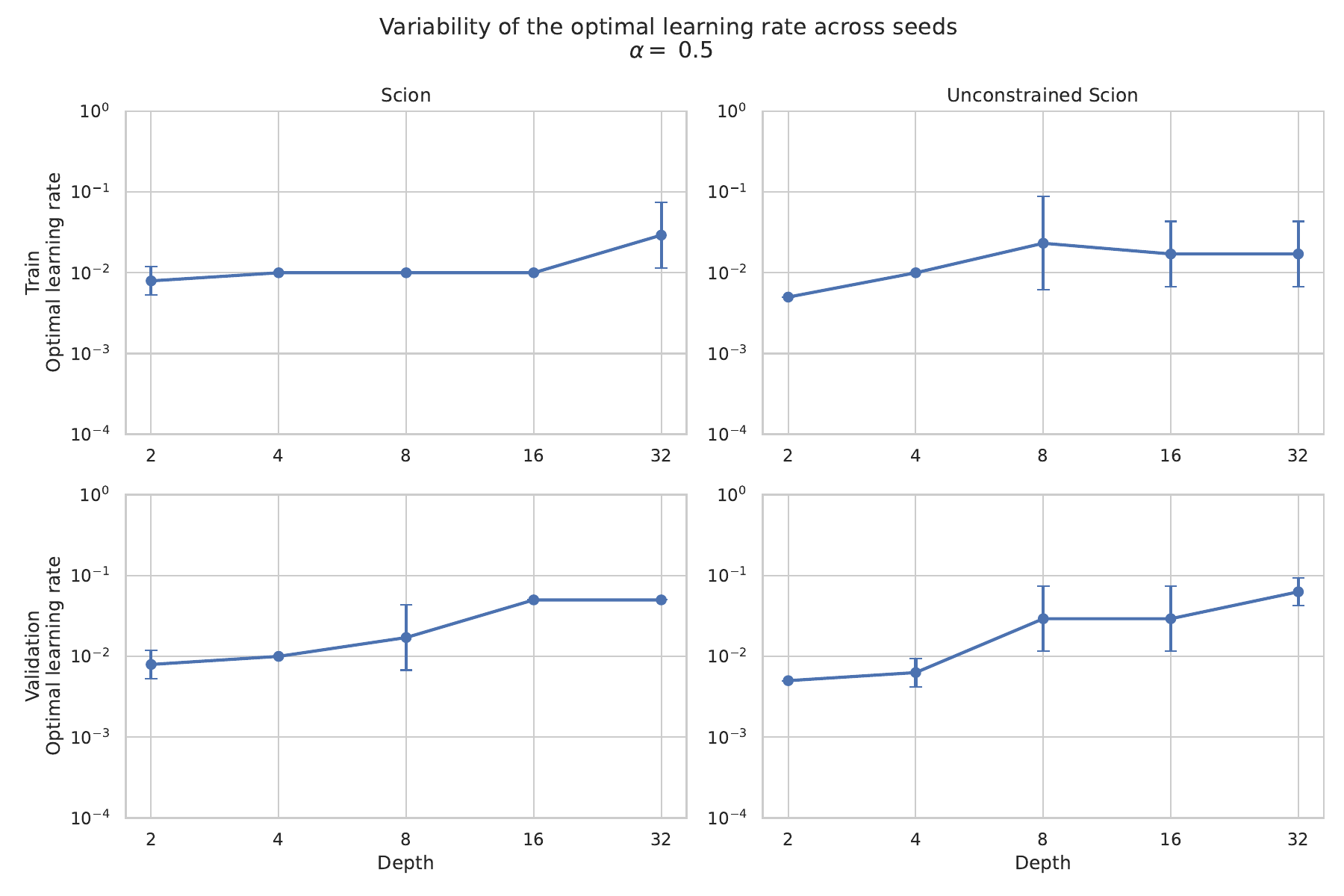}
    \end{subfigure}
    \hfill
     \begin{subfigure}{0.48\linewidth}
    \includegraphics[width=\linewidth]{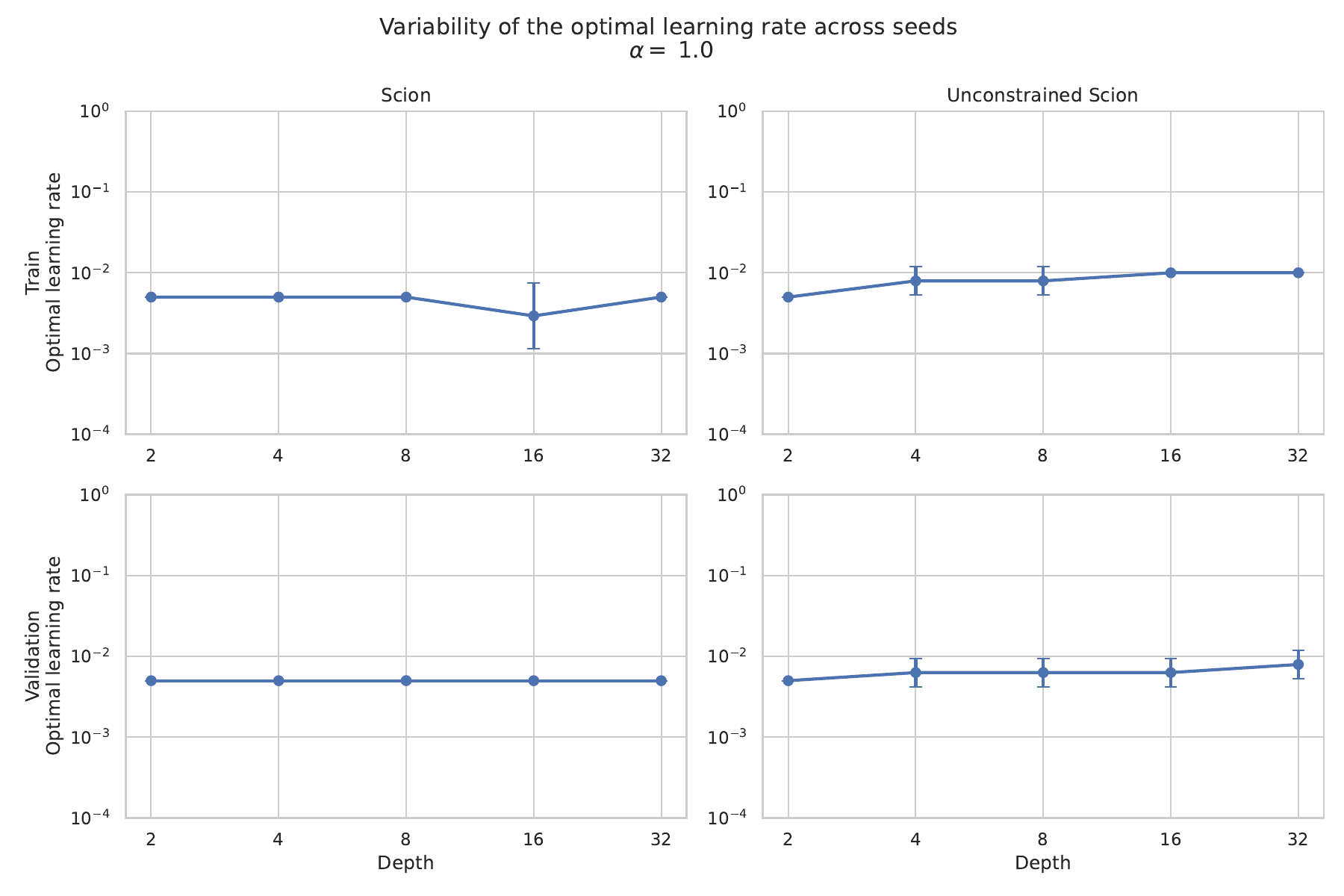}
    \end{subfigure}
    \caption{
    Variability across seeds of the optimal learning rates per depths with \Scion and \uScion for $\alpha=0.5$ (\textit{left plots}) and $\alpha=1$ (\textit{right plots}) on MLP-Mixer trained on CIFAR-10.
    The experimental details are in Table \ref{tab:mlpmixer_hp}.
    }
    \label{fig:error_bars_mlpmixer}
\end{figure}

To further assess the stability across depth observed in the experiments, we plot in Figures \ref{fig:error_bars_nanogpt} and \ref{fig:error_bars_mlpmixer} the optimal learning rate for each depth, together with error bars capturing the variability across three seeds.


\clearpage 
\section{Weight alignment and non-cancellation}
\label{appdx-sec:weight_alignment_and_non_cancellation}
The experiments presented below are performed under a training regime that is closer to standard practice than the one adopted in \cite{Yang2023ASpectralCondtion}. Our implementation builds on the code provided by the authors, where we modify the experimental setup so that all network layers are updated jointly over multiple optimization steps. By contrast, the experiments in \cite{Yang2023ASpectralCondtion} are restricted to a single optimization step, with only the layer of interest being updated and all remaining layers held fixed.

\subsection{Alignment of the weights with the incoming vector}
\begin{figure}[ht!]
    \centering
    \begin{subfigure}{0.45\linewidth}
        \includegraphics[width=\linewidth]{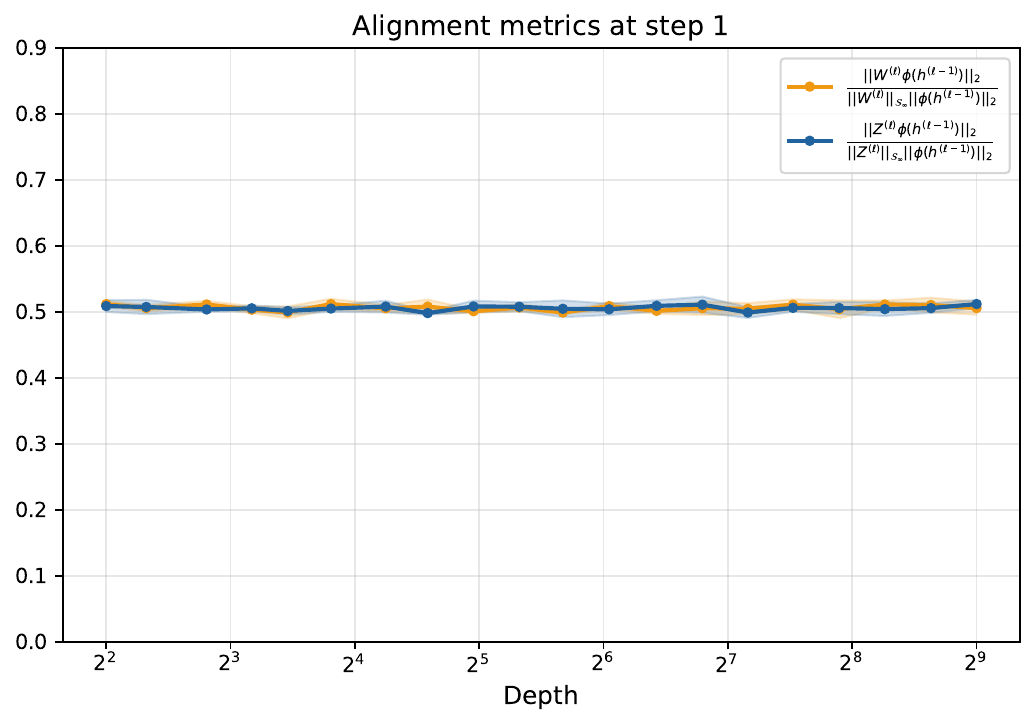}
    \end{subfigure}
    \hfill
    \begin{subfigure}{0.45\linewidth}
    \includegraphics[width=\linewidth]{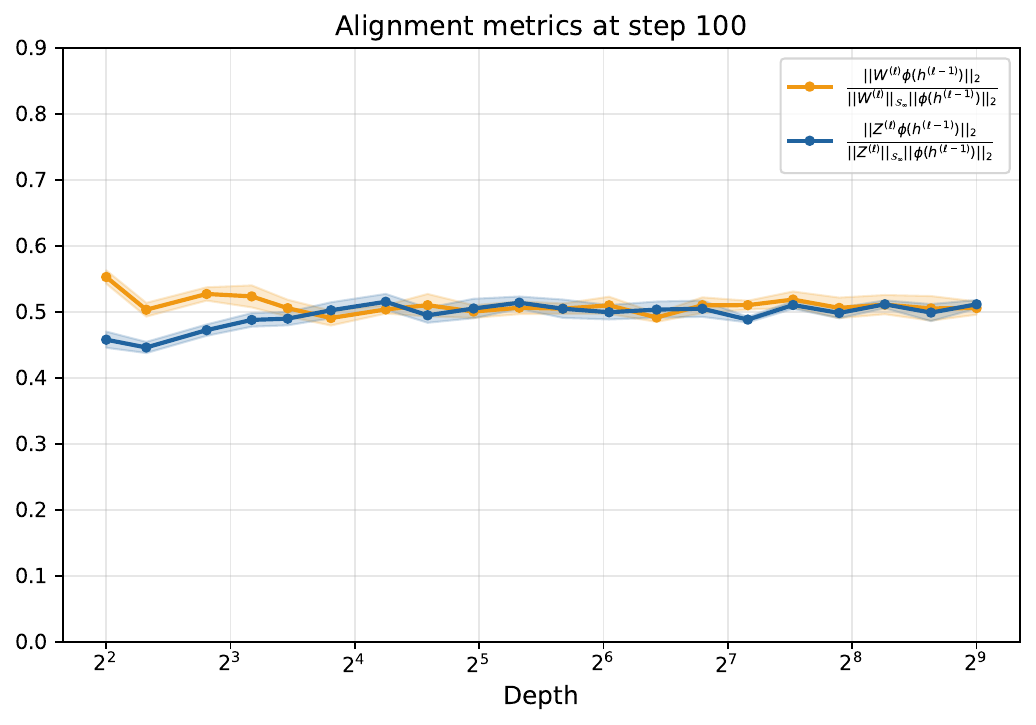}
    \end{subfigure}
    \\
    \begin{subfigure}{0.45\linewidth}
    \includegraphics[width=\linewidth]{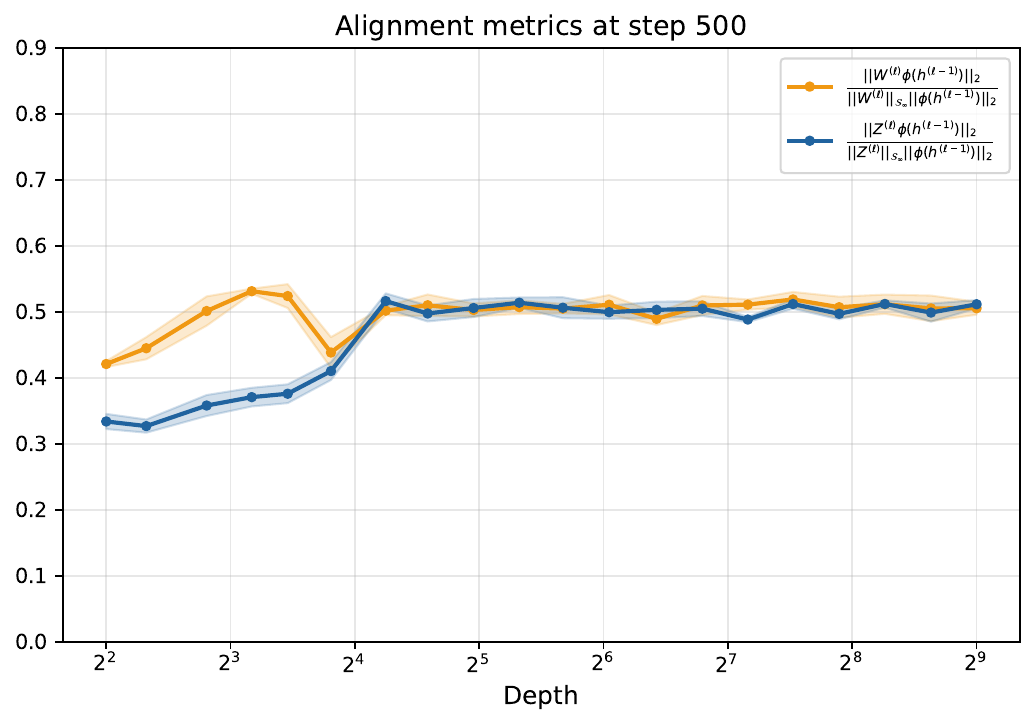}
    \end{subfigure}
    \hfill
    \begin{subfigure}{0.45\linewidth}
    \includegraphics[width=\linewidth]{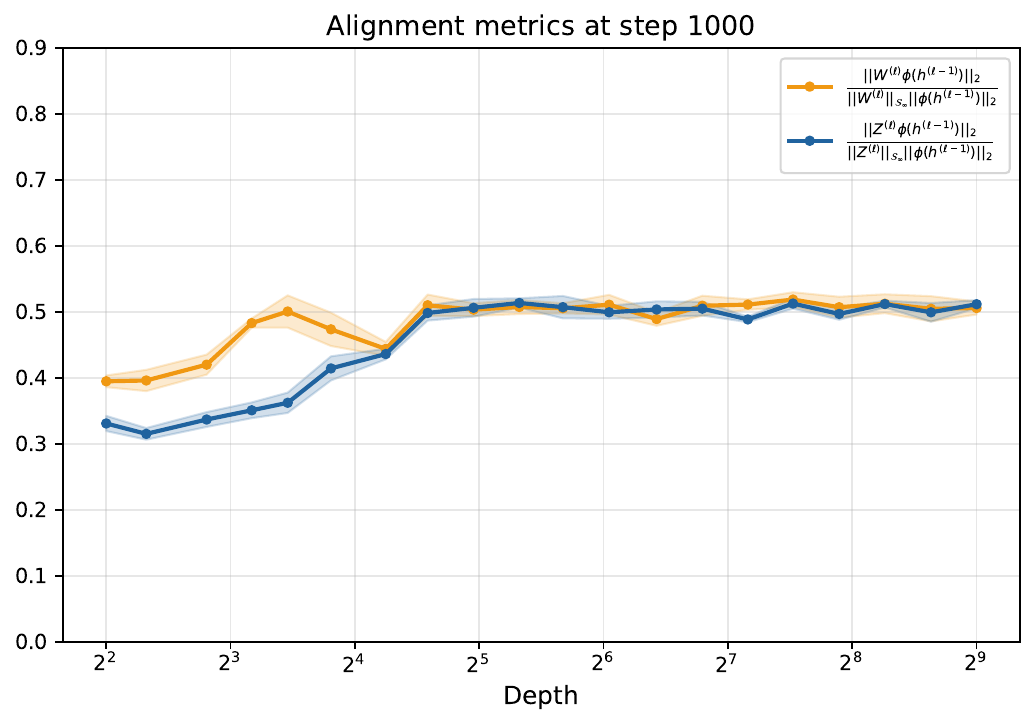}
    \end{subfigure}
    \caption{
    Alignment of weights $W^{(\ell)}$ and $Z^{(\ell)}$ with the vector $\phi(h^{(\ell-1)})$, across a wide range of depths $L$, on CIFAR-10. Both $W^{(\ell)}$ and $Z^{(\ell)}$ exhibit strong alignment regardless of depth. The network is a ResNet where we learn the matrices $Z^{(\ell)}$, trained with \uScion on CIFAR-10, on 5 trials. The layer of interest here is the third hidden layer. Experimental details can be found in Table \ref{tab:resnet_hp_alignment}.}
    \label{fig:weight_alignment_W_and_Z_with_phi_h}
\end{figure}

In Figure \ref{fig:weight_alignment_W_and_Z_with_phi_h}, we report 
\begin{equation}
    \frac{\lVert W^{(\ell)} \phi(h^{(\ell -1)}) \rVert_2}{\lVert W^{(\ell)} \rVert_{\mathcal{S}_\infty} \lVert \phi(h^{(\ell -1)})\rVert_2} 
    \quad \text{and} \quad
    \frac{\lVert Z^{(\ell)} \phi(h^{(\ell -1)}) \rVert_2}{\lVert Z^{(\ell)} \rVert_{\mathcal{S}_\infty} \lVert \phi(h^{(\ell -1)}) \rVert_2}
\end{equation}
for various depths and iteration steps.

These experiments provide an empirical assessment of Assumption \ref{eq:hyp_alignment} beyond initialization and throughout training.

\subsection{Alignment of the sum of the residual block with the previous state}
\begin{figure}[ht!]
    \centering
    \begin{subfigure}{0.48\linewidth}
        \centering
        \includegraphics[width=\linewidth]{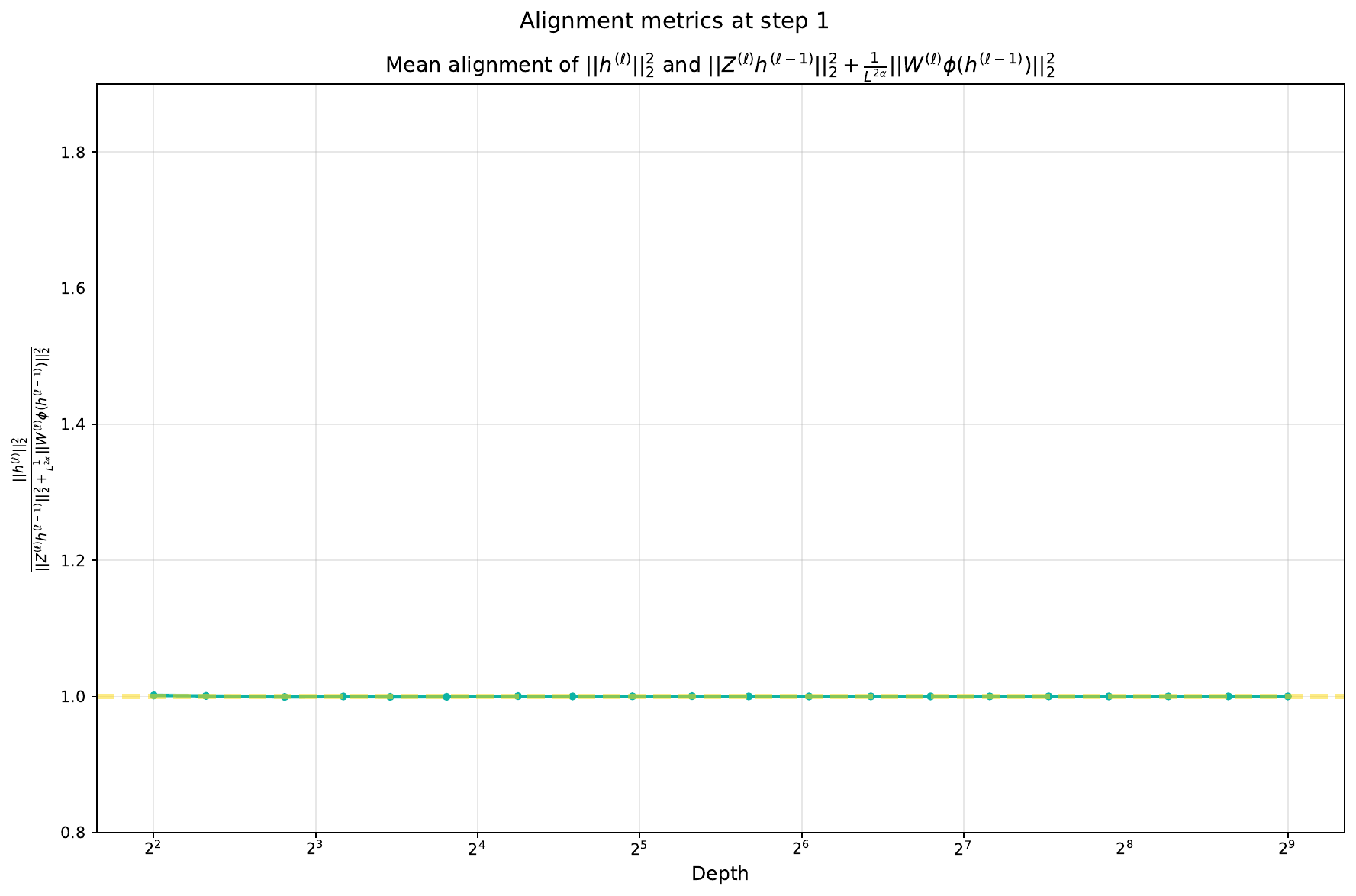}
    \end{subfigure}
    \hfill
    \begin{subfigure}{0.48\linewidth}
        \centering
        \includegraphics[width=\linewidth]{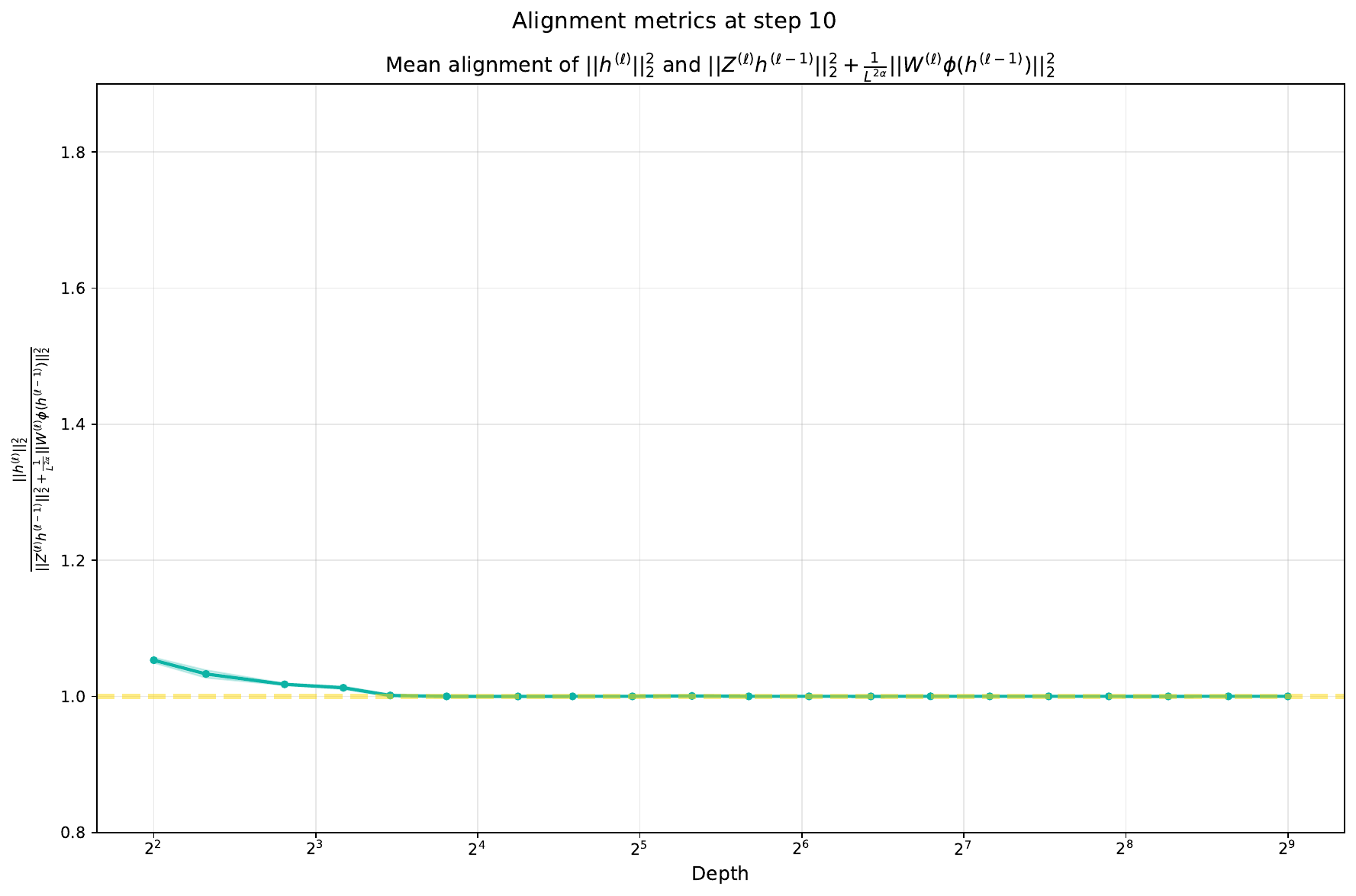}
    \end{subfigure}

    \vspace{0.5em}

    \begin{subfigure}{0.48\linewidth}
        \centering
        \includegraphics[width=\linewidth]{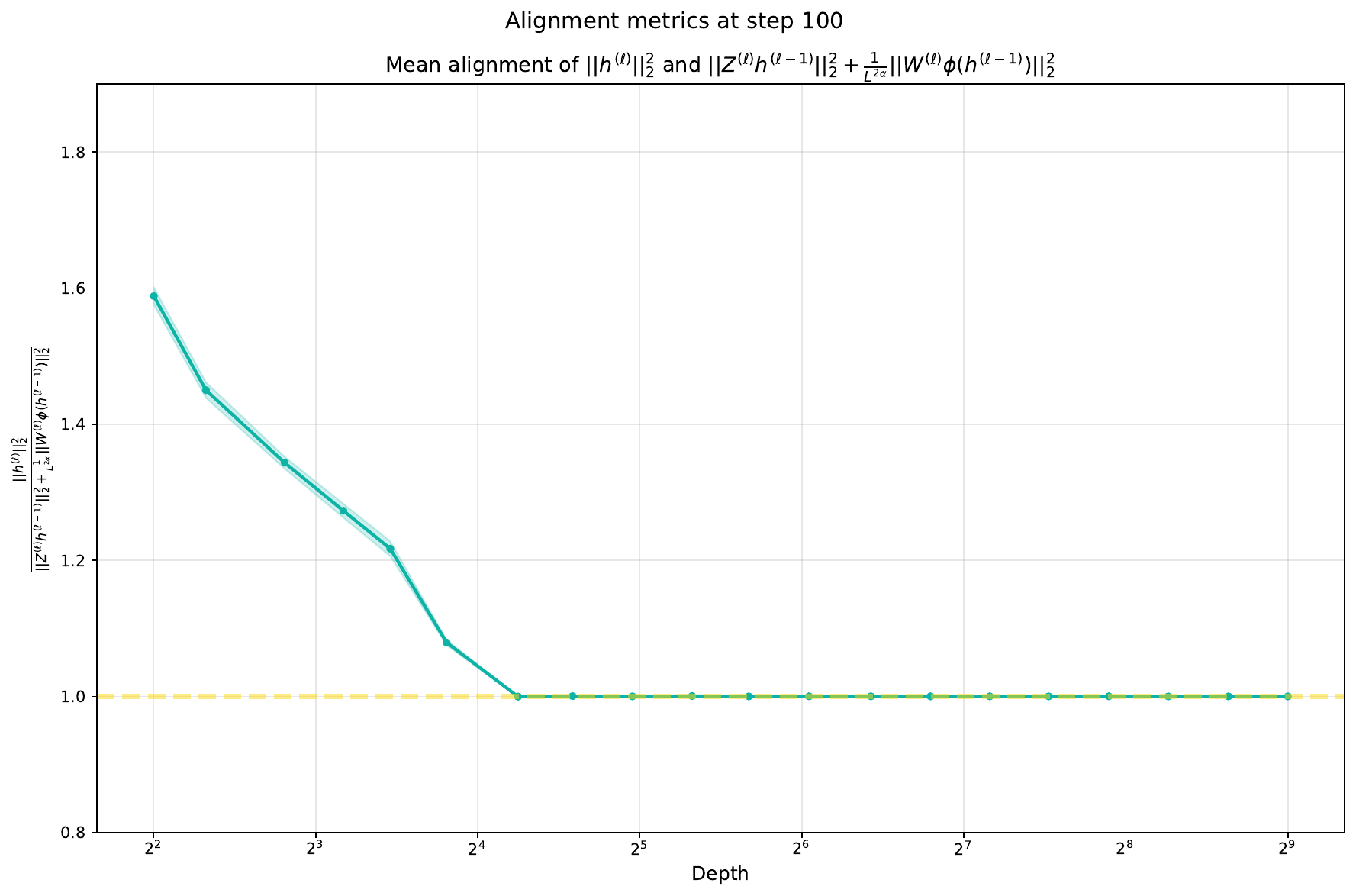}
    \end{subfigure}
    \hfill
    \begin{subfigure}{0.48\linewidth}
        \centering
        \includegraphics[width=\linewidth]{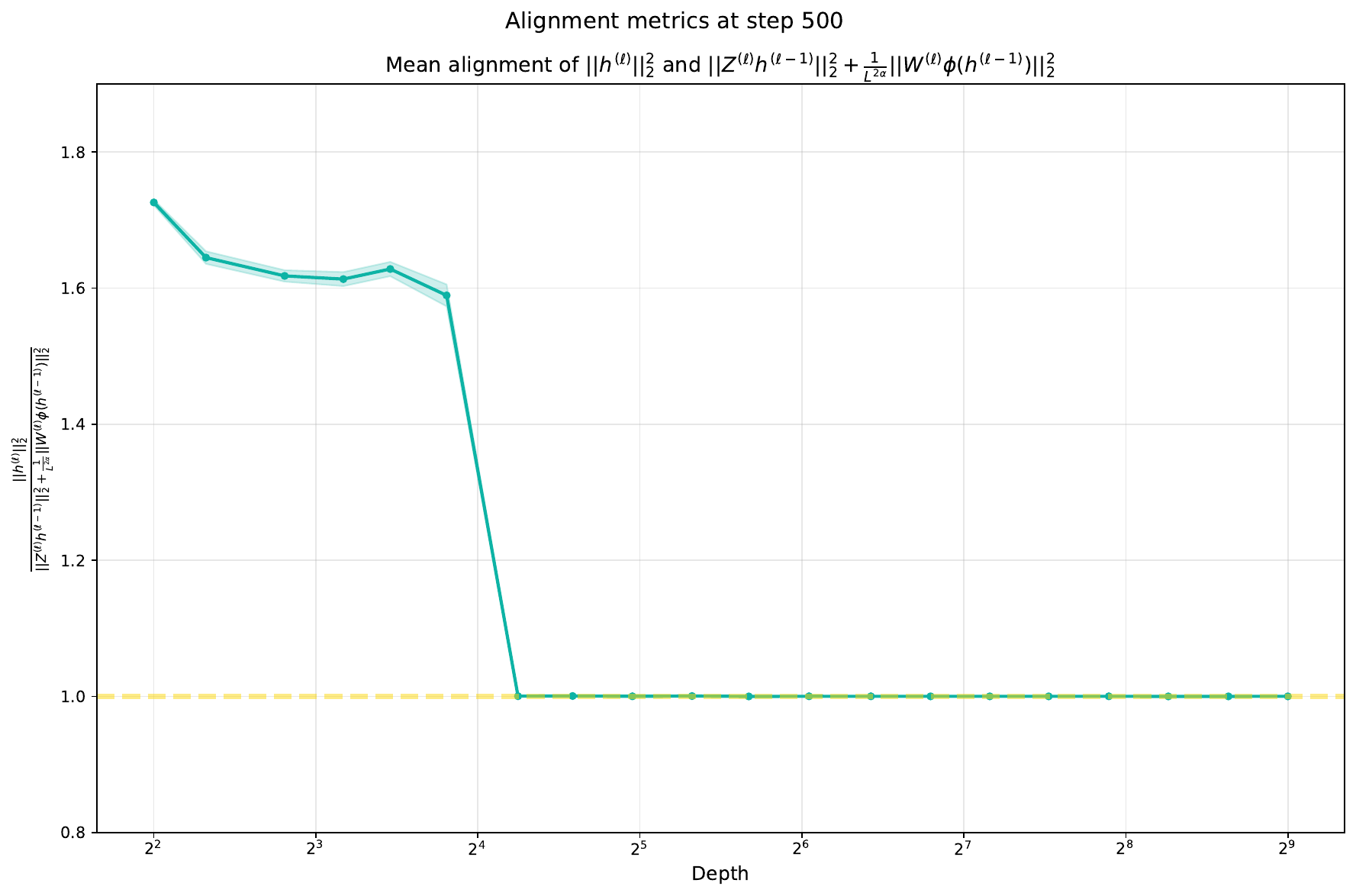}
    \end{subfigure}
    \caption{
    Alignment of the hidden features $h^{(\ell)}$ with the sum of the previous block and the residual block, scaled by the factor $L^{-\alpha}$, across a wide range of depths $L$, on CIFAR-10. 
    Despite the non stability we observe at small depth, we still note that the ratio is always above $1$, which empirically shows that the lower bound holds. The network is a ResNet where we learn the matrices $Z^{(\ell)}$, trained with \uScion on CIFAR-10, on 5 trials. The layer of interest here is the third hidden layer. Experimental details can be found in Table \ref{tab:resnet_hp_alignment}.
    }
    \label{fig:alignment_sum_res_block}
\end{figure}

In Figure \ref{fig:alignment_sum_res_block}, we report 
\begin{equation}
    \frac{ \lVert h^{(\ell)}\rVert_2^2 }{ \lVert Z^{(\ell)} h^{(\ell -1)} \rVert_2^2 + \frac{1}{L^{2\alpha}}  \lVert W^{(\ell)} \phi(h^{(\ell -1)}) \rVert_2^2}. 
\end{equation}
for various depths and iteration steps.

\subsection{Alignment of the updates of the layer with \ubScion}
\begin{figure}[ht!]
    \centering
    \begin{subfigure}{0.48\linewidth}
        \centering
        \includegraphics[width=\linewidth]{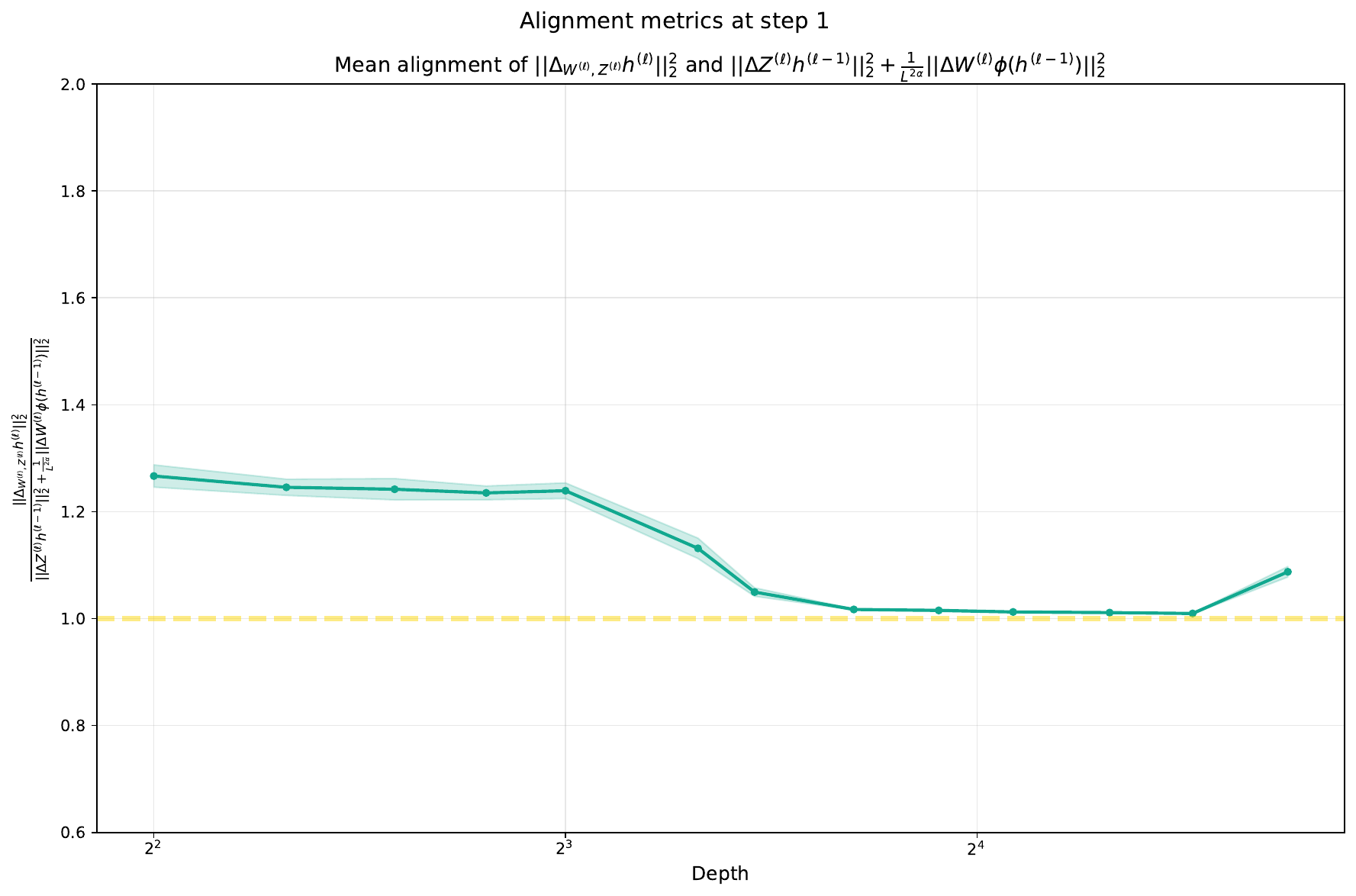}
    \end{subfigure}
    \hfill
    \begin{subfigure}{0.48\linewidth}
        \centering
        \includegraphics[width=\linewidth]{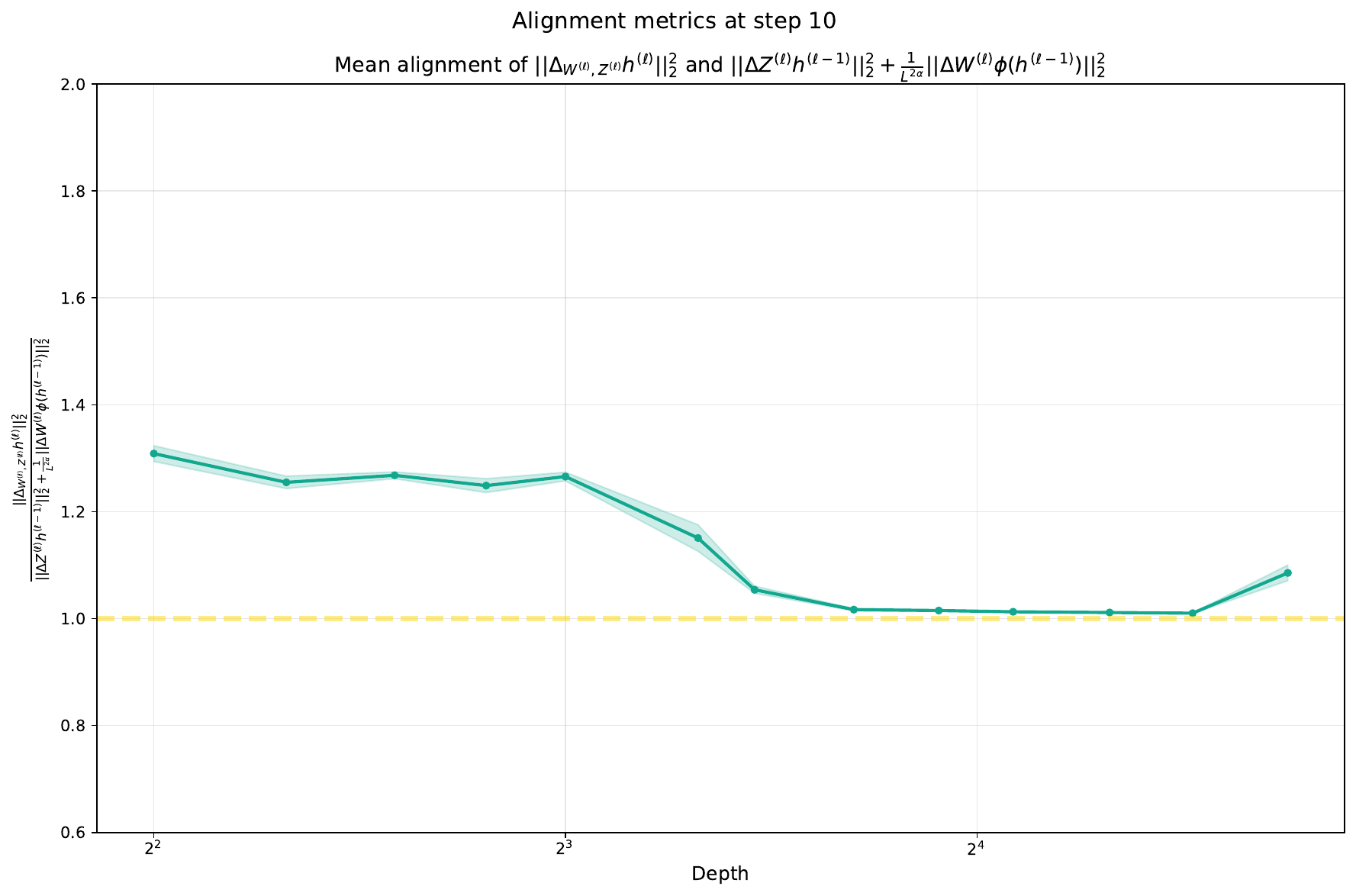}
    \end{subfigure}

    \vspace{0.5em}

    \begin{subfigure}{0.48\linewidth}
        \centering
        \includegraphics[width=\linewidth]{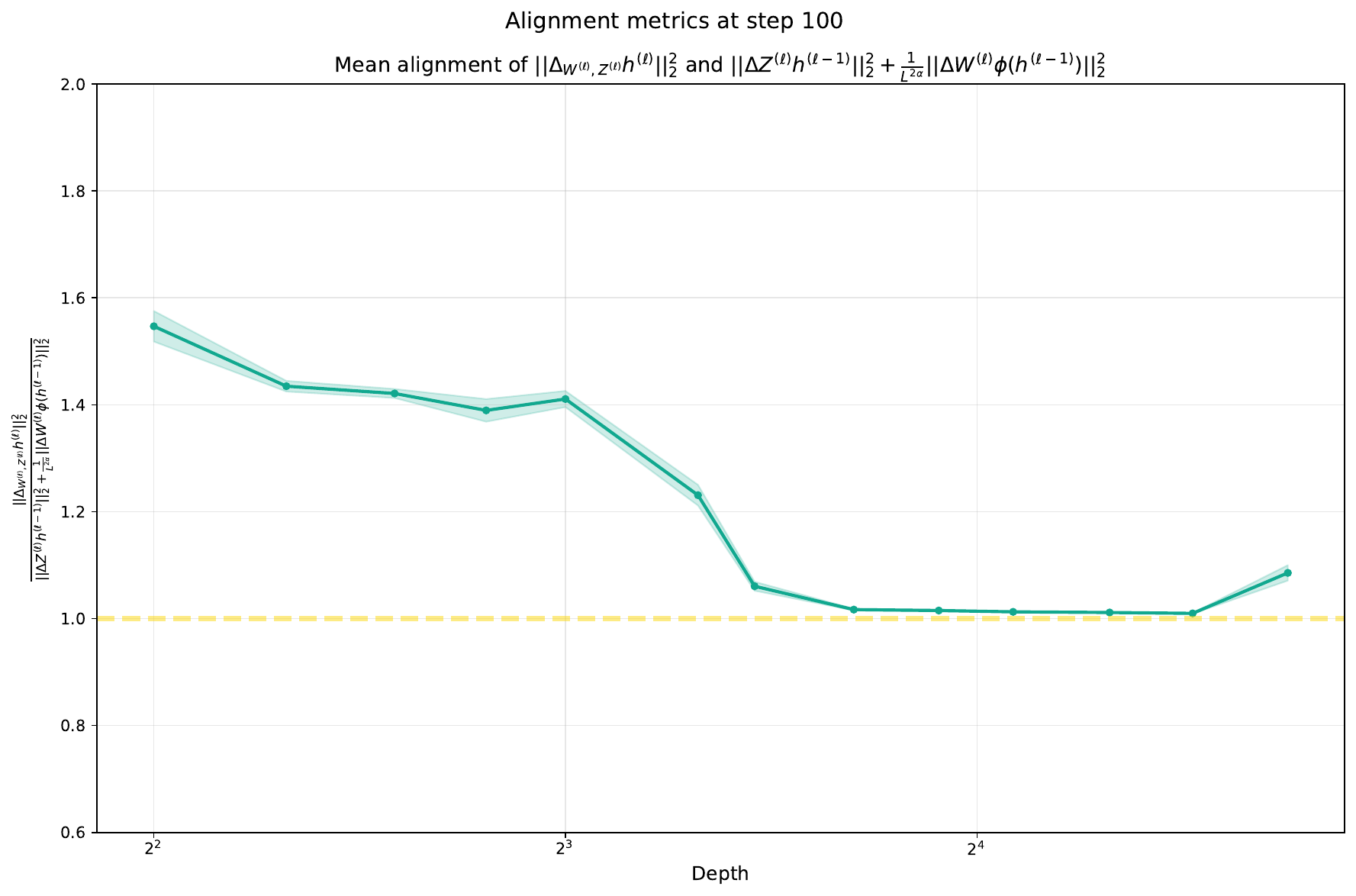}
    \end{subfigure}
    \hfill
    \begin{subfigure}{0.48\linewidth}
        \centering
        \includegraphics[width=\linewidth]{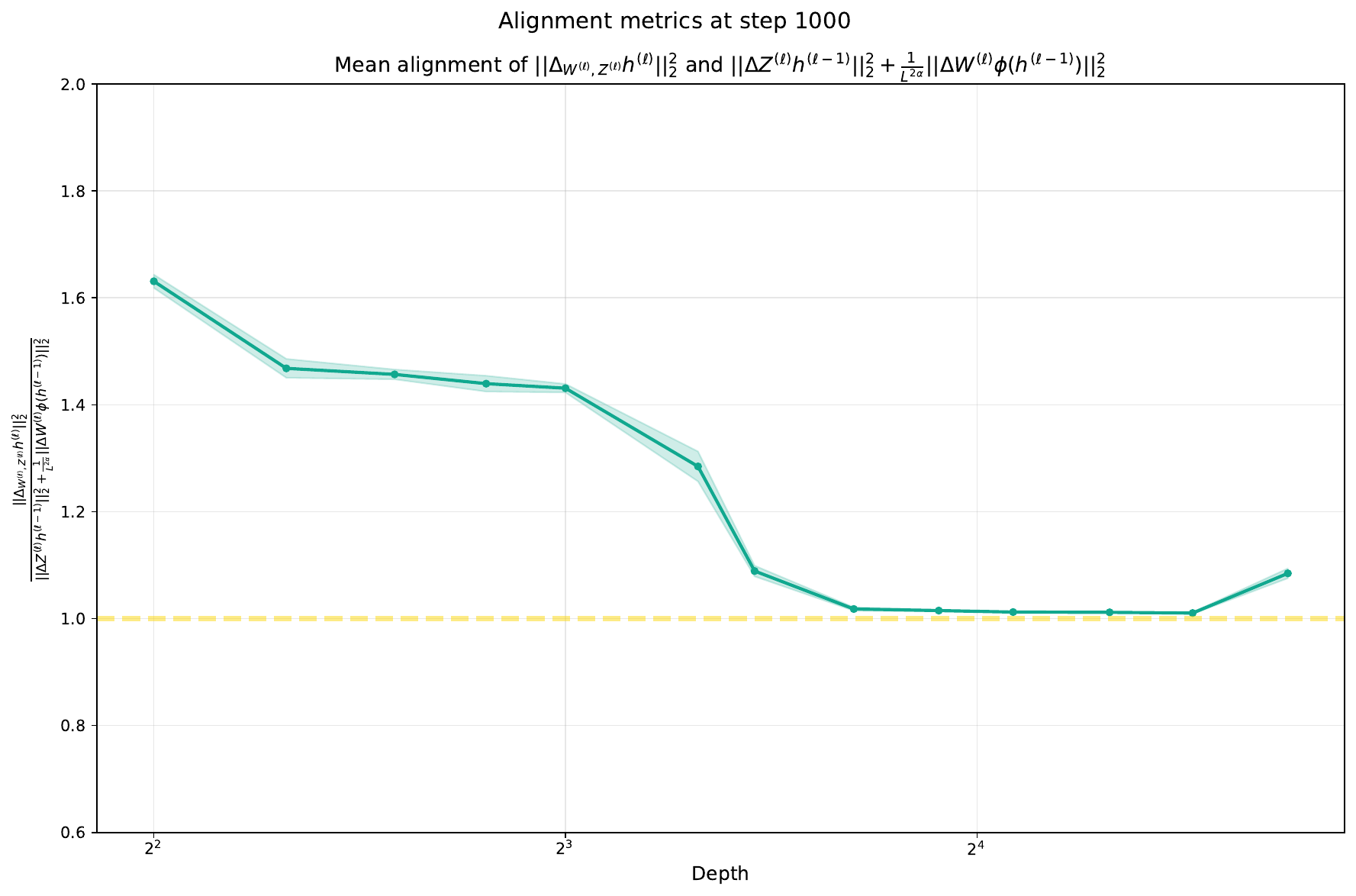}
    \end{subfigure}
    \caption{
    Alignment of the update of the hidden features $\Delta_{W^{(\ell)}, Z^{(\ell)}}h^{(\ell)}$ with the sum of the updates applied to the layer $\ell$, across a wide range of depths $L$, on CIFAR-10. 
    Despite the non stability we observe at small depth, we still note that the ratio is always above $1$, which empirically show that the lower bound holds.
    The network is a ResNet where we learn the matrices $Z^{(\ell)}$, trained with \uScion, on CIFAR-10, on 5 trials. The layer of interest here is the third hidden layer. Experimental details can be found in Table \ref{tab:resnet_hp_alignment}.}
    \label{fig:alignment_sum_update_res_block}
\end{figure}

In Figure \ref{fig:alignment_sum_update_res_block}, we report 
\begin{equation}
    \frac{\lVert \Delta_{W^{(\ell)}, Z^{(\ell)}}h^{(\ell)} \rVert_2^2}{\lVert \Delta Z^{(\ell)} h^{(\ell-1)} \lVert_2^2+ \frac{1}{L^{2\alpha}} \lVert \Delta W^{(\ell)} \phi(h^{(\ell -1)}) \rVert_2^2}
\end{equation}
for various depths and iteration steps, where the weights are updated with \uScion.

\subsection{Alignment of the updates of the layer with \bScion}
\begin{figure}[ht!]
    \centering
    \begin{subfigure}{0.32\linewidth}
        \centering
        \includegraphics[width=\linewidth]{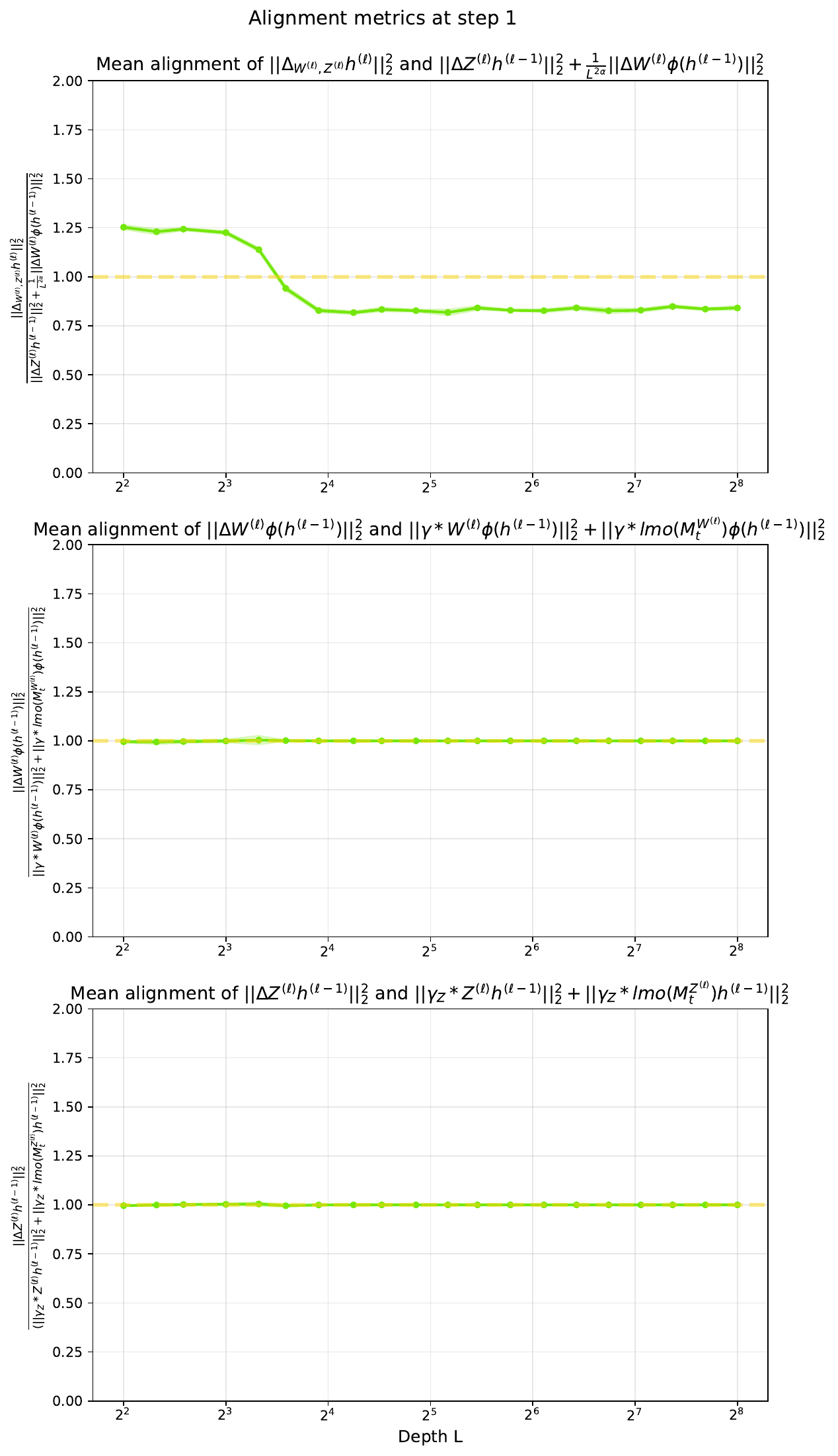}
    \end{subfigure}
    \hfill
    \begin{subfigure}{0.32\linewidth}
        \centering
        \includegraphics[width=\linewidth]{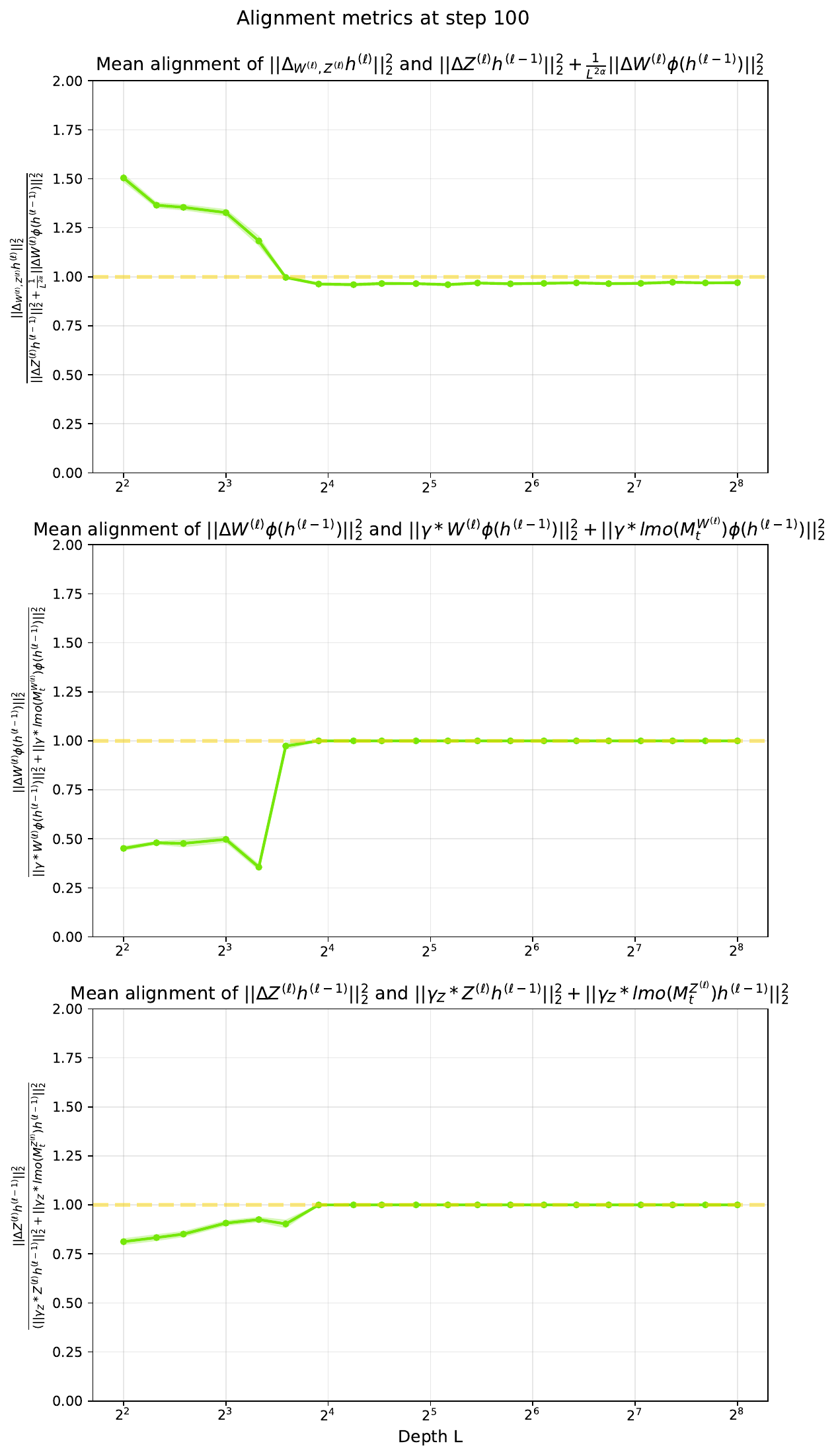}
    \end{subfigure}
    \hfill
    \begin{subfigure}{0.32\linewidth}
        \centering
        \includegraphics[width=\linewidth]{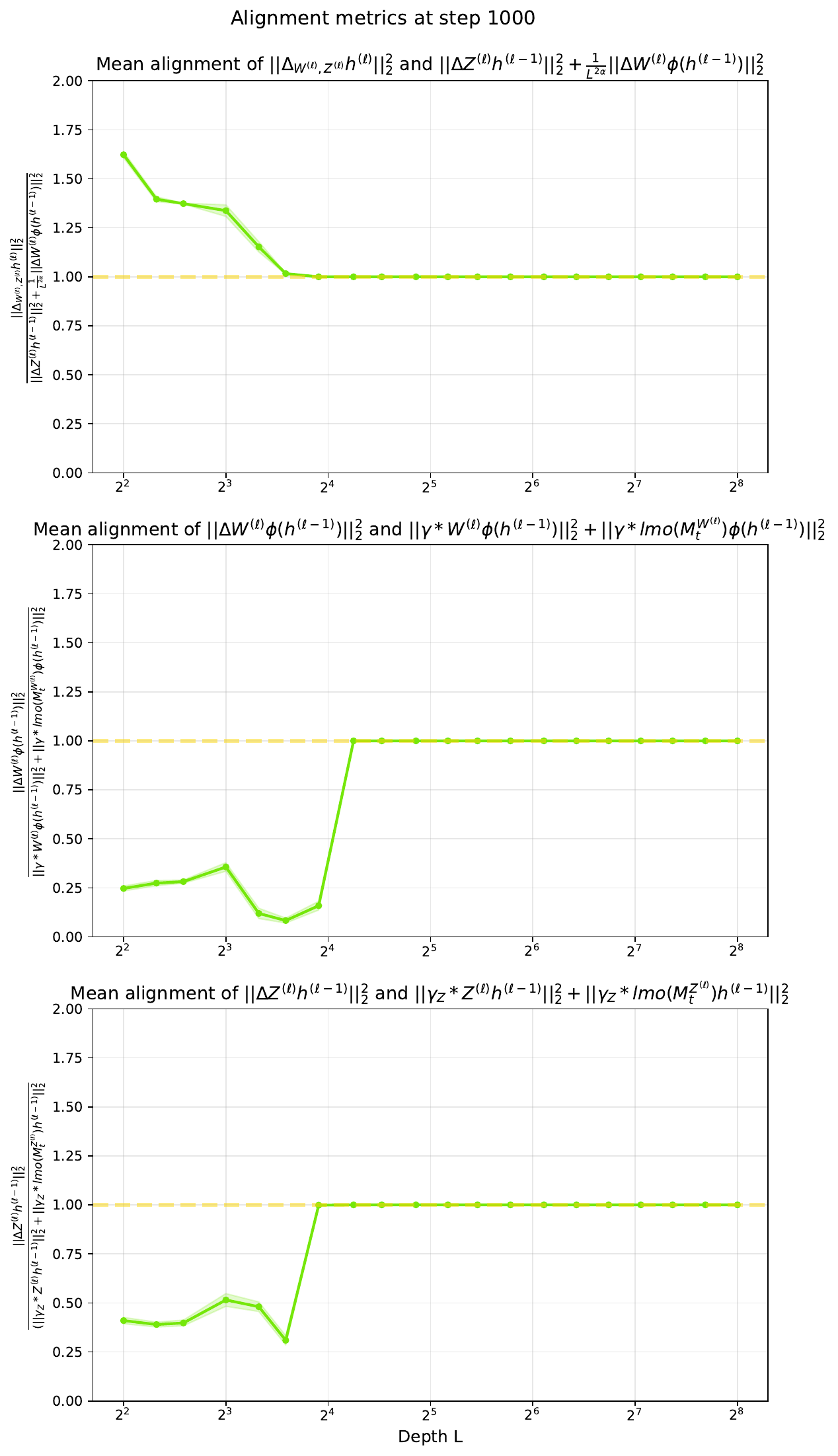}
    \end{subfigure}
    \caption{
    \textit{First row: } Alignment of the update of the hidden features $\Delta_{W^{(\ell)}, Z^{(\ell)}}h^{(\ell)}$ with the sum of the updates applied to the layer $\ell$, across a wide range of depths $L$, on CIFAR-10. 
    \textit{Second row: } Alignment of the update of $W^{(\ell)}$ with the sum of the current $W^{(\ell)}$ and $\lmo({M_t^{W^{(\ell)}}})$.
    \textit{Third row: } Alignment of the update of $Z^{(\ell)}$ with the sum of the current $Z^{(\ell)}$ and $\lmo({M_t^{Z^{(\ell)}}})$.  
    Despite the non stability we observe at small depth, we still note that the ratio stabilizes at larger depth to a constant close to $1$, which empirically show that the alignment is a reasonable assumption.
    The network is a ResNet where we learn the matrices $Z^{(\ell)}$, trained with \Scion, on CIFAR-10, on 5 trials. The layer of interest here is the third hidden layer. Experimental details can be found in Table \ref{tab:resnet_hp_alignment}.}
    \label{fig:alignment_sum_update_res_block_with_scion}
\end{figure}

In Figure \ref{fig:alignment_sum_update_res_block_with_scion}, we report 
\begin{equation}
\begin{aligned}
    & \frac{\lVert \Delta_{W^{(\ell)}, Z^{(\ell)}}h^{(\ell)} \rVert_2^2}{\lVert \Delta Z^{(\ell)} h^{(\ell-1)} \lVert_2^2+ \frac{1}{L^{2\alpha}} \lVert \Delta W^{(\ell)} \phi(h^{(\ell -1)}) \rVert_2^2} \\
\end{aligned}
\end{equation}
for various depths and iteration steps, where the weights are updated with \Scion.\\

We also report

\begin{equation}
\begin{aligned}
     &\frac{\lVert\Delta W^{(\ell)} \phi(h^{(\ell-1)})\rVert_2^2}{\lVert \gamma * W^{(\ell)} \phi(h^{(\ell-1)})\rVert_2^2 + \lVert \gamma  \lmo({M_t^{W^{(\ell)}}}) \phi(h^{(\ell-1)})\rVert_2^2} \\
    \text{ and } \quad
    &\frac{\lVert\Delta Z^{(\ell)} h^{(\ell-1)}\rVert_2^2}{(\lVert \gamma_Z * Z^{(\ell)} h^{(\ell-1)}\rVert_2^2 + \lVert \gamma_Z  \lmo({M_t^{Z^{(\ell)}}})  h^{(\ell-1)}\rVert_2^2}.
\end{aligned}
\end{equation}

\subsection{Alignment of the updates of the layer with \bScion for the width transfer}
\begin{figure}[ht!]
    \centering
    \begin{subfigure}{0.32\linewidth}
        \centering
        \includegraphics[width=\linewidth]{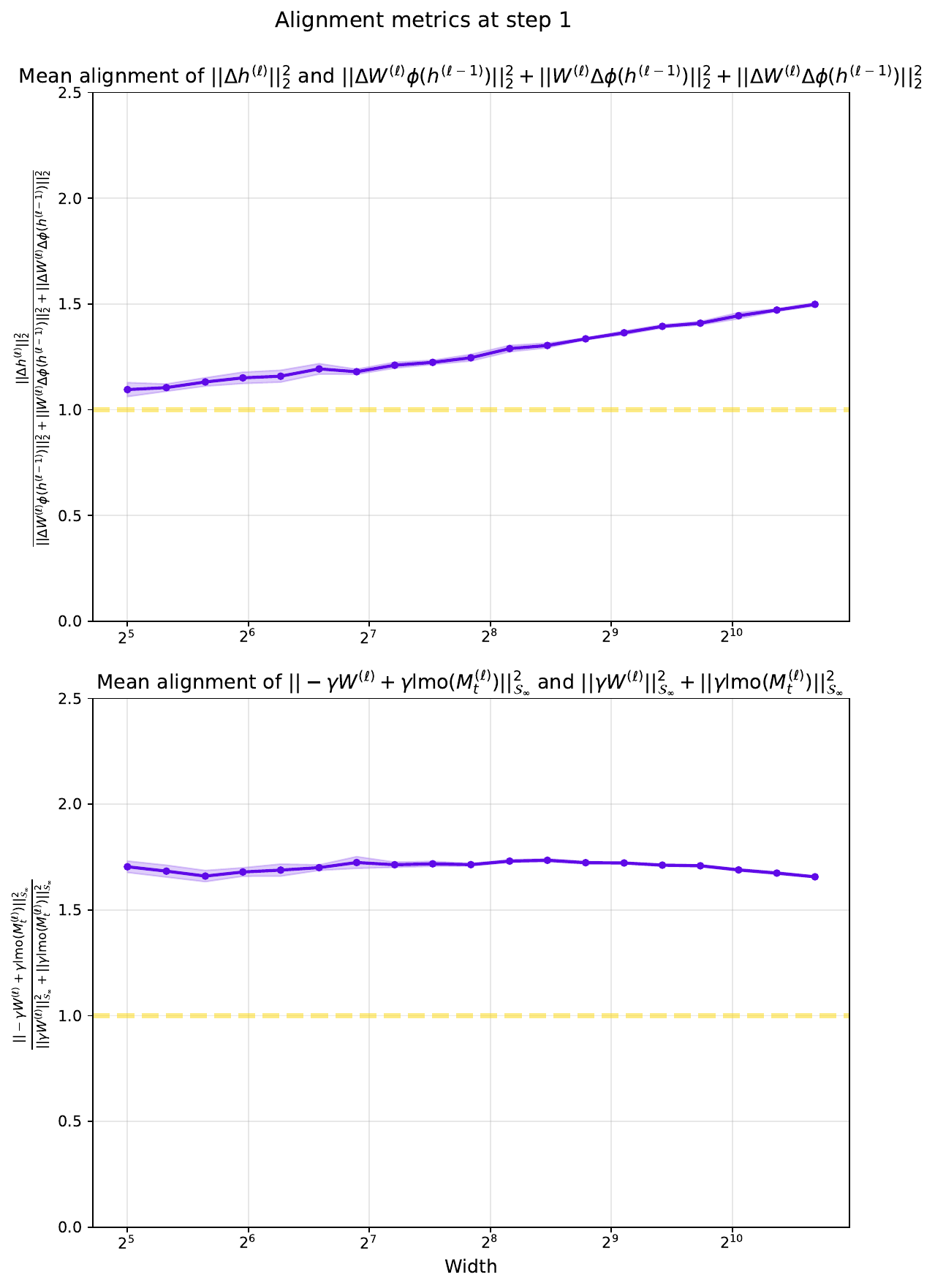}
    \end{subfigure}
    \hfill
    \begin{subfigure}{0.32\linewidth}
        \centering
        \includegraphics[width=\linewidth]{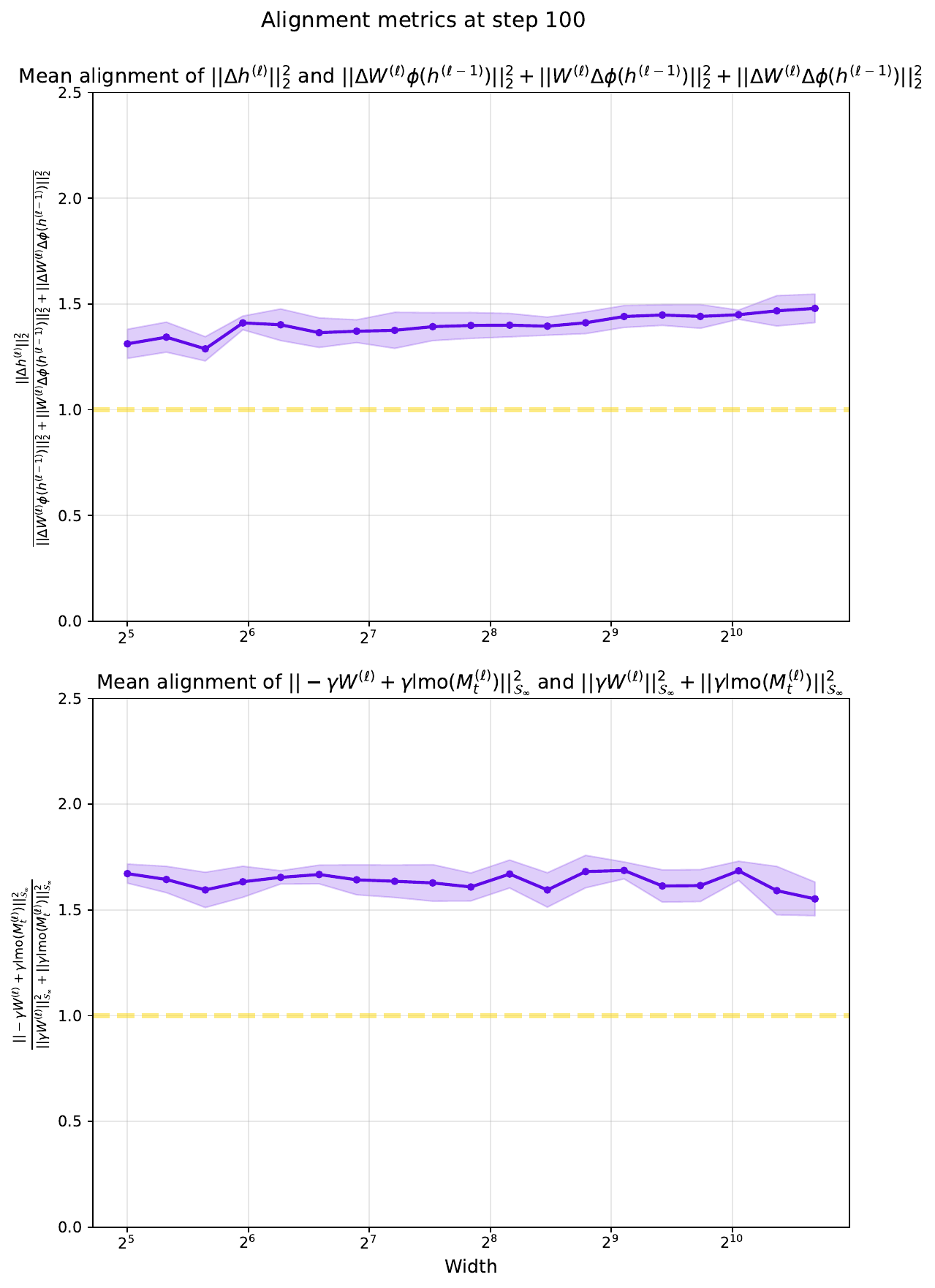}
    \end{subfigure}
    \hfill
    \begin{subfigure}{0.32\linewidth}
        \centering
        \includegraphics[width=\linewidth]{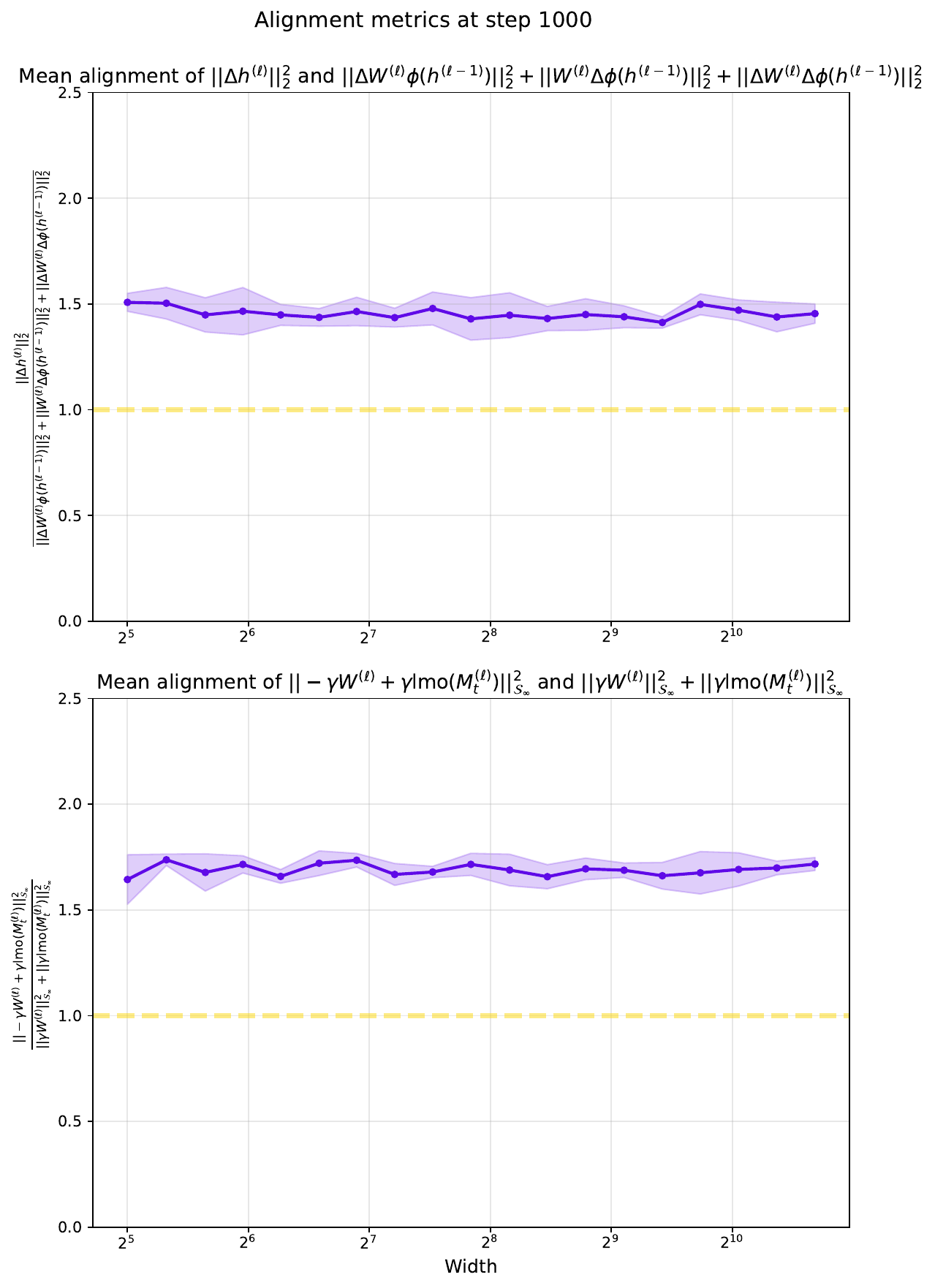}
    \end{subfigure}
    \caption{
    \textit{First row: } Alignment between the update of the hidden features $\Delta h^{(\ell)}$ and the three terms appearing from the update of $W^{(\ell)}\phi(h^{(\ell-1)})$, across a range of network widths, on CIFAR-10.
    \textit{Second row: } Alignment between the update of $W^{(\ell)}$ and the sum of the squared spectral norms of $W^{(\ell)}$ and the corresponding $\lmo$.
    The network is an MLP trained with \Scion on CIFAR-10, over 5 trials. The layer of interest is the second hidden layer. The plots are shown at different training steps and across a wide range of network widths.
    Experimental details can be found in Table \ref{tab:mlp_hp_alignment}.}
    \label{fig:alignment_sum_update_with_scion_width}
\end{figure}

In Figure \ref{fig:alignment_sum_update_with_scion_width}, we study the non-cancellation properties of the updates of \Scion, as we scale the width of the MLP. 
We report:
\begin{equation}
\begin{aligned}
\frac{\lVert \Delta h^{(\ell)} \rVert_2^2}{
    \lVert \Delta W^{(\ell)} \phi(h^{(\ell-1)}) \rVert_2^2
    + \lVert W^{(\ell)} \Delta \phi(h^{(\ell-1)}) \rVert_2^2
    + \lVert \Delta W^{(\ell)} \Delta \phi(h^{(\ell-1)}) \rVert_2^2
},
\end{aligned}
\end{equation}
for various network widths and iteration steps $t$, where the weights are updated with \Scion.

We also report the alignment of the current quantities with the update,
\begin{equation}
\frac{
    \lVert
    -\gamma W^{(\ell)} + \gamma \lmo(M_t^{(\ell)})
    \rVert_{\mathcal{S}_\infty}^2
}{
    \lVert
    \gamma W^{(\ell)}
    \rVert_{\mathcal{S}_\infty}^2
    +
    \lVert
    \gamma  \lmo(M_t^{(\ell)})
    \rVert_{\mathcal{S}_\infty}^2
}.
\end{equation}

%% file: bibliography.bib
@misc{dey2026dontlazycompletepenables,
      title={Don't be lazy: {CompleteP} enables compute-efficient deep transformers}, 
      author={Nolan Dey and Bin Claire Zhang and Lorenzo Noci and Mufan Li and Blake Bordelon and Shane Bergsma and Cengiz Pehlevan and Boris Hanin and Joel Hestness},
      year={2026},
      eprint={2505.01618},
      archivePrefix={arXiv},
      primaryClass={cs.LG},
      url={https://arxiv.org/abs/2505.01618}, 
}

@misc{jordan2024muon,
  author       = {Keller Jordan and Yuchen Jin and Vlado Boza and Jiacheng You and
                  Franz Cesista and Laker Newhouse and Jeremy Bernstein},
  title        = {Muon: An optimizer for hidden layers in neural networks},
  year         = {2024},
  url          = {https://kellerjordan.github.io/posts/muon/}
}

@misc{modded_nanogpt_2024,
  author = {Keller Jordan and Jeremy Bernstein and Brendan Rappazzo and
            {\tt @fernbear.bsky.social} and Boza Vlado and You Jiacheng and
            Franz Cesista and Braden Koszarsky and {\tt @Grad62304977}},
  title  = {Modded-{NanoGPT}: Speedrunning the {NanoGPT} Baseline},
  year   = {2024},
  url    = {https://github.com/KellerJordan/modded-nanogpt},
}

@inproceedings{yang2021TPIVFeatureLeanringWidth,
  author    = {Greg Yang and Edward J. Hu},
  title     = {Tensor Programs {IV}: Feature Learning in Infinite-Width Neural Networks},
  booktitle = {International Conference on Machine Learning},
  pages     = {11727--11737},
  year      = {2021},
  publisher = {PMLR},
}

@inproceedings{yang2021TPVZeroShot,
  author    = {Greg Yang and Edward J. Hu and Igor Babuschkin and Szymon Sidor and Xiaodong Liu and David Farhi and Nick Ryder and Jakub Pachocki and Weizhu Chen and Jianfeng Gao},
  title     = {Tensor Programs {V}: Tuning Large Neural Networks via Zero-Shot Hyperparameter Transfer},
  booktitle = {Advances in Neural Information Processing Systems (NeurIPS)},
  year      = {2021},
}

@inproceedings{yang2023TPIVb,
    title={Tensor Programs {IVb}: Adaptive Optimization in the Infinite-Width Limit},
    author={Yang, Greg and Littwin, Etai},
    booktitle={International Conference on Learning Representations (ICLR)},
    year={2023}
}

@inproceedings{yang2023TPIVFeatureLearningDepth,
  title={Tensor Programs {VI}: Feature Learning in Infinite-Depth Neural Networks},
  author={Yang, Greg and Yu, Dingli and Zhu, Chen and Hayou, Soufiane},
  booktitle={NeurIPS 2023 Workshop on Mathematics of Modern Machine Learning},
  year={2023}
}

@article{Yang2023ASpectralCondtion,
  author  = {Greg Yang and James B. Simon and Jeremy Bernstein},
  title   = {A Spectral Condition for Feature Learning},
  journal = {arXiv preprint arXiv:2310.17813},
  year    = {2023},
}

@article{bordelon2022self,
  author  = {Blake Bordelon and Cengiz Pehlevan},
  title   = {Self-Consistent Dynamical Field Theory of Kernel Evolution in Wide Neural Networks},
  journal = {Advances in Neural Information Processing Systems},
  volume  = {35},
  pages   = {32240--32256},
  year    = {2022},
}

@inproceedings{bordelon2024depthwise,
  author    = {Blake Bordelon and Lorenzo Noci and Mufan Bill Li and Boris Hanin and Cengiz Pehlevan},
  title     = {Depthwise Hyperparameter Transfer in Residual Networks: Dynamics and Scaling Limit},
  booktitle = {The Twelfth International Conference on Learning Representations},
  year      = {2024},
}

@article{bordelon2024infinite,
  title={Infinite limits of multi-head transformer dynamics},
  author={Bordelon, Blake and Chaudhry, Hamza and Pehlevan, Cengiz},
  journal={Advances in Neural Information Processing Systems},
  volume={37},
  pages={35824--35878},
  year={2024}
}

@article{zheng2026spectral,
  author  = {Chenyu Zheng and Rongzhen Wang and Xinyu Zhang and Chongxuan Li},
  title   = {Spectral Condition for {$\mu$P} under Width-Depth Scaling},
  journal = {arXiv preprint arXiv:2603.00541},
  year    = {2026},
}

@article{qiu2025hyperparameter,
  author  = {Shikai Qiu and Zixi Chen and Hoang Phan and Qi Lei and Andrew Gordon Wilson},
  title   = {Hyperparameter Transfer Enables Consistent Gains of Matrix-Preconditioned  Optimizers Across Scales},
  journal = {arXiv preprint arXiv:2512.05620},
  year    = {2025},
}

@article{kosson2025weight,
  author  = {Atli Kosson and Jeremy Welborn and Yang Liu and Martin Jaggi and Xi Chen},
  title   = {Weight Decay May Matter More Than {muP} for Learning Rate Transfer in Practice},
  journal = {arXiv preprint arXiv:2510.19093},
  year    = {2025},
}

@article{chizat2019lazy,
  author  = {L{\'e}na{\"i}c Chizat and Edouard Oyallon and Francis Bach},
  title   = {On Lazy Training in Differentiable Programming},
  journal = {Advances in Neural Information Processing Systems},
  volume  = {32},
  year    = {2019},
}

@article{chennuru2024feature,
  author  = {Leena Chennuru Vankadara and Jin Xu and Moritz Haas and Volkan Cevher},
  title   = {On Feature Learning in Structured State Space Models},
  journal = {Advances in Neural Information Processing Systems},
  volume  = {37},
  pages   = {86145--86179},
  year    = {2024},
}

@misc{xu2026widthscalingneuraloptimizers,
      title={On the Width Scaling of Neural Optimizers Under Matrix Operator Norms {I}: Row/Column Normalization and Hyperparameter Transfer}, 
      author={Ruihan Xu and Jiajin Li and Yiping Lu},
      year={2026},
      eprint={2603.09952},
      archivePrefix={arXiv},
      primaryClass={cs.LG},
      url={https://arxiv.org/abs/2603.09952}, 
}

@misc{cesista2025spectralclipping,
  author = {Franz Louis Cesista},
  title  = {Fast, Numerically Stable, and Auto-Differentiable Spectral Clipping via
            {Newton-Schulz} Iteration},
  year   = {2025},
  url    = {http://leloykun.github.io/ponder/spectral-clipping/},
}

@inproceedings{miyato2018spectralnormalizationgenerativeadversarial,
  author    = {Takeru Miyato and Toshiki Kataoka and Masanori Koyama and Yuichi Yoshida},
  title     = {Spectral Normalization for Generative Adversarial Networks},
  booktitle = {6th International Conference on Learning Representations ({ICLR} 2018),
               Vancouver, BC, Canada, April 30 -- May 3, 2018, Conference Track Proceedings},
  publisher = {OpenReview.net},
  year      = {2018},
}

@article{large2024scalable,
  author  = {Tim Large and Yang Liu and Minyoung Huh and Hyojin Bahng and
             Phillip Isola and Jeremy Bernstein},
  title   = {Scalable Optimization in the Modular Norm},
  journal = {Advances in Neural Information Processing Systems},
  volume  = {37},
  pages   = {73501--73548},
  year    = {2024},
}

@misc{filatov2026optimalscalingneedsoptimal,
      title={Optimal Scaling Needs Optimal Norm}, 
      author={Oleg Filatov and Jiangtao Wang and Jan Ebert and Stefan Kesselheim},
      year={2026},
      eprint={2510.03871},
      archivePrefix={arXiv},
      primaryClass={cs.LG},
      url={https://arxiv.org/abs/2510.03871}, 
}

@article{frank_wolfe,
  title={An algorithm for quadratic programming},
  author={Marguerite Frank and Philip Wolfe},
  journal={Naval Research Logistics Quarterly},
  year={1956},
  volume={3},
  pages={95-110},
  url={https://api.semanticscholar.org/CorpusID:122654717}
}

@article{hazan2015beyond,
  author  = {Elad Hazan and Kfir Levy and Shai Shalev-Shwartz},
  title   = {Beyond Convexity: Stochastic Quasi-Convex Optimization},
  journal = {Advances in Neural Information Processing Systems},
  volume  = {28},
  year    = {2015},
}

@inproceedings{bernstein2018signsgd,
  author    = {Jeremy Bernstein and Yu-Xiang Wang and Kamyar Azizzadenesheli and
               Animashree Anandkumar},
  title     = {{signSGD}: Compressed Optimisation for Non-Convex Problems},
  booktitle = {International Conference on Machine Learning},
  pages     = {560--569},
  year      = {2018},
  publisher = {PMLR},
}

@inproceedings{pethick2025trainingdeeplearningmodels,
  author    = {Thomas Pethick and Wanyun Xie and Kimon Antonakopoulos and Zhenyu Zhu and
               Antonio Silveti-Falls and Volkan Cevher},
  title     = {Training Deep Learning Models with Norm-Constrained {LMOs}},
  booktitle = {Forty-Second International Conference on Machine Learning ({ICML} 2025)},
  year      = {2025},
}

@article{pethick2025training,
  title={Training Neural Networks at Any Scale},
  author={Pethick, Thomas and Antonakopoulos, Kimon and Silveti-Falls, Antonio and Vankadara, Leena Chennuru and Cevher, Volkan},
  journal={arXiv preprint arXiv:2511.11163},
  year={2025}
}

@article{kingma2014adam,
  title={Adam: A Method for Stochastic Optimization},
  author={Diederik P. Kingma and Jimmy Ba},
  journal={CoRR},
  year={2014},
  volume={abs/1412.6980},
}

@inproceedings{vyas2025soap,
  author    = {Nikhil Vyas and Depen Morwani and Rosie Zhao and Itai Shapira and
               David Brandfonbrener and Lucas Janson and Sham M. Kakade},
  title     = {{SOAP}: Improving and Stabilizing {Shampoo} Using {Adam} for Language
               Modeling},
  booktitle = {International Conference on Learning Representations (ICLR)},
  year      = {2025},
}

@inproceedings{gupta2018shampoo,
title = "Shampoo: Preconditioned stochastic tensor optimization",
author = "Vineet Gupta and Tomer Koren and Yoram Singer",
note = "Publisher Copyright: {\textcopyright} 2018 35th International Conference on Machine Learning, ICML 2018. All rights reserved.; 35th International Conference on Machine Learning, ICML 2018 ; Conference date: 10-07-2018 Through 15-07-2018",
year = "2018",
language = "English (US)",
series = "35th International Conference on Machine Learning, ICML 2018",
publisher = "International Machine Learning Society (IMLS)",
pages = "2956--2964",
editor = "Jennifer Dy and Andreas Krause",
booktitle = "35th International Conference on Machine Learning, ICML 2018",
}

@article{Riedmiller1993ADA,
  title={A direct adaptive method for faster backpropagation learning: the {RPROP} algorithm},
  author={Martin A. Riedmiller and Heinrich Braun},
  journal={IEEE International Conference on Neural Networks},
  year={1993},
  pages={586-591 vol.1}
}

@inproceedings{he2016deep,
  author    = {Kaiming He and Xiangyu Zhang and Shaoqing Ren and Jian Sun},
  title     = {Deep Residual Learning for Image Recognition},
  booktitle = {Proceedings of the IEEE Conference on Computer Vision and Pattern Recognition (CVPR)},
  pages     = {770--778},
  year      = {2016},
}

@inproceedings{imagenet_NIPS2012_c399862d,
  author    = {Alex Krizhevsky and Ilya Sutskever and Geoffrey E. Hinton},
  title     = {{ImageNet} Classification with Deep Convolutional Neural Networks},
  booktitle = {Advances in Neural Information Processing Systems},
  editor    = {F. Pereira and C. J. Burges and L. Bottou and K. Q. Weinberger},
  publisher = {Curran Associates, Inc.},
  volume    = {25},
  year      = {2012},
}

@article{Simonyan2014VeryDC,
  author  = {Karen Simonyan and Andrew Zisserman},
  title   = {Very Deep Convolutional Networks for Large-Scale Image Recognition},
  journal = {CoRR},
  volume  = {abs/1409.1556},
  year    = {2014},
}

@inproceedings{SzegedyGoingDeeper,
  author    = {Christian Szegedy and Wei Liu and Yangqing Jia and Pierre Sermanet and
               Scott Reed and Dumitru Erhan and Vincent Vanhoucke and Andrew Rabinovich},
  title     = {Going Deeper with Convolutions},
  booktitle = {IEEE Conference on Computer Vision and Pattern Recognition (CVPR)},
  pages     = {1--9},
  year      = {2015},
}

@InProceedings{pmlr-v139-touvron21a,
  title =     {Training data-efficient image transformers {\&} distillation through attention},
  author =    {Touvron, Hugo and Cord, Matthieu and Douze, Matthijs and Massa, Francisco and Sablayrolles, Alexandre and J{\'e}egou, Herv{\'e}},
  booktitle = {International Conference on Machine Learning},
  pages =     {10347--10357},
  year =      {2021},
  volume =    {139},
  month =     {July}
}

@inproceedings{tolstikhin2021mlpmixerallmlparchitecturevision,
 author = {Tolstikhin, Ilya O and Houlsby, Neil and Kolesnikov, Alexander and Beyer, Lucas and Zhai, Xiaohua and Unterthiner, Thomas and Yung, Jessica and Steiner, Andreas and Keysers, Daniel and Uszkoreit, Jakob and Lucic, Mario and Dosovitskiy, Alexey},
 booktitle = {Advances in Neural Information Processing Systems},
 editor = {M. Ranzato and A. Beygelzimer and Y. Dauphin and P.S. Liang and J. Wortman Vaughan},
 pages = {24261--24272},
 publisher = {Curran Associates, Inc.},
 title = {{MLP-Mixer}: An All-{MLP} Architecture for Vision},
 url = {https://proceedings.neurips.cc/paper_files/paper/2021/file/cba0a4ee5ccd02fda0fe3f9a3e7b89fe-Paper.pdf},
 volume = {34},
 year = {2021}
}

@article{vaswani2017attention,
  author  = {Ashish Vaswani and Noam Shazeer and Niki Parmar and Jakob Uszkoreit and
             Llion Jones and Aidan N. Gomez and {\L}ukasz Kaiser and Illia Polosukhin},
  title   = {Attention Is All You Need},
  journal = {Advances in Neural Information Processing Systems},
  volume  = {30},
  year    = {2017},
}

@inproceedings{devlin2019bert,
  author    = {Jacob Devlin and Ming-Wei Chang and Kenton Lee and Kristina Toutanova},
  title     = {{BERT}: Pre-Training of Deep Bidirectional Transformers for Language
               Understanding},
  booktitle = {Proceedings of the 2019 Conference of the North American Chapter of the
               Association for Computational Linguistics: Human Language Technologies,
               Volume 1 (Long and Short Papers)},
  pages     = {4171--4186},
  year      = {2019},
}

@article{hoffmann2022training,
  author  = {Jordan Hoffmann and Sebastian Borgeaud and Arthur Mensch and Elena Buchatskaya
             and Trevor Cai and Eliza Rutherford and Diego de Las Casas and Lisa Anne Hendricks
             and Johannes Welbl and Aidan Clark and others},
  title   = {Training Compute-Optimal Large Language Models},
  journal = {arXiv preprint arXiv:2203.15556},
  year    = {2022},
}

@misc{deepseekai2026deepseekv4,
  author = {{DeepSeek-AI}},
  title  = {{DeepSeek-V4}: Towards Highly Efficient Million-Token Context Intelligence},
  year   = {2026},
}

@article{singh2026arcee,
  author  = {Varun Singh and Lucas Krauss and Sami Jaghouar and Matej Sirovatka and
             Charles Goddard and Fares Obied and Jack Min Ong and Jannik Straube and
             Aria Harley and Conner Stewart and others},
  title   = {{Arcee Trinity Large} Technical Report},
  journal = {arXiv preprint arXiv:2602.17004},
  year    = {2026},
}

@article{team2025kimi,
  author  = {{Kimi Team} and Yifan Bai and Yiping Bao and Y. Charles and Cheng Chen and
             Guanduo Chen and Haiting Chen and Huarong Chen and Jiahao Chen and Ningxin Chen
             and others},
  title   = {Kimi {K2}: Open Agentic Intelligence},
  journal = {arXiv preprint arXiv:2507.20534},
  year    = {2025},
}

@article{team2026kimi,
  author  = {{Kimi Team} and Tongtong Bai and Yifan Bai and Yiping Bao and S. H. Cai and
             Yuan Cao and Y. Charles and H. S. Che and Cheng Chen and Guanduo Chen and others},
  title   = {Kimi {K2.5}: Visual Agentic Intelligence},
  journal = {arXiv preprint arXiv:2602.02276},
  year    = {2026},
}

@misc{karpathy2023nanogpt,
  author       = {Andrej Karpathy},
  title        = {{nanoGPT}},
  year         = {2023},
  howpublished = {\url{https://github.com/karpathy/nanoGPT}},
  note         = {Accessed: 2025-01-25},
}

@inproceedings{penedo2024the,
  author    = {Guilherme Penedo and Hynek Kydl{\'i}{\v{c}}ek and Loubna {Ben Allal} and
               Anton Lozhkov and Margaret Mitchell and Colin Raffel and Leandro {Von Werra}
               and Thomas Wolf},
  title     = {The {FineWeb} Datasets: Decanting the Web for the Finest Text Data at Scale},
  booktitle = {The Thirty-Eighth Conference on Neural Information Processing Systems
               Datasets and Benchmarks Track},
  year      = {2024},
}

@article{sutskever2014sequence,
  author  = {Ilya Sutskever and Oriol Vinyals and Quoc V. Le},
  title   = {Sequence to Sequence Learning with Neural Networks},
  journal = {Advances in Neural Information Processing Systems},
  volume  = {27},
  year    = {2014},
}

@article{HintonDeepNeuralNetworksForAcousticModeling,
  author  = {Geoffrey Hinton and Li Deng and Dong Yu and George Dahl and
             Abdel-rahman Mohamed and Navdeep Jaitly and Andrew Senior and
             Vincent Vanhoucke and Phuongtrang Nguyen and Tara Sainath and Brian Kingsbury},
  title   = {Deep Neural Networks for Acoustic Modeling in Speech Recognition:
             The Shared Views of Four Research Groups},
  journal = {IEEE Signal Processing Magazine},
  volume  = {29},
  pages   = {82--97},
  year    = {2012},
}

@article{Graves2013SpeechRW,
  author  = {Alex Graves and Abdel-rahman Mohamed and Geoffrey E. Hinton},
  title   = {Speech Recognition with Deep Recurrent Neural Networks},
  journal = {2013 IEEE International Conference on Acoustics, Speech and Signal Processing},
  pages   = {6645--6649},
  year    = {2013},
}

@inproceedings{Amodei2015DeepS2,
author = {Amodei, Dario and Ananthanarayanan, Sundaram and Anubhai, Rishita and Bai, Jingliang and Battenberg, Eric and Case, Carl and Casper, Jared and Catanzaro, Bryan and Cheng, Qiang and Chen, Guoliang and Chen, Jie and Chen, Jingdong and Chen, Zhijie and Chrzanowski, Mike and Coates, Adam and Diamos, Greg and Ding, Ke and Du, Niandong and Elsen, Erich and Engel, Jesse and Fang, Weiwei and Fan, Linxi and Fougner, Christopher and Gao, Liang and Gong, Caixia and Hannun, Awni and Han, Tony and Johannes, Lappi Vaino and Jiang, Bing and Ju, Cai and Jun, Billy and LeGresley, Patrick and Lin, Libby and Liu, Junjie and Liu, Yang and Li, Weigao and Li, Xiangang and Ma, Dongpeng and Narang, Sharan and Ng, Andrew and Ozair, Sherjil and Peng, Yiping and Prenger, Ryan and Qian, Sheng and Quan, Zongfeng and Raiman, Jonathan and Rao, Vinay and Satheesh, Sanjeev and Seetapun, David and Sengupta, Shubho and Srinet, Kavya and Sriram, Anuroop and Tang, Haiyuan and Tang, Liliang and Wang, Chong and Wang, Jidong and Wang, Kaifu and Wang, Yi and Wang, Zhijian and Wang, Zhiqian and Wu, Shuang and Wei, Likai and Xiao, Bo and Xie, Wen and Xie, Yan and Yogatama, Dani and Yuan, Bin and Zhan, Jun and Zhu, Zhenyao},
title = {Deep speech 2: end-to-end speech recognition in English and mandarin},
year = {2016},
publisher = {JMLR.org},
booktitle = {Proceedings of the 33rd International Conference on International Conference on Machine Learning - Volume 48},
pages = {173–182},
numpages = {10},
location = {New York, NY, USA},
series = {ICML'16}
}

@article{esteva2017dermatologist,
  author    = {Andre Esteva and Brett Kuprel and Roberto A. Novoa and Justin Ko and
               Susan M. Swetter and Helen M. Blau and Sebastian Thrun},
  title     = {Dermatologist-Level Classification of Skin Cancer with Deep Neural Networks},
  journal   = {Nature},
  volume    = {542},
  number    = {7639},
  pages     = {115--118},
  year      = {2017},
  publisher = {Nature Publishing Group UK London},
}

@article{topol2019high,
  author    = {Eric J. Topol},
  title     = {High-Performance Medicine: The Convergence of Human and Artificial Intelligence},
  journal   = {Nature Medicine},
  volume    = {25},
  number    = {1},
  pages     = {44--56},
  year      = {2019},
  publisher = {Nature Publishing Group US New York},
}

@inproceedings{julian2019guaranteeing,
  author       = {Kyle D. Julian and Mykel J. Kochenderfer},
  title        = {Guaranteeing Safety for Neural Network-Based Aircraft Collision Avoidance
                  Systems},
  booktitle    = {2019 IEEE/AIAA 38th Digital Avionics Systems Conference (DASC)},
  pages        = {1--10},
  year         = {2019},
  organization = {IEEE},
}

@article{sun2019review,
  author    = {Gang Sun and Shuyue Wang},
  title     = {A Review of the Artificial Neural Network Surrogate Modeling in Aerodynamic
               Design},
  journal   = {Proceedings of the Institution of Mechanical Engineers, Part {G}: Journal of
               Aerospace Engineering},
  volume    = {233},
  number    = {16},
  pages     = {5863--5872},
  year      = {2019},
  publisher = {SAGE Publications Sage UK: London, England},
}

@article{tang2022deep,
  author    = {Chenwei Tang and Caiyang Yu and Yi Gao and Jianming Chen and Jiaming Yang and
               Jiuling Lang and Chuan Liu and Ling Zhong and Zhenan He and Jiancheng Lv},
  title     = {Deep Learning in Nuclear Industry: A Survey},
  journal   = {Big Data Mining and Analytics},
  volume    = {5},
  number    = {2},
  pages     = {140--160},
  year      = {2022},
  publisher = {TUP},
}

@article{hu2021data,
  author    = {Guang Hu and Taotao Zhou and Qianfeng Liu},
  title     = {Data-Driven Machine Learning for Fault Detection and Diagnosis in Nuclear
               Power Plants: A Review},
  journal   = {Frontiers in Energy Research},
  volume    = {9},
  pages     = {663296},
  year      = {2021},
  publisher = {Frontiers Media SA},
}

@inproceedings{GenBound17,
  author    = {Peter L. Bartlett and Dylan J. Foster and Matus J. Telgarsky},
  title     = {Spectrally-Normalized Margin Bounds for Neural Networks},
  booktitle = {Advances in Neural Information Processing Systems},
  editor    = {I. Guyon and U. Von Luxburg and S. Bengio and H. Wallach and R. Fergus
               and S. Vishwanathan and R. Garnett},
  publisher = {Curran Associates, Inc.},
  volume    = {30},
  year      = {2017},
}

@article{sokolicrobustlargemargin,
title = "Robust Large Margin Deep Neural Networks",
author = "Jure Sokoli{\'c} and Raja Giryes and Guillermo Sapiro and Rodrigues, \{Miguel R.D.\}",
note = "Publisher Copyright: {\textcopyright} 2017 IEEE.",
year = "2017",
month = aug,
day = "15",
doi = "10.1109/TSP.2017.2708039",
language = "English (US)",
volume = "65",
pages = "4265--4280",
journal = "IEEE Transactions on Signal Processing",
issn = "1053-587X",
publisher = "Institute of Electrical and Electronics Engineers Inc.",
number = "16",

}

@inproceedings{goodfellow2015explainingharnessingadversarialexamples,
  author    = {Ian J. Goodfellow and Jonathon Shlens and Christian Szegedy},
  title     = {Explaining and Harnessing Adversarial Examples},
  booktitle = {3rd International Conference on Learning Representations ({ICLR} 2015),
               San Diego, CA, USA, May 7--9, 2015, Conference Track Proceedings},
  editor    = {Yoshua Bengio and Yann LeCun},
  year      = {2015},
}

@article{kovarik1970some,
  title={Some iterative methods for improving orthonormality},
  author={Kovarik, Zdislav},
  journal={SIAM Journal on Numerical Analysis},
  volume={7},
  number={3},
  pages={386--389},
  year={1970},
  publisher={SIAM}
}

@article{bjorck1971iterative,
  title={An iterative algorithm for computing the best estimate of an orthogonal matrix},
  author={Bj{\"o}rck, {\AA}ke and Bowie, Clazett},
  journal={SIAM Journal on Numerical Analysis},
  volume={8},
  number={2},
  pages={358--364},
  year={1971},
  publisher={SIAM}
}

@article{newhouse2025training,
  title={Training transformers with enforced {Lipschitz} constants},
  author={Newhouse, Laker and Hess, R Preston and Cesista, Franz and Zahorodnii, Andrii and Bernstein, Jeremy and Isola, Phillip},
  journal={arXiv preprint arXiv:2507.13338},
  year={2025}
}

@misc{nemo_emerging_optimizers,
  author       = {{NVIDIA NeMo Team}},
  title        = {Emerging Optimizers},
  year         = {2025},
  publisher    = {GitHub},
  howpublished = {\url{https://github.com/NVIDIA-NeMo/Emerging-Optimizers}},
  note         = {NVIDIA NeMo, \url{https://docs.nvidia.com/nemo/emerging-optimizers/}}
}

@article{ren2026rethinking,
  title={Rethinking language model scaling under transferable hypersphere optimization},
  author={Ren, Liliang and Liu, Yang and Shen, Yelong and Chen, Weizhu},
  journal={arXiv preprint arXiv:2603.28743},
  year={2026}
}

@article{shigida2026learning,
  title={Learning Rate Transfer in Normalized Transformers},
  author={Shigida, Boris and Hanin, Boris and Gromov, Andrey},
  journal={arXiv preprint arXiv:2604.27077},
  year={2026}
}

@article{xie2025mhc,
  title={{mHC}: Manifold-constrained hyper-connections},
  author={Xie, Zhenda and Wei, Yixuan and Cao, Huanqi and Zhao, Chenggang and Deng, Chengqi and Li, Jiashi and Dai, Damai and Gao, Huazuo and Chang, Jiang and Yu, Kuai and others},
  journal={arXiv preprint arXiv:2512.24880},
  year={2025}
}

@article{team2026attention,
  title={Attention residuals},
  author={{Kimi Team} and Chen, Guangyu and Zhang, Yu and Su, Jianlin and Xu, Weixin and Pan, Siyuan and Wang, Yaoyu and Wang, Yucheng and Chen, Guanduo and Yin, Bohong and others},
  journal={arXiv preprint arXiv:2603.15031},
  year={2026}
}

@article{islamov2026role,
  title={On the Role of Batch Size in Stochastic Conditional Gradient Methods},
  author={Islamov, Rustem and Machacek, Roman and Lucchi, Aurelien and Silveti-Falls, Antonio and Gorbunov, Eduard and Cevher, Volkan},
  journal={arXiv preprint arXiv:2603.21191},
  year={2026}
}

@inproceedings{carlson2015stochastic,
  title={Stochastic spectral descent for restricted {Boltzmann} machines},
  author={Carlson, David and Cevher, Volkan and Carin, Lawrence},
  booktitle={Artificial intelligence and statistics},
  pages={111--119},
  year={2015},
  organization={PMLR}
}

@article{carlson2015preconditioned,
  title={Preconditioned spectral descent for deep learning},
  author={Carlson, David E and Collins, Edo and Hsieh, Ya-Ping and Carin, Lawrence and Cevher, Volkan},
  journal={Advances in neural information processing systems},
  volume={28},
  year={2015}
}

@article{carlson2015graphical,
  title={Stochastic spectral descent for discrete graphical models},
  author={Carlson, David and Hsieh, Ya-Ping and Collins, Edo and Carin, Lawrence and Cevher, Volkan},
  journal={IEEE Journal of Selected Topics in Signal Processing},
  volume={10},
  number={2},
  pages={296--311},
  year={2015},
  publisher={IEEE}
}

@article{chou2025correction,
  title={Correction of Decoupled Weight Decay},
  author={Chou, Jason Chuan-Chih},
  journal={arXiv preprint arXiv:2512.08217},
  year={2025}
}
